\documentclass[preprint,11pt]{elsarticle}

\usepackage{amssymb}
\usepackage{amsmath}

\usepackage{graphicx}
\usepackage[]{epstopdf}
\usepackage{algorithm2e}
\usepackage{algpseudocode}

\def\Bbb#1{\mathbb{#1}}
\def\Mca#1{\mathcal{#1}}
\DeclareMathOperator{\mean}{mean}
\def\supp{\mbox{\rm supp }}

\input comment.sty
\excludecomment{estensione}

\journal{Applied Numerical Mathematics}

\begin{document}

\begin{frontmatter}

\title{Analytic heuristics for a fast DSC-MRI}

\author{M.Virgulin$^1$, M.Castellaro$^2$, E.Grisan$^2$, F.Marcuzzi*$^1$\footnote{* Corresponding author}}
\address{$^1$Department of Mathematics, Padua University, Via Trieste 63, 35131 Padova, Italy
$^2$Department of Information Engineering, Padua University, Via Gradenigo 6/B, 35131 Padova, Italy}
\ead{marcuzzi@math.unipd.it}

\begin{abstract}
In this paper we propose a deterministic approach for the reconstruction of Dynamic Susceptibility Contrast magnetic resonance imaging data and compare it with the compressed sensing solution existing in the literature for the same problem. Our study is based on the mathematical analysis of the problem, which is computationally intractable because of its non polynomial complexity, but suggests simple heuristics that perform quite well. We give results on real images and on artificial phantoms with added noise.
\end{abstract}


\end{frontmatter}

\section{Introduction}
\label{intro}

The magnetic resonance images formation process (MRI) is based on radio-frequency energy absorption and emission of the hydrogen atoms nucleus: by varying the intensity of two orthogonal magnetic fields on a particular region, we can obtain an echo signal on a wide frequency spectrum. The physical image in the spatial coordinates, which represents the spatial distribution of hydrogen atoms density, is reconstructed by applying the bi-dimensional inverse Fourier transform of the echo frequency signal \cite{LL00}.

In this paper we consider in particular the Dynamic Susceptibility Contrast MRI (DSC-MRI) \cite{Os96a, Os96b}: a temporal sequence of images is reconstructed while a para-magnetic contrast liquid flows into the vessels of the human brain. This experiment is fundamental for the analysis of the emo-dynamics of the region. This sequence of images, whose purpose is to highlight the flow path of the contrast liquid, exhibits a strong spatiotemporal correlation, that can be exploited to predict which measurements will bring mostly the new information contained in the next frames.

This is crucial because in principle, for each frame, in order to reconstruct the whole spatial image we need to measure the complete spectrum in the frequency domain, but in practice, it is highly recommended a rapid-MRI scheme involving the smallest possible number of in-vivo measurements (i.e. a frequency under-sampling), so the time to sample each frame is reduced.
In general, the sampling speed is an important issue in many applications of the MRI, e.g. some medical exams need several minutes to be done, and this can represent a limit to this kind of exams. Sampling speed is limited by constraints that can be physical (e.g. sampling frequency) and physiological (e.g. bloodstream). The focus of many current researches is to study methods to reduce the number of measurement samples needed for each frame without degrading the image quality \cite{LDP07}.
For the DSC-MRI, the frequency under-sampling of single images does not imply automatically the decreasing of the duration of the whole clinic procedure (the flow of the contrast liquid has inevitable physical speed limits), but it can be exploited to make more frequent space or time acquisitions: the second option increases the time resolution and allows the analysis of fast dynamics not observed at this moment.

When the frequency domain is under-sampled, the Nyquist criterion is in general violated and the Fourier reconstruction shows aliasing artifacts. 
In DSC-MRI, however, by exploiting the redundancy in MR images, it is possible to define methods for rapid reconstructions that produce minimal artifacts in the recovered images.
Several techniques can be used to reduce the number of measures and at the same time keep the alias in artifacts small: some methods generates incoherent or less visual apparent artifacts at the cost of a reduced signal-noise ratio (SNR), other methods exploit the redundancy of the signal to recover, in its domain or in other domains \cite{LDP07}.

A novel set of techniques exploits the fact that, although the image to recover can be dense, its representation in other domains can be sparse. This is the case of medical images, that many transforms can make sparse, e.g. the wavelet transform \cite{Da92, GM84}. The possibility of reducing the memory space needed to store the information of an image has set up the question if it could be possible even to undersample the measured signal (in our case the frequency spectrum) in order to recover the image, apparently violating the Nyquist criterion. This approach goes under the name of \textit{Compressed Sensing} \cite{Can06, DDEK12, Do06, Fo10} and from the last recent years there's a growing literature exploting this approach on several applications, and also the MRI \cite{LDP07, LCAMP}.

In the literature, notable results in sparse signals recovery have been achieved by Compressed Sensing (CS) techniques, and also for MRI applications \cite{LDP07, LCAMP}. However, these methods are based on statistical properties rather than deterministic ones. In particular they do not make any assumption on the possible relations between sampled and recovered signals. 
This lack of information limits performances; a recently proposed solution is introducing deterministic constraints in the CS procedure \cite{LCAMP}: the space-time correlation of the sequence of images allows to define \textit{locations contraints}, i.e. the prediction of where the image will evolve mostly in the next frames.
The difficult task is to choose the measurement points to form the minimal set that best describes this evolution. In CS this is made with a random approach.

In this paper we want to study analytically the compound \textit{Fourier + wavelet} transform, involved in the processes of reconstruction and sparsification of MR images, in order to define a deterministic technique for a rapid-MRI.
By studying the wavelet transform, known for producing sparse representations for natural images, we exploit deterministically the relations between the wavelet sparse representation of the recovered signal (MR image) and the frequency samples. We then compared image reconstructions implemented with deterministic and CS methods.

The article structure is the following: in Section \ref{problem} we define the problem of recovering a signal with sparse wavelet representation by undersampling its frequency spectrum; in Section \ref{transforms} we study the involved transforms (two particular compositions of the Fourier and wavelet transforms); in Section \ref{algorithms} we describe four different heuristic approaches for solving the problem; in Section \ref{iterative algorithms} we recall two well-known iterative algorithms for sparse recovery; in Section \ref{data} we describe the datasets used for the simulations; in Section \ref{comparison} we compare the results of all the implemented algorithms, while in Section \ref{conclusions} we propose some possible directions we can explore to obtain better results.

\section{The problem}
\label{problem}

Let $x\in \Bbb{R}^N$ be the signal to recover in the spatial coordinates (for simplicity we suppose in this section that the signal be one-dimensional), and $N \gg 0$. Let $y\in \Bbb{R}^N, y = Wx$ be the vector of its wavelet coefficients, where $W$ is the Discrete Wavelet Transform (DWT). Let $y$ be $n$-sparse, i.e. $y_i = 0\hspace{.2cm} \forall i\in I,\  I\subseteq\{1,..,N\},\  |I| = n,\  n\ll N,$ (except an arrangement of the terms). Let $f\in \Bbb{C}^N, f = Fx$ be the vector of the Fourier coefficients of $x$, where $F$ is the Discrete Fourier Transform (DFT).

We want to find an optimal $m$-undersampling of $f$ for recovering $y$, i.e. given the number of measurements $m$, with $m \geq n$ and possibly also $m \ll N$, we want to determine the set of indexes $J\subseteq\{1,..,N\}, |J| = m,$ such that the recovered signal $\hat{y} = WF^T \left( \begin{array}{c}
f_J \\
0 \end{array} \right)$, where $f_J = (f_h)_{h\in J}$, leads to a minimal error with respect to the original signal $y$, measured with the norm $||\hat{y}-y||_2$.

Note that between the wavelet and Fourier coefficients it holds the relation $FW^Ty = f$. Now, let $J$ a frequency undersampling, it holds $F_JW^Ty=f_J$, where $F_J$ denotes the submatrix of the rows of $F$ that correspond to the non-zero coefficients $f_J$ of $f$. Let us use the notation
$$F = \left[ \begin{array}{c}
F_J \\
F_R \end{array} \right], \quad  f = \left( \begin{array}{c}
f_J \\
f_R \end{array} \right), \quad   y = \left( \begin{array}{c}
y_I \\
y_S \end{array} \right).$$
By applying the recover $WF^T$ to the zero-filled $f_J$, we have
$$\begin{array}{ccc}
WF^T \left( \begin{array}{c}
f_J \\
0 \end{array} \right) & = & WF^T \left( f - \left( \begin{array}{c}
0 \\
f_R \end{array} \right)\right)\\
& = & y - W \left[ F_J^T F_R^T \right] \left( \begin{array}{c}
0 \\
f_R \end{array} \right)\\
& = & y - W\left( \begin{array}{c}
0 \\
F_R^Tf_R \end{array} \right).\end{array}$$

So we need to find the set of indexes $R$ that makes minimal the norm of the non-zero coefficients of the vector $W\left( \begin{array}{c}
0 \\
F_R^Tf_R \end{array} \right)$. Being $y$ sparse, it becomes relevant only to minimize the $n$ coefficients at indexes $I$ of the error array
$$
W\left( \begin{array}{c}
0 \\
F_R^Tf_R \end{array} \right)
$$
that is, if we decompose $W$ as
$$\left[ \begin{array}{cc}
W_{IJ} & W_{IR} \\
W_{SJ} & W_{SR} \end{array} \right],$$
it means to minimize the norm of the vector
\begin{equation}
e_{rec} = W_{IR}F_R^Tf_R \quad .
\label{reconstruction_error}
\end{equation}

\begin{figure}[!htbp]
\centering
  {\includegraphics[scale=0.27,clip]{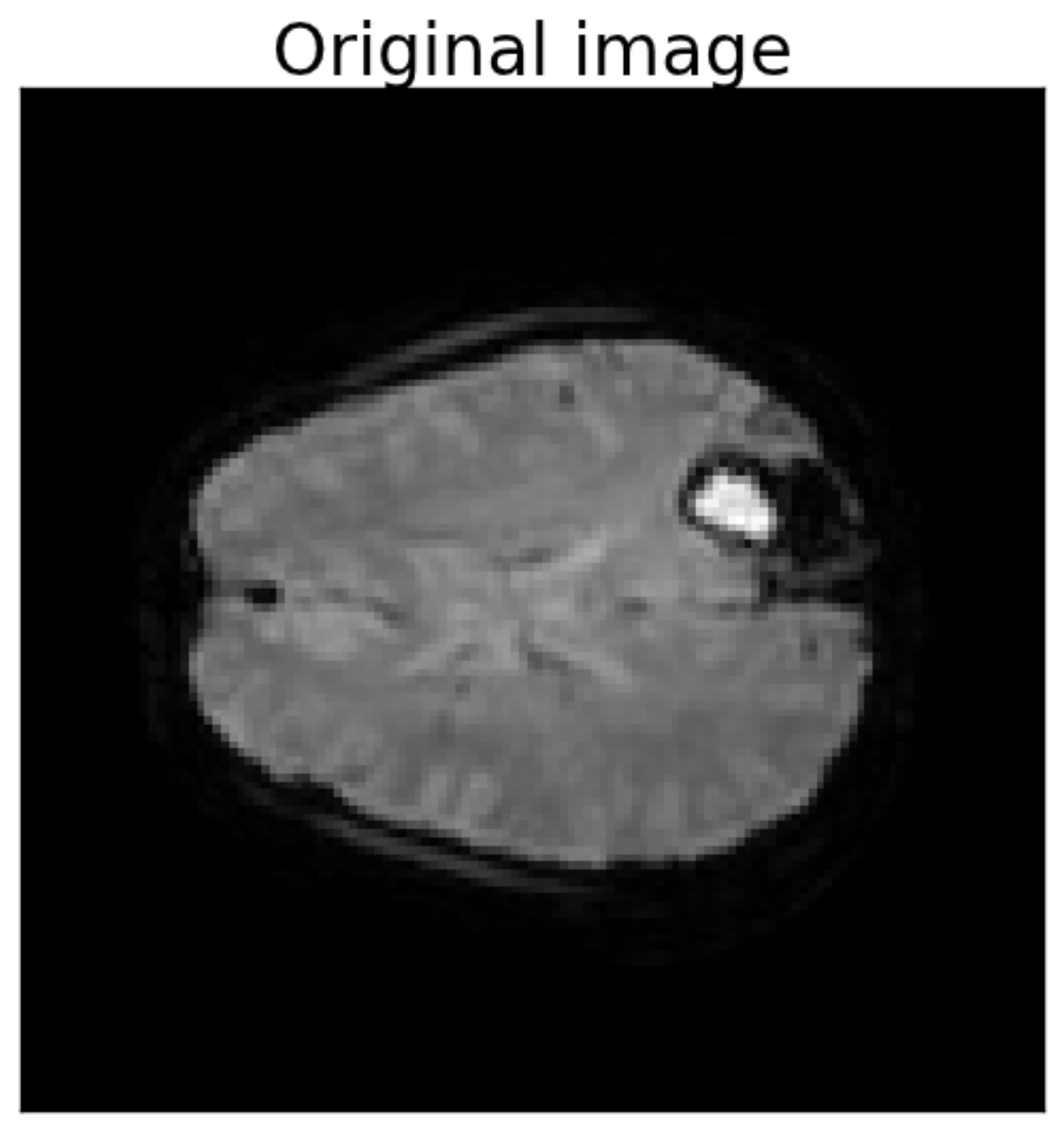}}\\
  {\includegraphics[scale=0.27,clip]{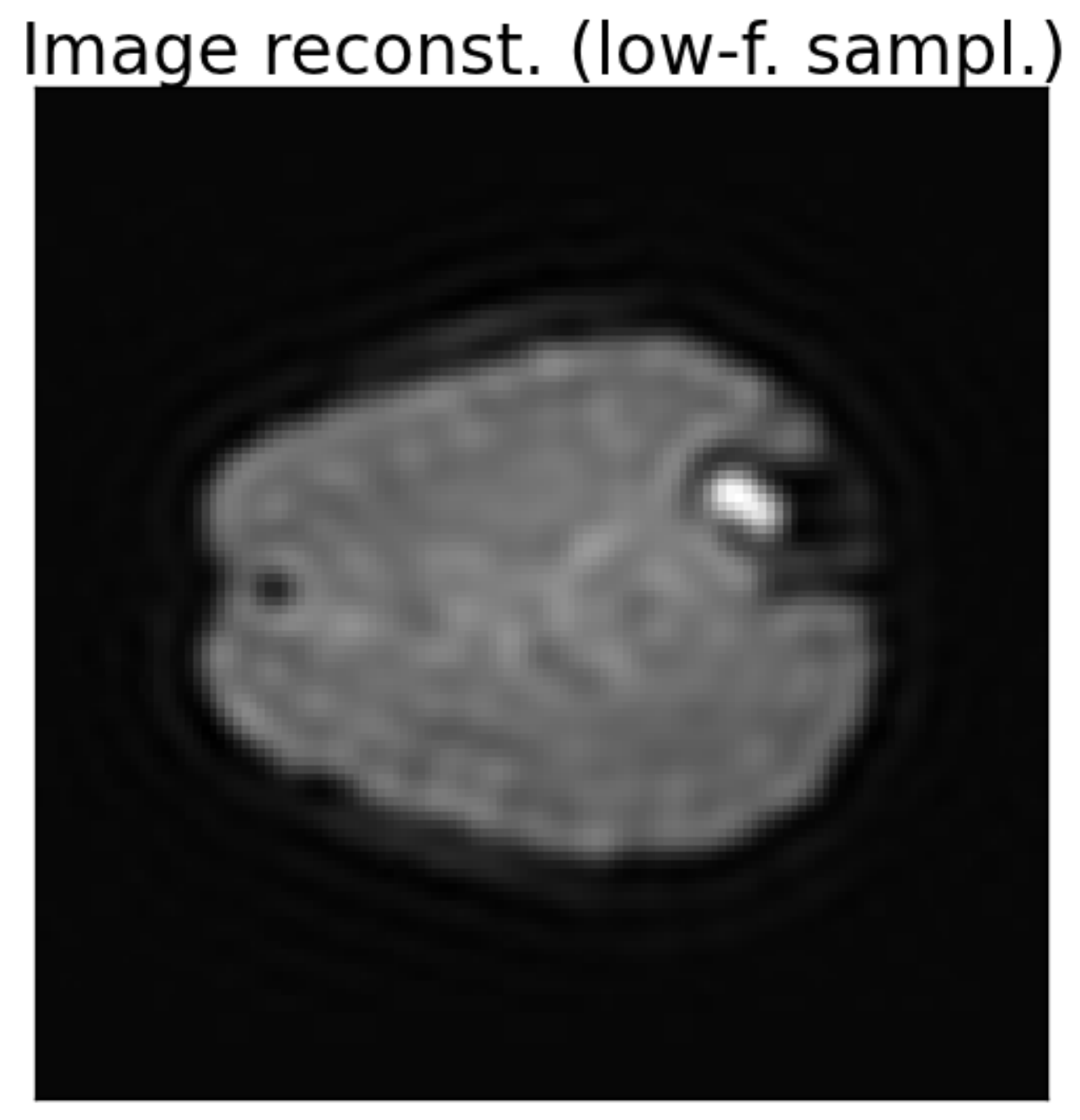}}
  {\includegraphics[scale=0.27,clip]{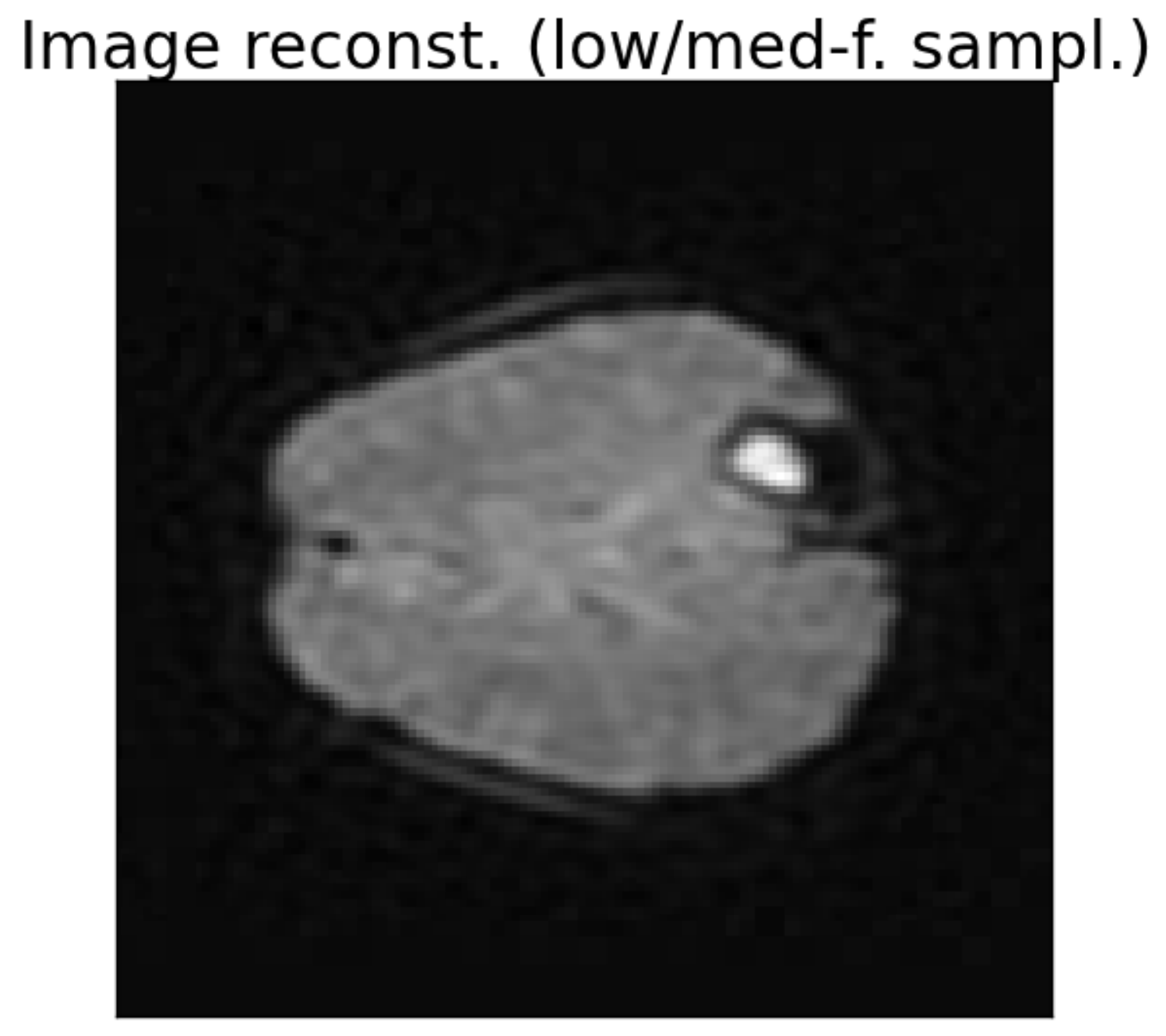}}
  {\includegraphics[scale=0.27,clip]{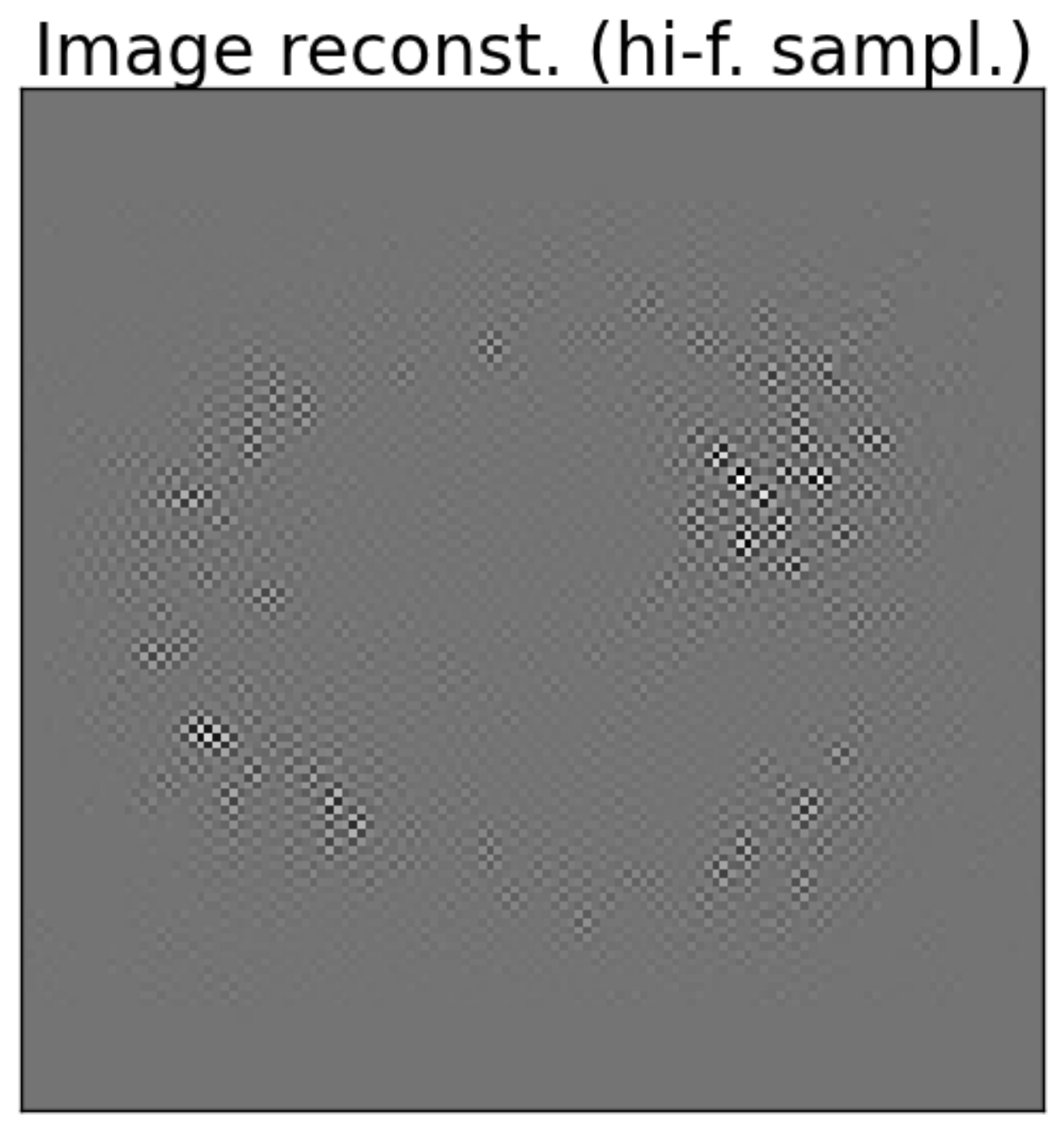}}\\
  {\includegraphics[scale=0.27,clip]{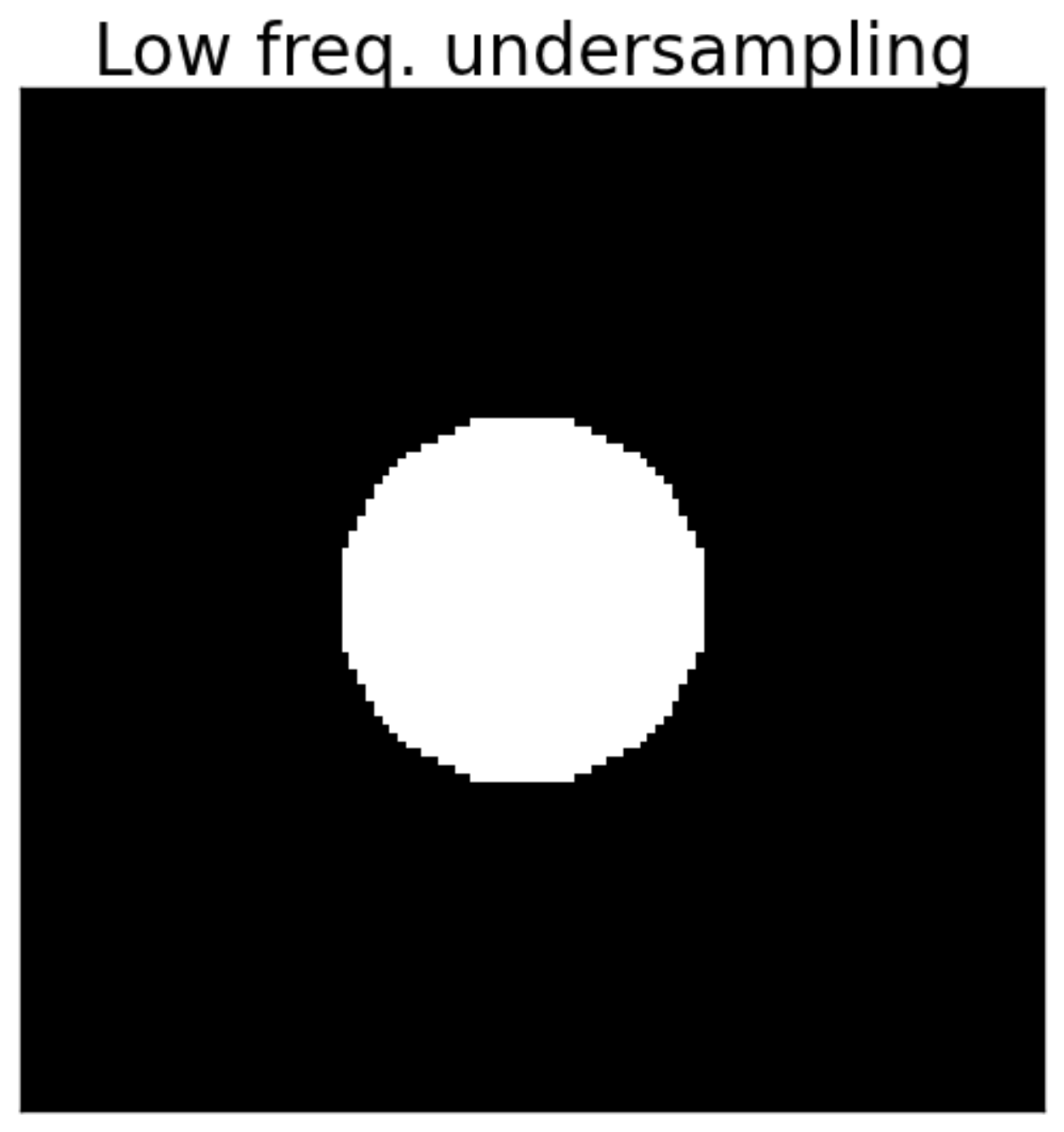}}
  {\includegraphics[scale=0.27,clip]{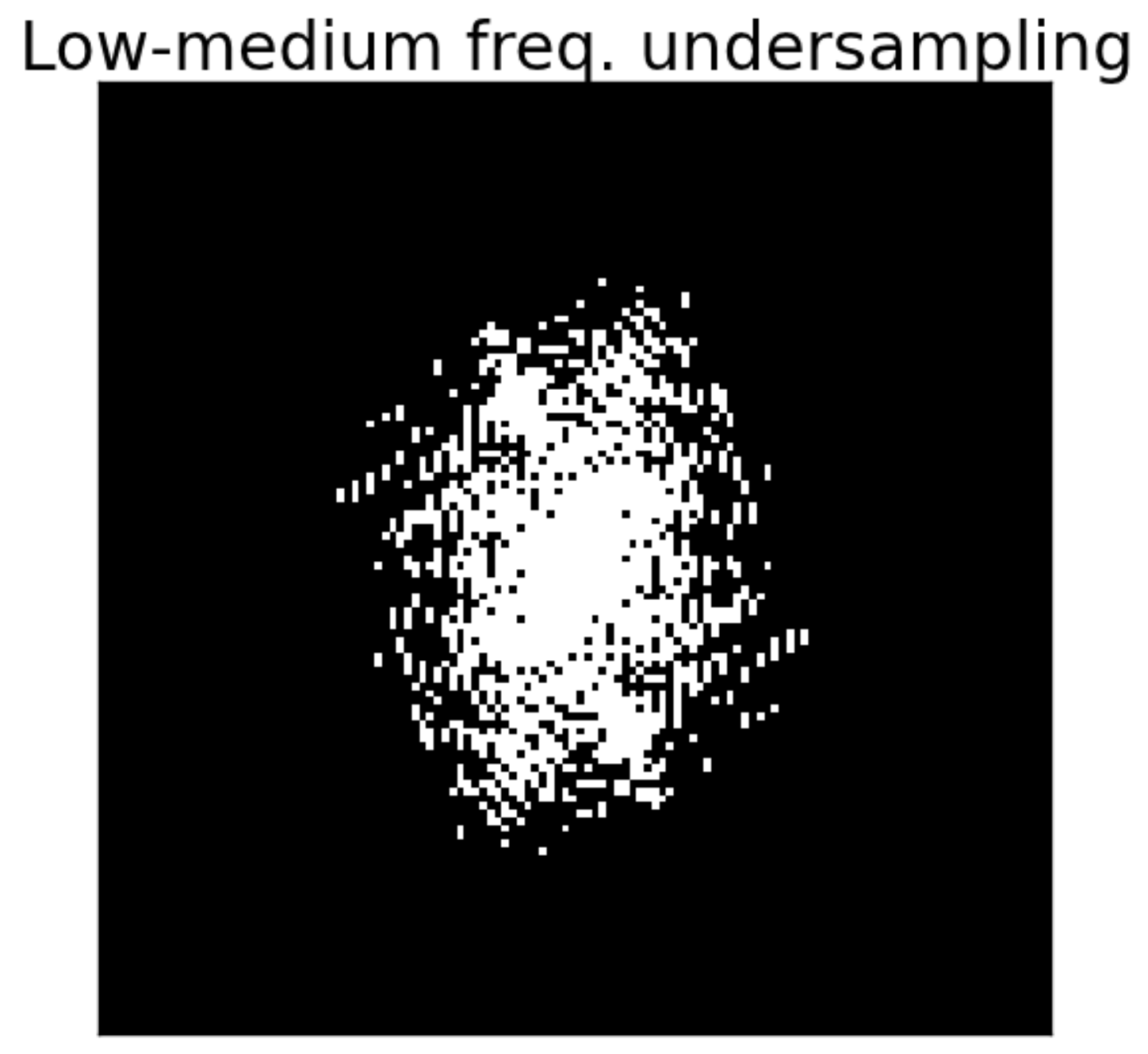}}
  {\includegraphics[scale=0.27,clip]{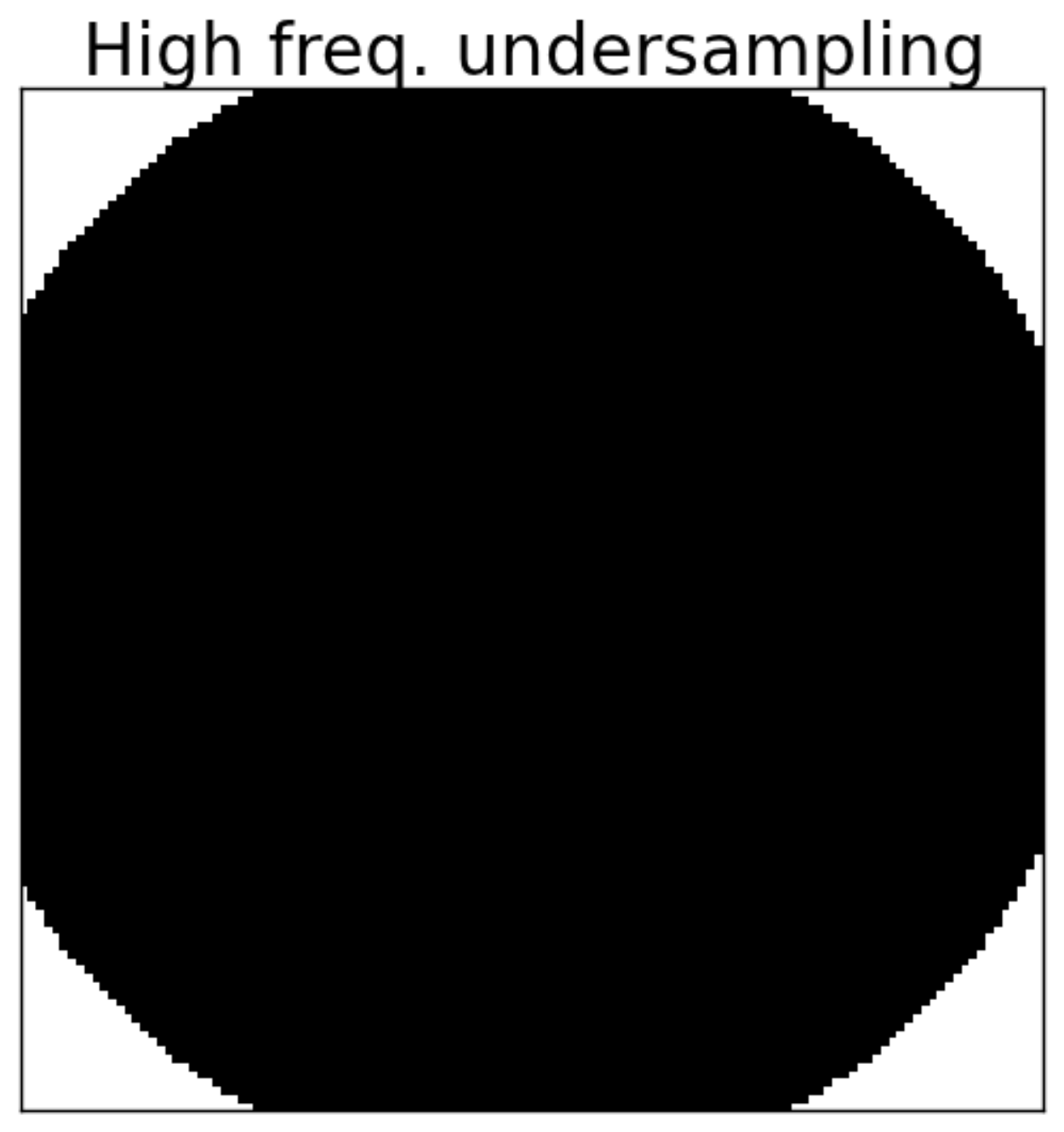}}
\caption{Examples of reconstructed images (second row) with three different sets of under-sampled frequency measures (third row) in low, medium/low and high frequency, fixed their cardinality to $m=10\%N$, compared to the original image (first row). The 2-norms of the error vectors $e_{rec}^{lo}, e_{rec}^{med/lo}, e_{rec}^{hi}$ are respectively $1384.28, 37147.66, 869.18.$ Note also the aliasing artifacts in the reconstructed images undersampled in low and medium/low frequency and how the second ones are more incoherent then the first ones.}
\end{figure}

The problem is in theory solvable combinatorially choosing the appropriate set $R$ (or $J$) of frequencies to disregard (or to measure). The number of possible choices for $J$, as $N$ and $m$ increase, becomes however too big to be solvable with brute-force search. As an example, the real images we analyzed had $N = 128^2 = 16384$ pixel, and if we wanted to perform the 25\% of the total possible measurements, the possible choices for $R$ or $J$ would be $\left( \begin{array}{c}
N \\
m \end{array} \right) = \left( \begin{array}{c}
16384 \\
4096 \end{array} \right) \approx 10^{4000}.$

The combinatorial approach is therefore intractable, so we defined other algorithms that could obtain good results to the same problem nevertheless.

\section{The compound transforms $\Mca{W}\circ\Mca{F}^{-1}$ and $\Mca{F}\circ\Mca{W}^{-1}$}
\label{transforms}

The analytic study of the problem is based on studying the composition of two transforms: the Inverse Fourier Transform $\Mca F^{-1}$ that reconstructs the image in the spatial domain from the measured frequencies, and the Wavelet Transform $\Mca W$ that sparsifies the image in its wavelet coefficients. In particular we focused on their compositions $\Mca{W}\circ\Mca{F}^{-1}$ and $\Mca{F}\circ\Mca{W}^{-1}$, as $\Mca W$ composed to the inverse of $\Mca F$ shows the error in the wavelet domain of every frequency disregarded by the undersampling, and $\Mca F$ composed to the inverse of $\Mca W$ shows the frequency spectrum of the sparse wavelet coefficients to sample.

In order to keep the notation as clean and simple as possible, we considered $\Mca F$ as the usual continue Fourier transform, while $\Mca W$ as the wavelet transform defined by a \textit{discrete} orthonormal basis of $L^2(\Bbb{R})$, since using the continuous wavelet transform would have requested to introduce some notions not relevant to our purposes.

Let $\psi\in L^2(\Bbb{R})$ be a \textit{discrete orthonormal mother wavelet}, i.e. let $\{\psi_{j,k}: \psi_{j,k}(z):=$ $2^{j/2}\psi(2^jz-k), j,k\in\Bbb{Z}\}$ be an orthonormal basis of $L^2(\Bbb{R})$.

Then it is easy to prove that for all $f:\Bbb{R}\rightarrow\Bbb{C}$ we have:
$$(\Mca{W}_\psi\circ\Mca{F}^{-1}f)(j,k) = \frac{1}{\sqrt{2\pi}}\int_{-\infty}^{+\infty} (\Mca{W}_\psi e^{i\omega\cdot})(j,k) f(\omega)d\omega, \hspace{.3cm}j,k\in\Bbb{Z},$$
and vice versa for all $y:\Bbb{Z}^2\rightarrow\Bbb{R}$ we have:
$$(\Mca{F}\circ\Mca{W}_\psi^{-1}y)(\omega)=\sum_{j,k=-\infty}^{+\infty}y_{j,k}\hat{\psi}_{j,k}(\omega), \hspace{.3cm}\omega\in\Bbb{R}.$$

\begin{figure}[!htbp]
 \centering
  {\includegraphics[scale=0.2,clip]{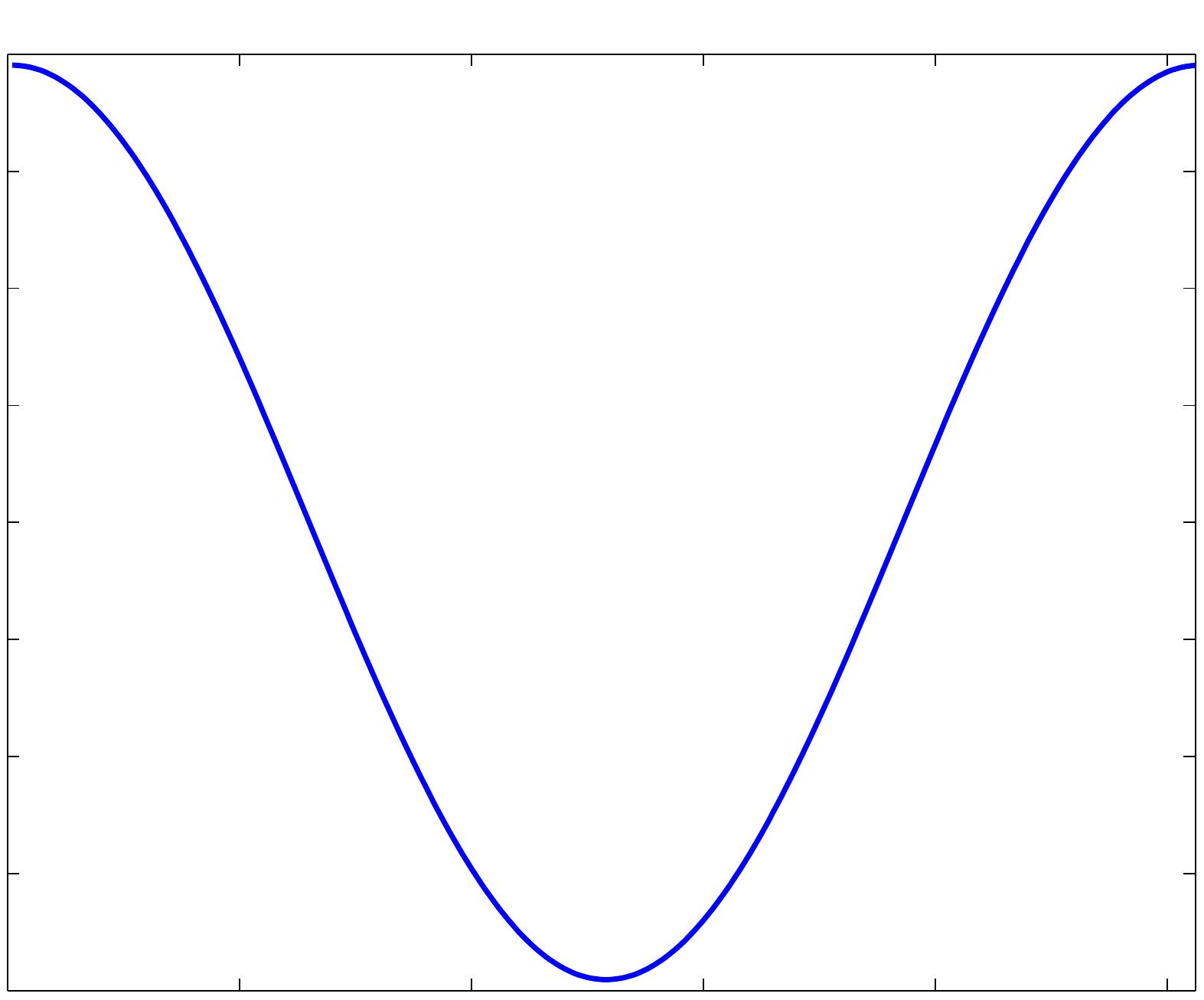}}
  {\includegraphics[scale=0.2,clip]{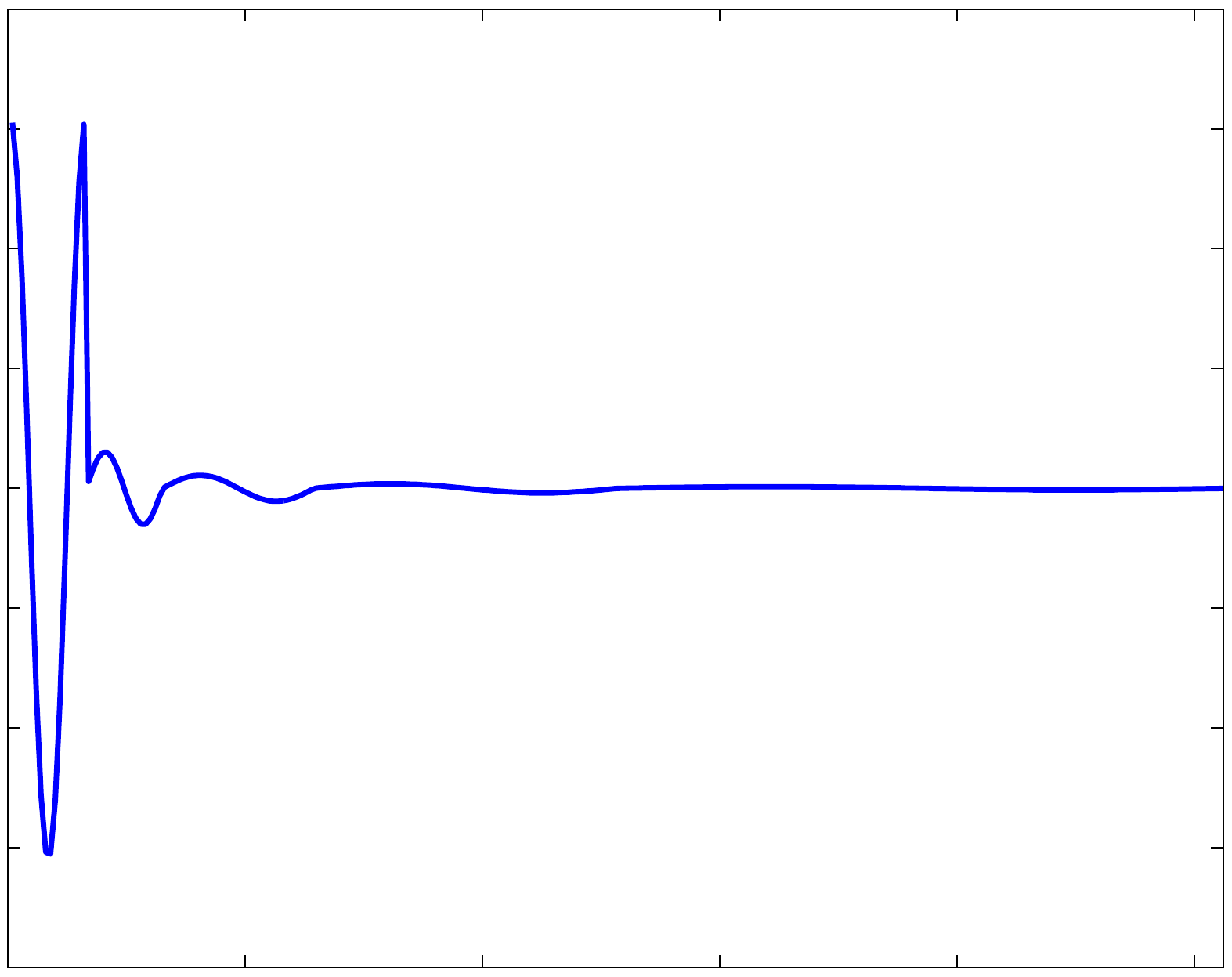}}\\
  {\includegraphics[scale=0.2,clip]{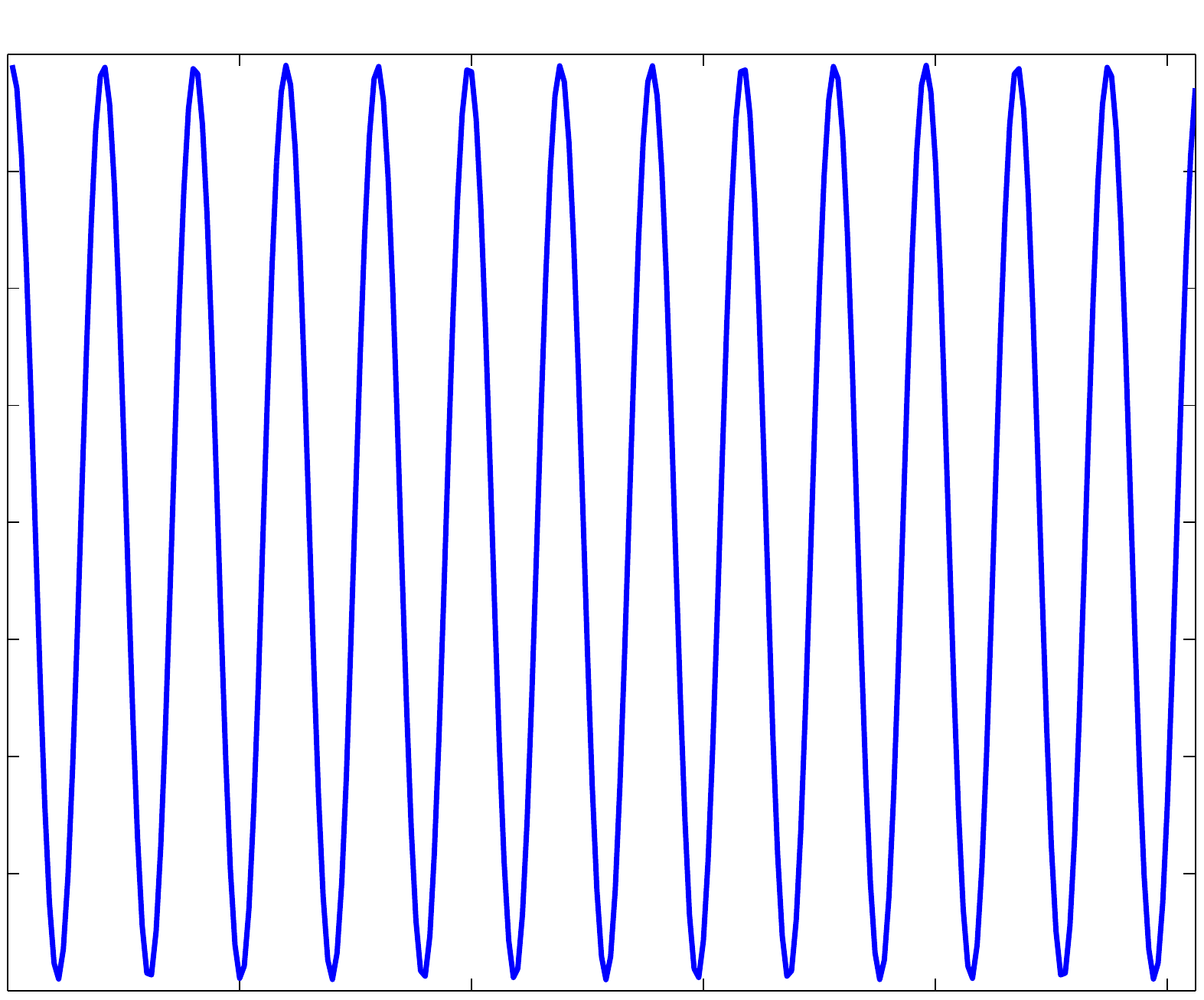}}
  {\includegraphics[scale=0.2,clip]{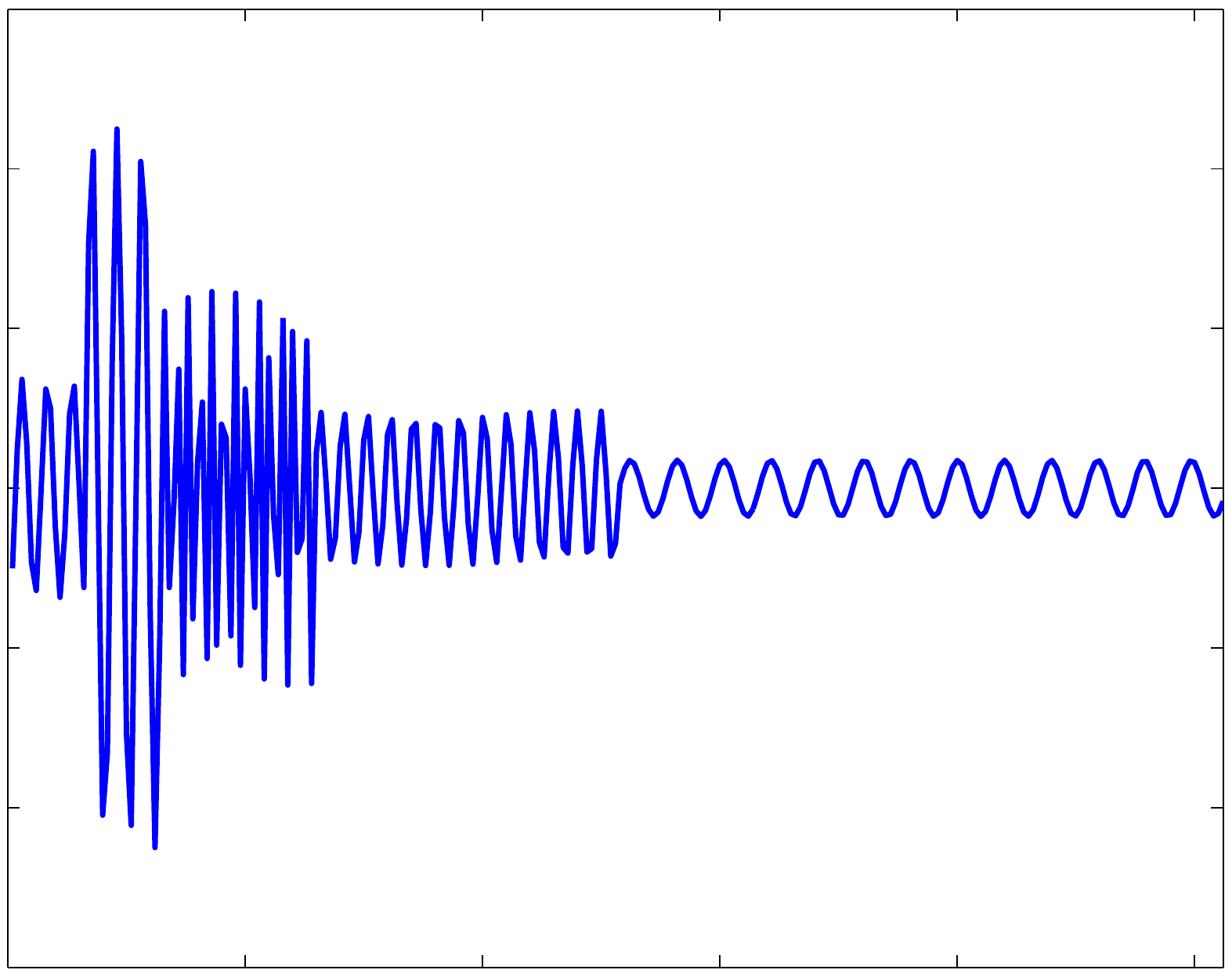}}\\
  {\includegraphics[scale=0.2,clip]{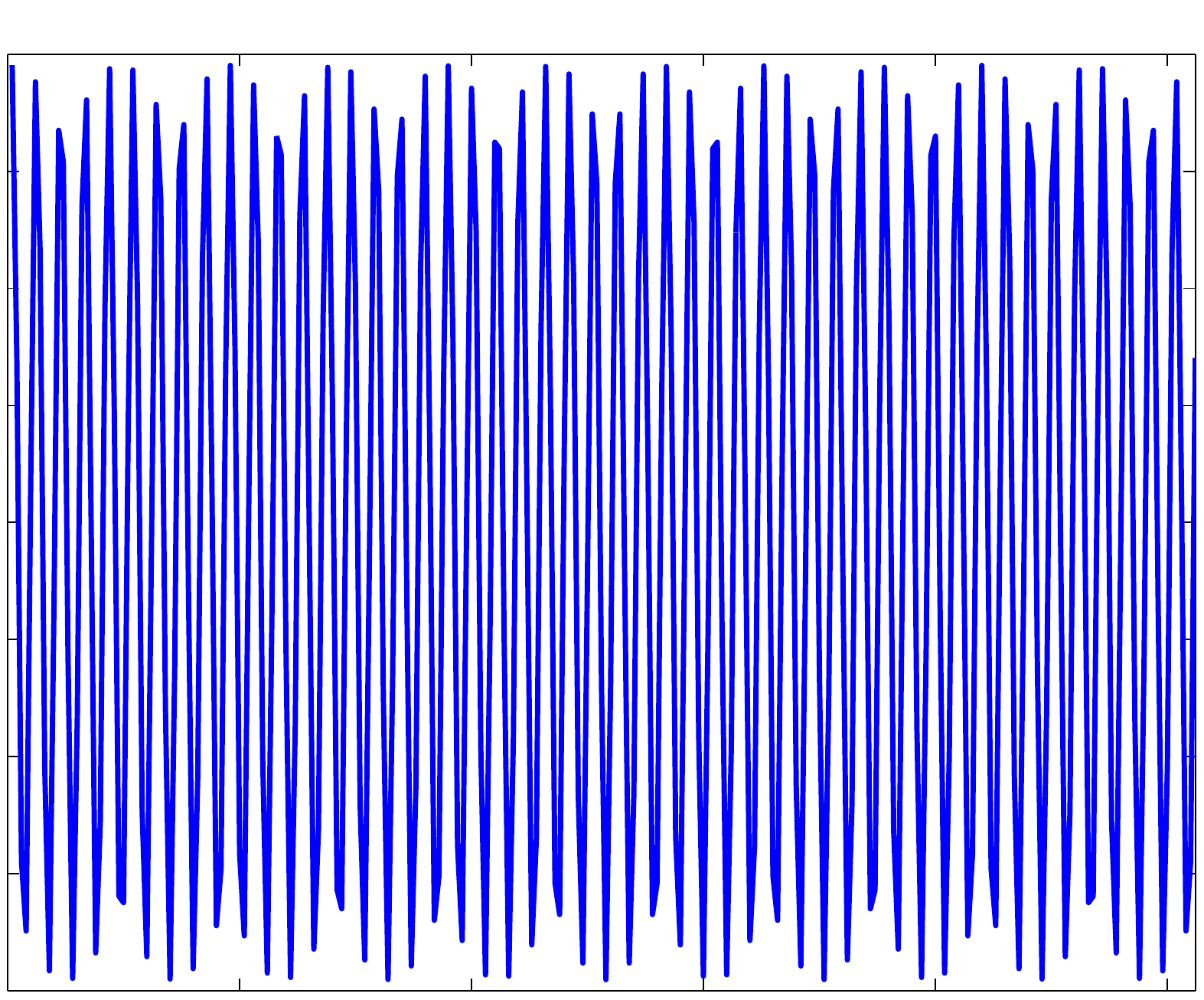}}
  {\includegraphics[scale=0.2,clip]{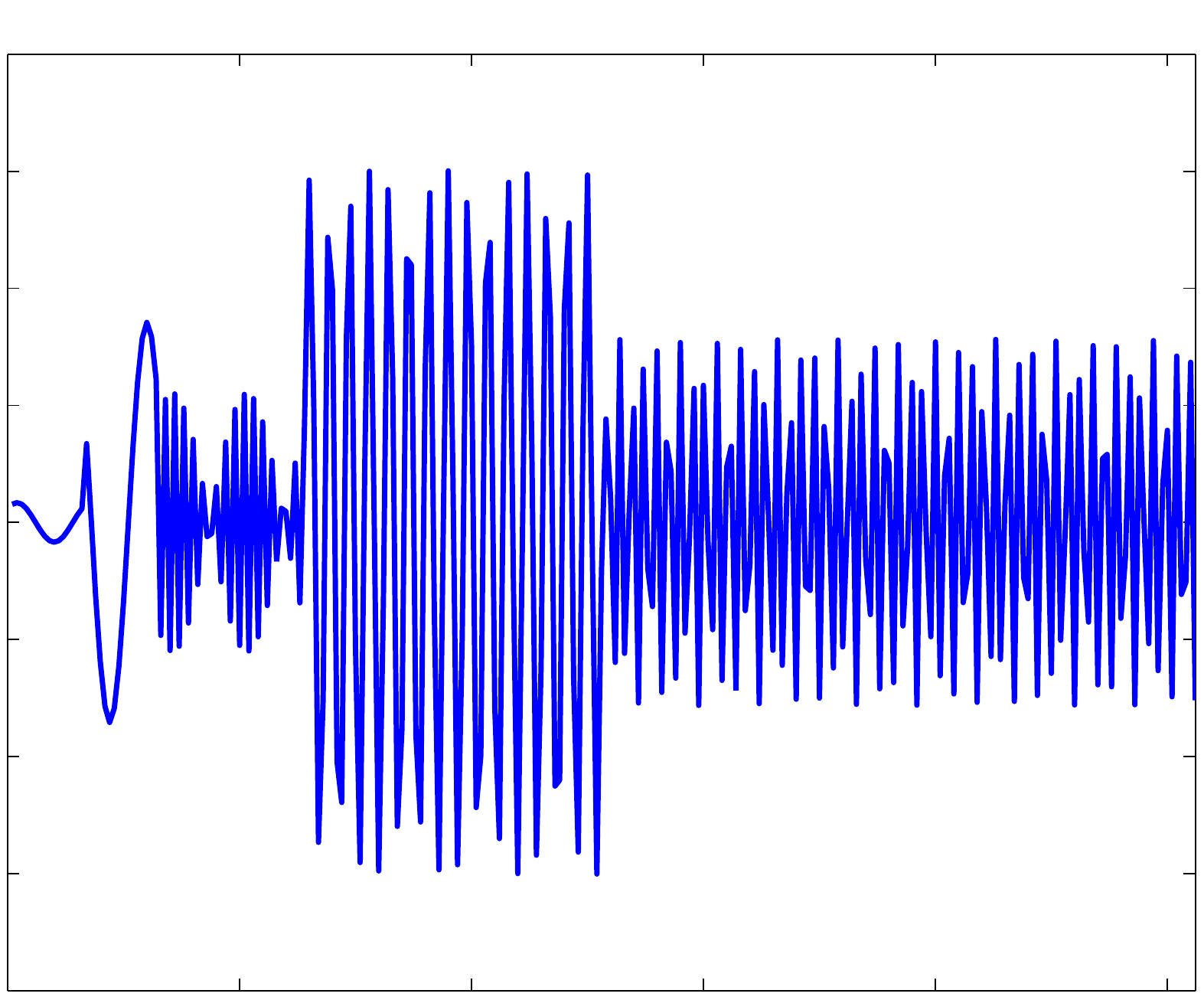}}\\
  {\includegraphics[scale=0.2,clip]{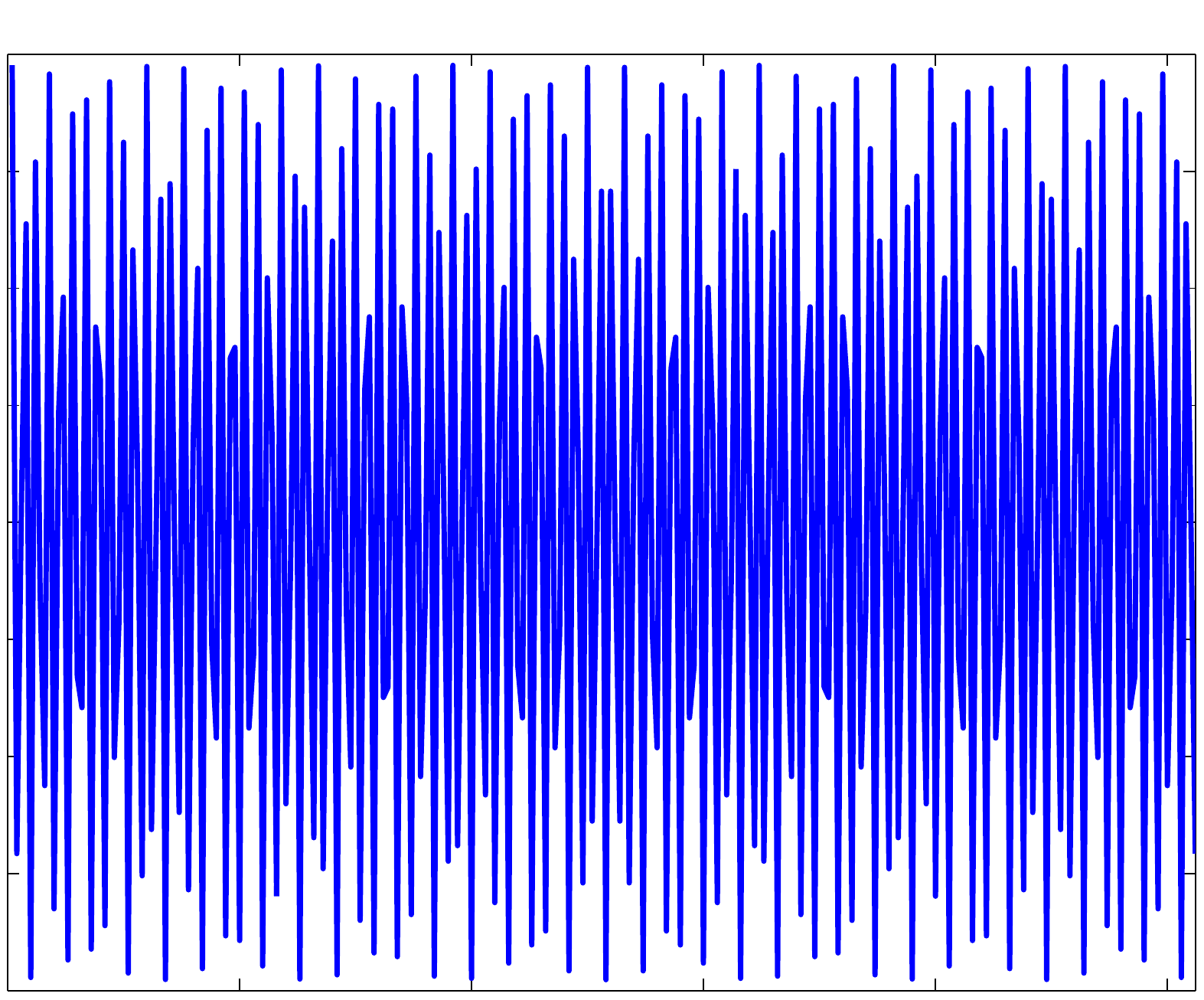}}
  {\includegraphics[scale=0.2,clip]{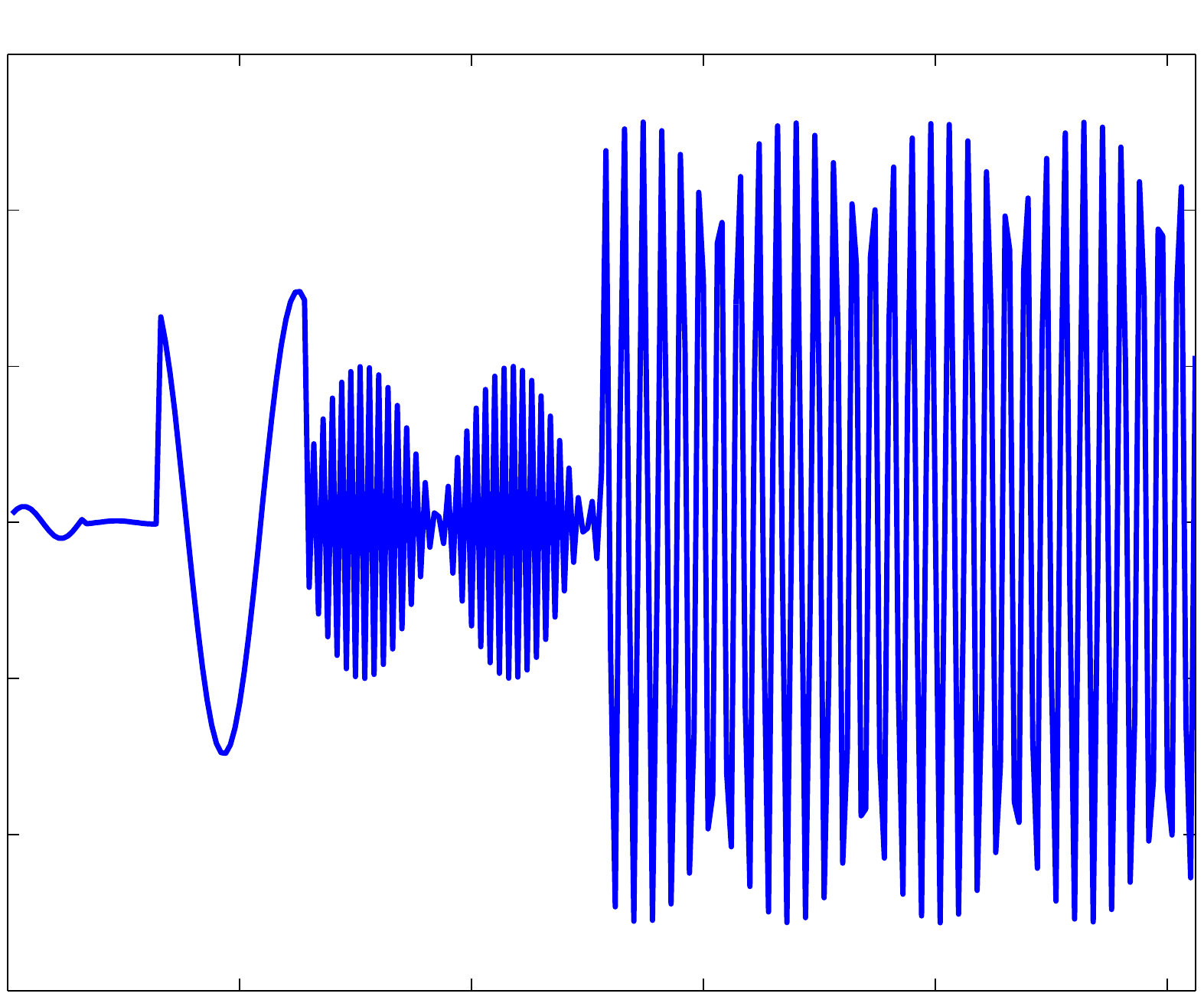}}
\caption{Examples of increasing frequencies and their one-dimensional Haar wavelet of level 4: note that higher frequencies excite higher resolution wavelet coefficients.}
\label{fig:FtoW}
\end{figure}

In particular, if we consider a single non-zero Fourier coefficient $f=\delta(\omega-\bar{\omega})$, then we have:
$$(\Mca{W}_\psi\circ\Mca{F}^{-1}f)(j,k) = \frac{1}{\sqrt{2\pi}}(\Mca{W}_\psi e^{i\bar{\omega}\cdot})(j,k), \hspace{.3cm}j,k\in\Bbb{Z},$$
that are wavelet representations of single pure sinusoids; so \textit{low frequencies} produce mainly \textit{low resolution} components in the error $e_{rec}$ (\ref{reconstruction_error}), while \textit{high frequencies} produce mainly \textit{high resolution} components in the error (figure \ref{fig:FtoW}).

\begin{figure}
 \centering
  {\includegraphics[scale=.19,clip]{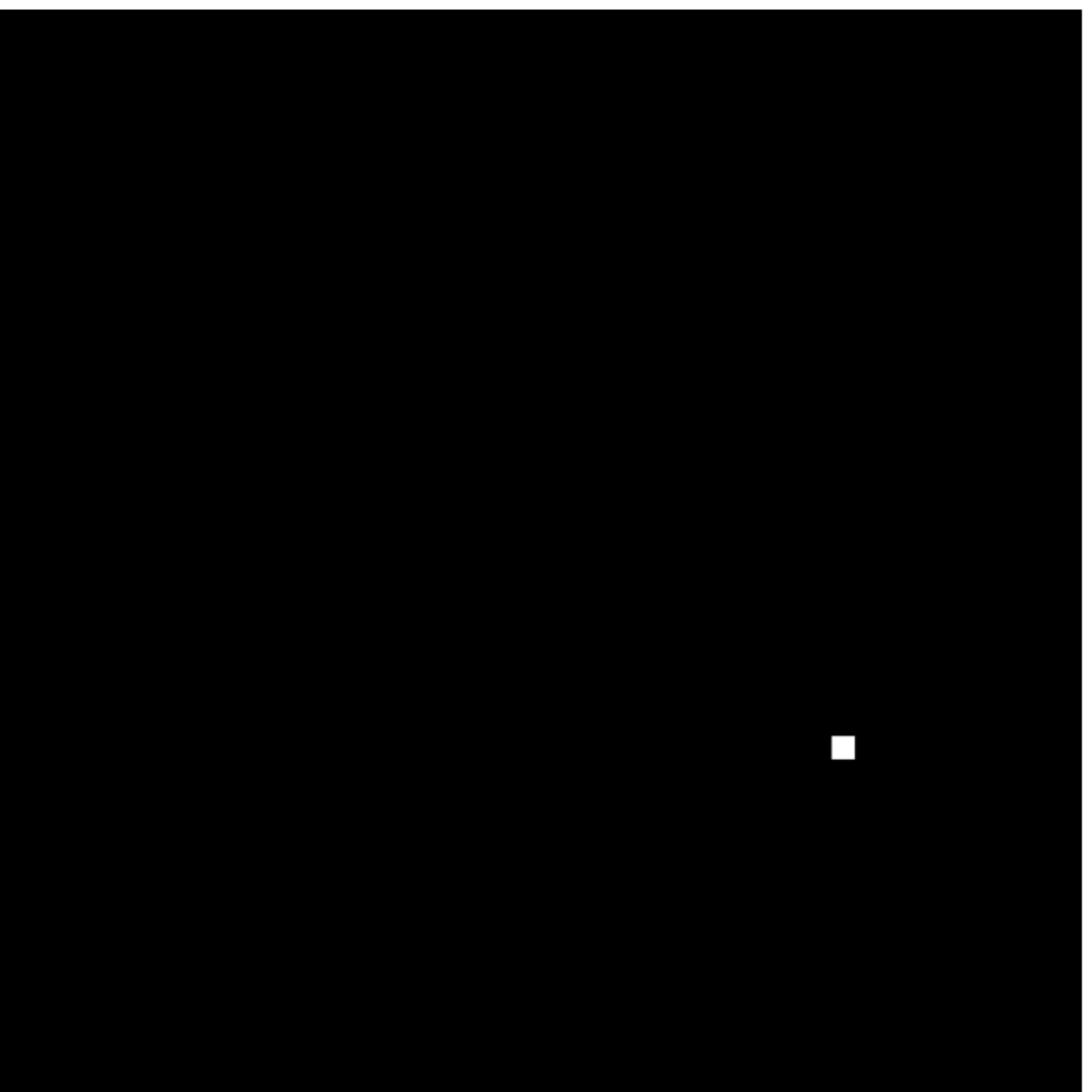}}
  {\includegraphics[scale=.19,clip]{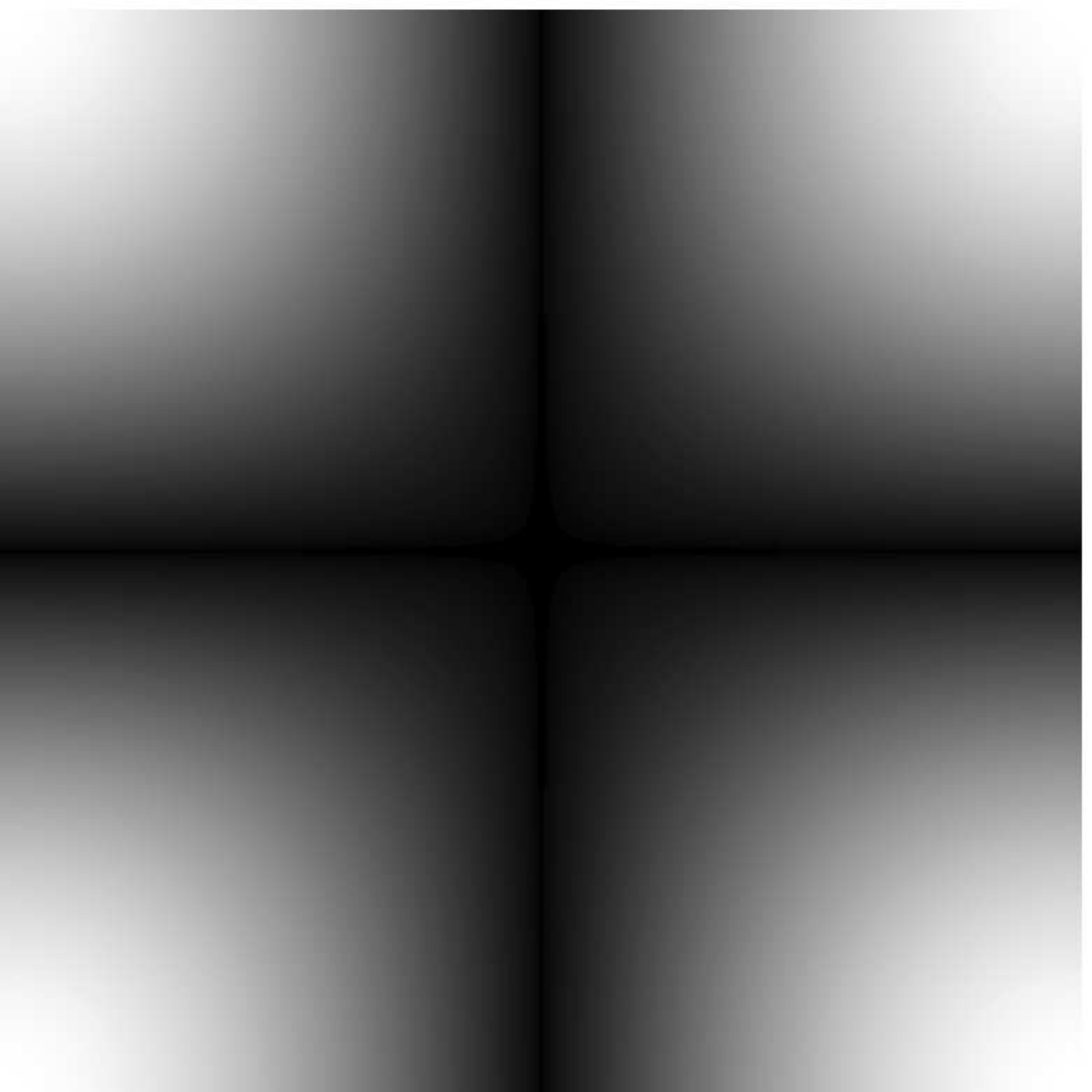}}
  {\includegraphics[scale=.19,clip]{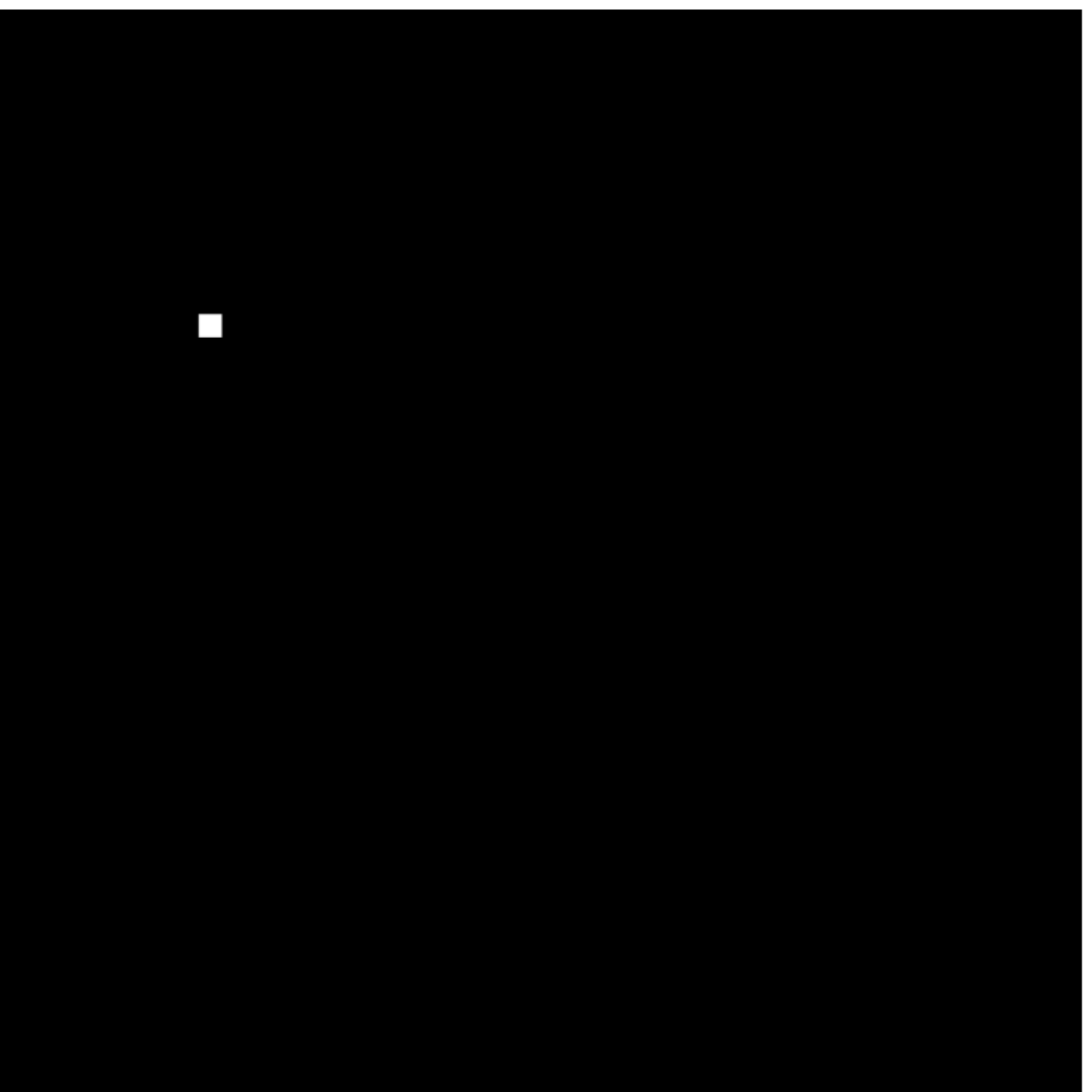}}
  {\includegraphics[scale=.19,clip]{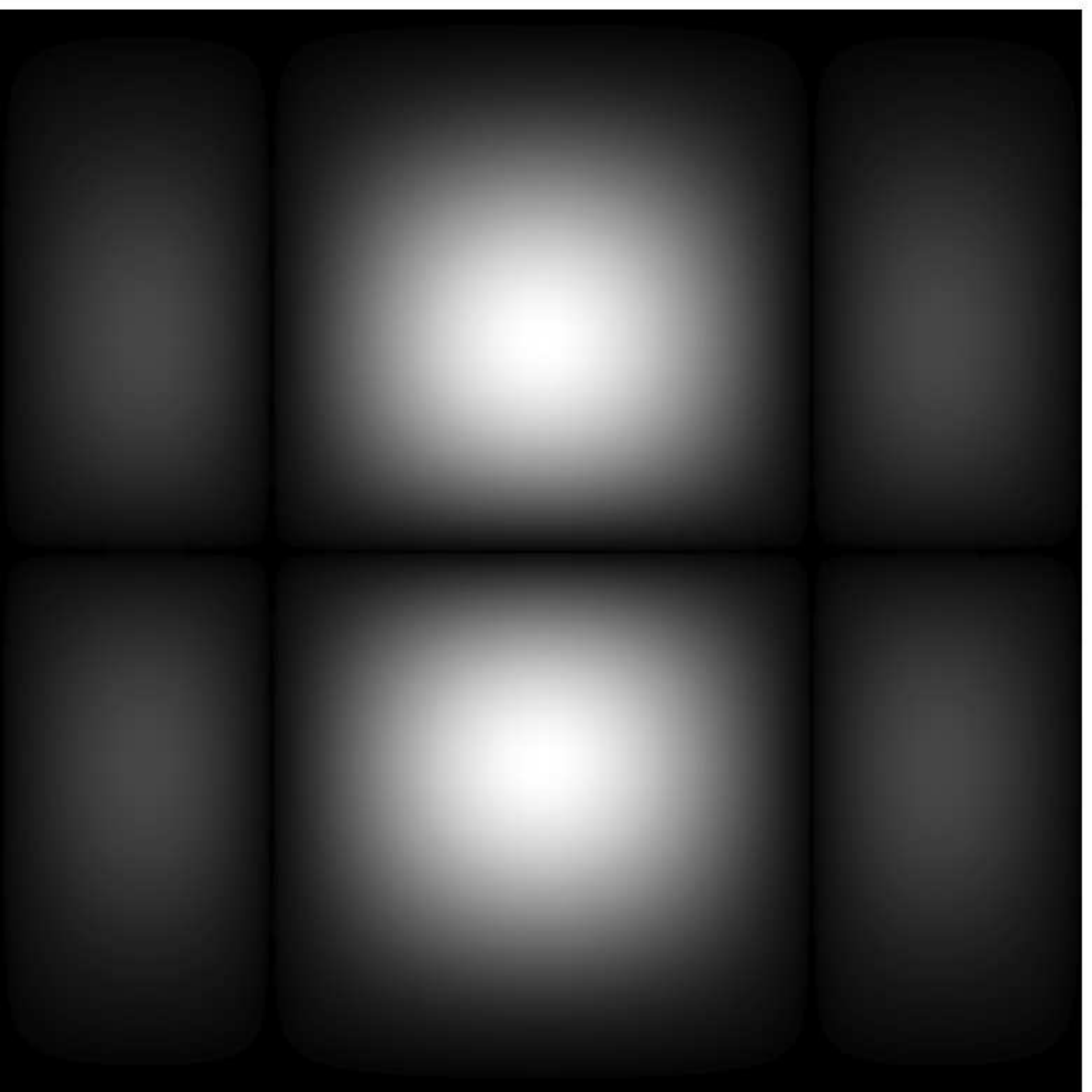}}\\
  {\includegraphics[scale=.19,clip]{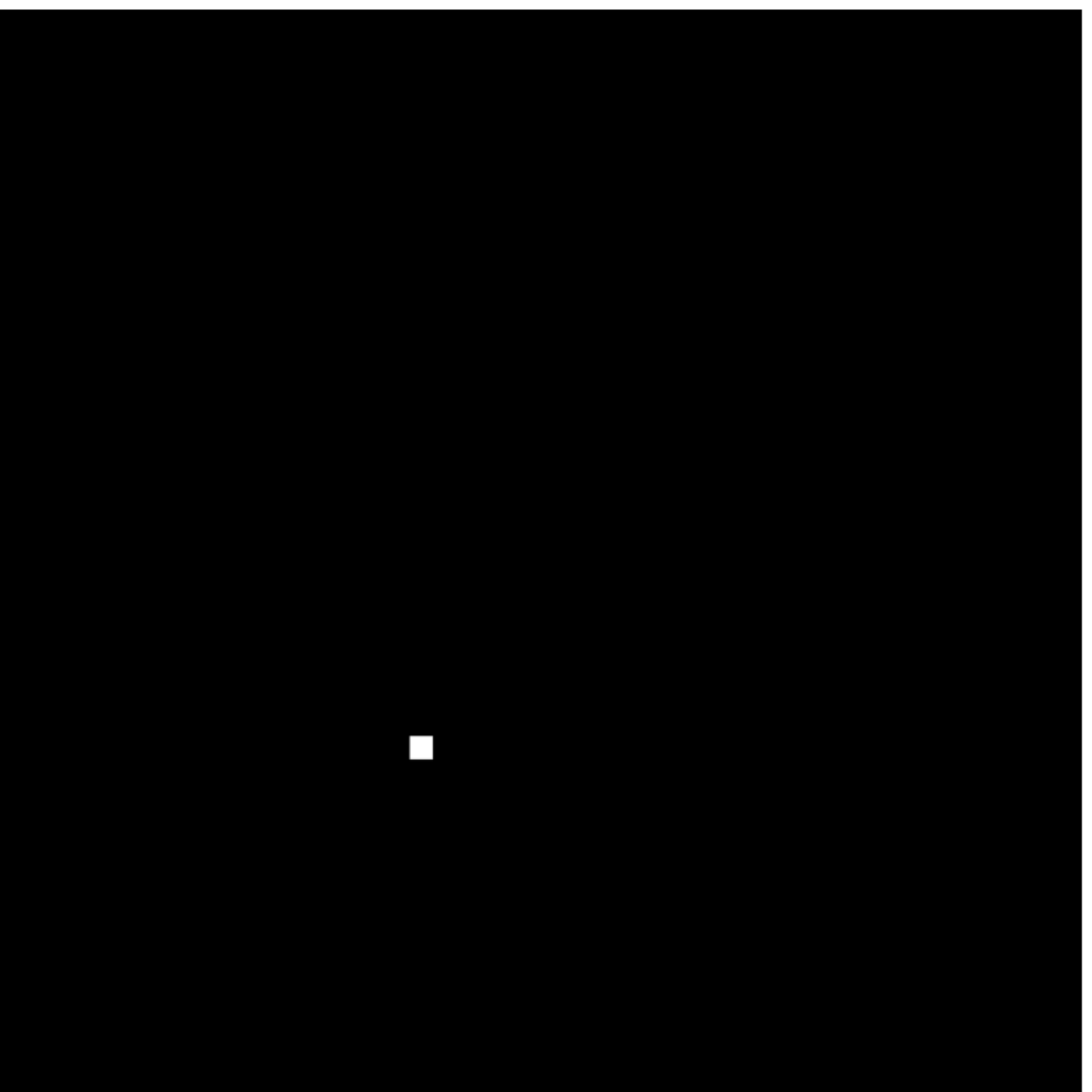}}
  {\includegraphics[scale=.19,clip]{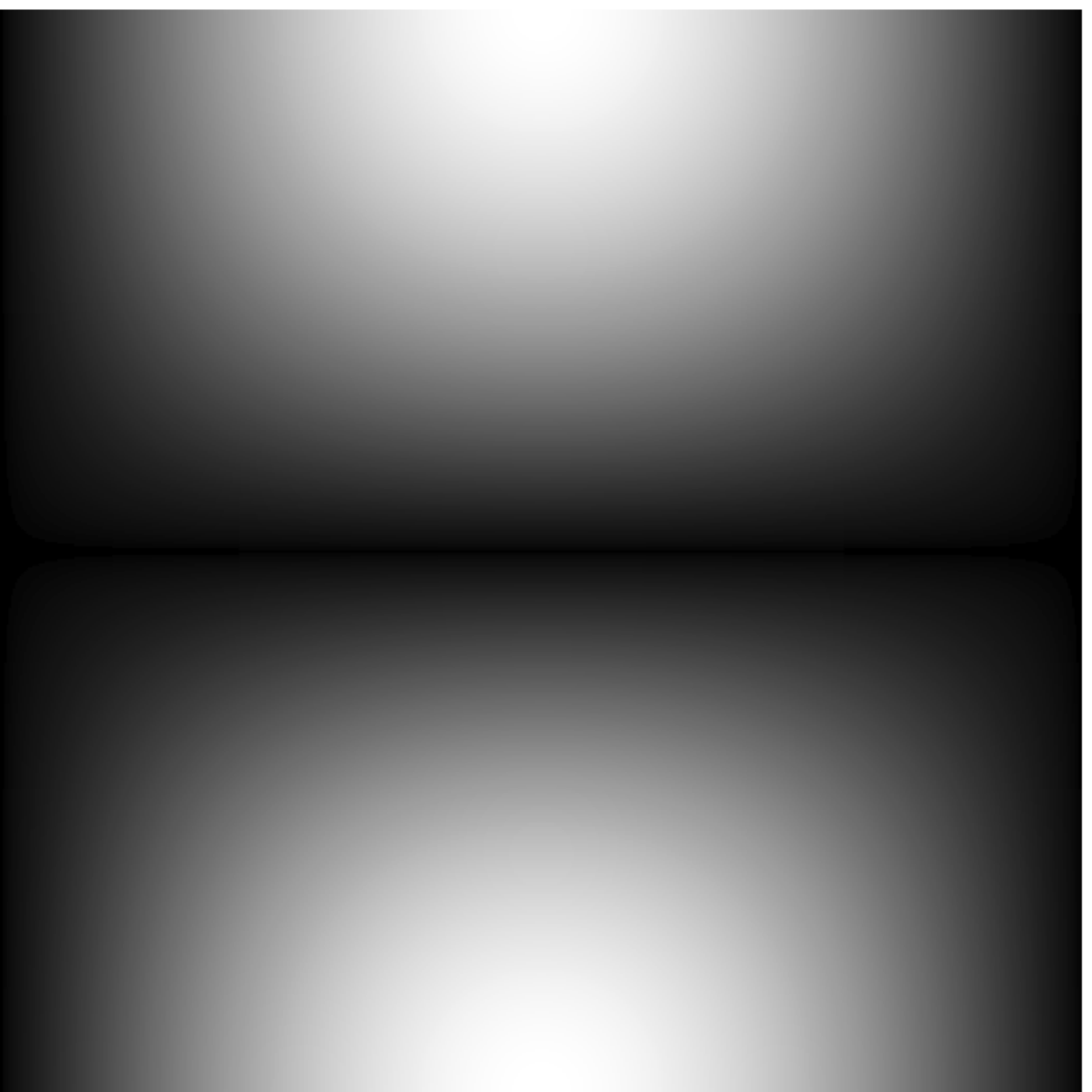}}
  {\includegraphics[scale=.19,clip]{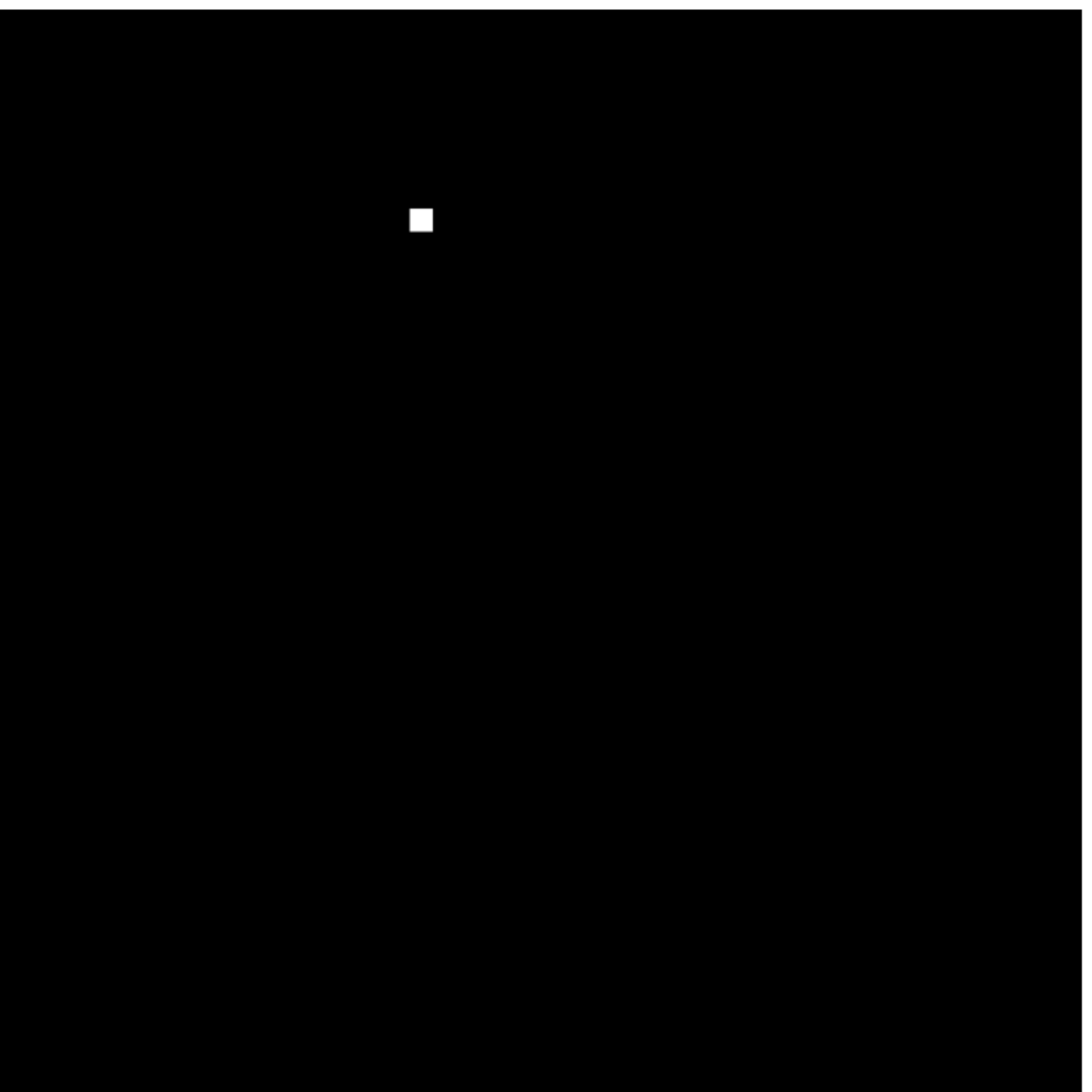}}
  {\includegraphics[scale=.19,clip]{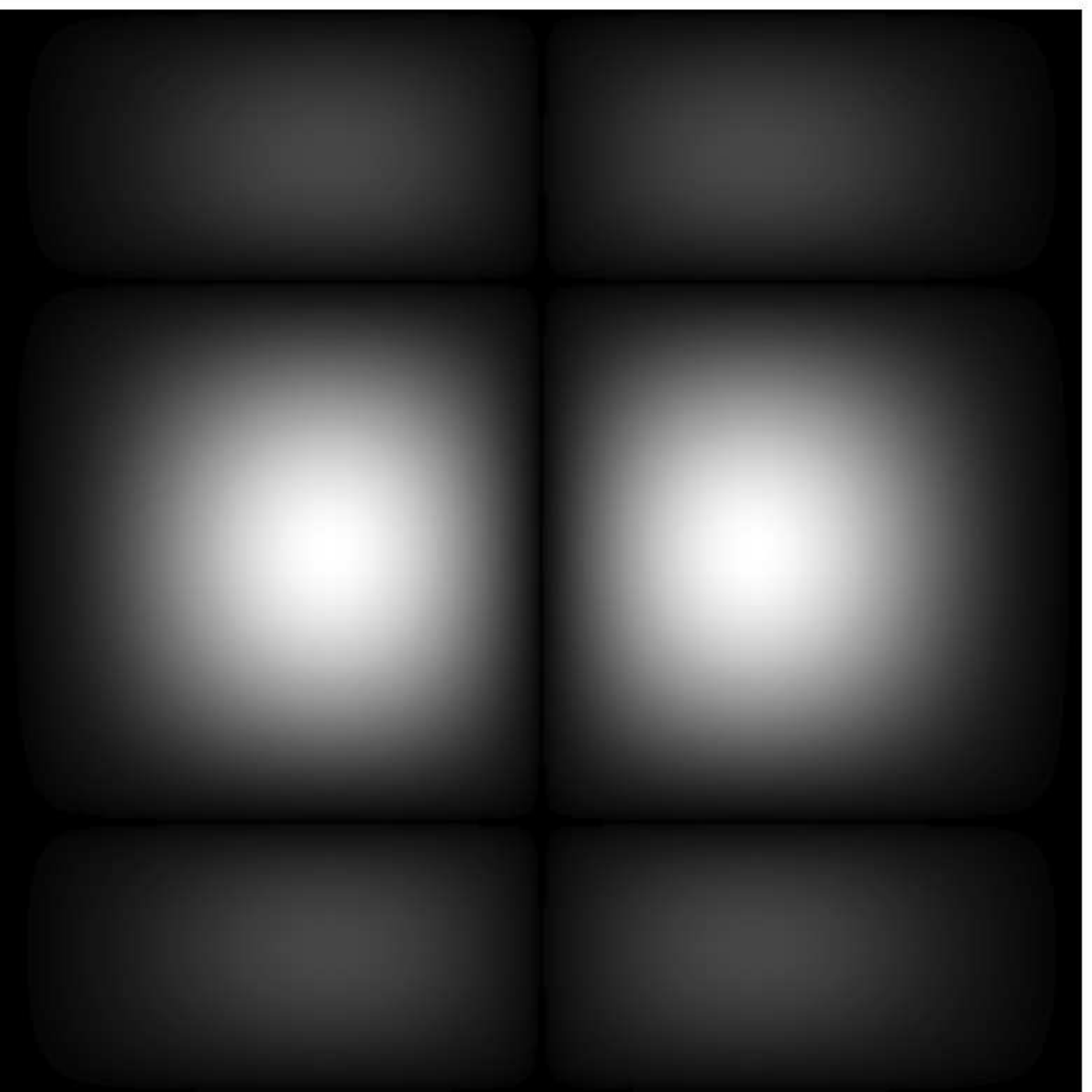}}\\
  {\includegraphics[scale=.19,clip]{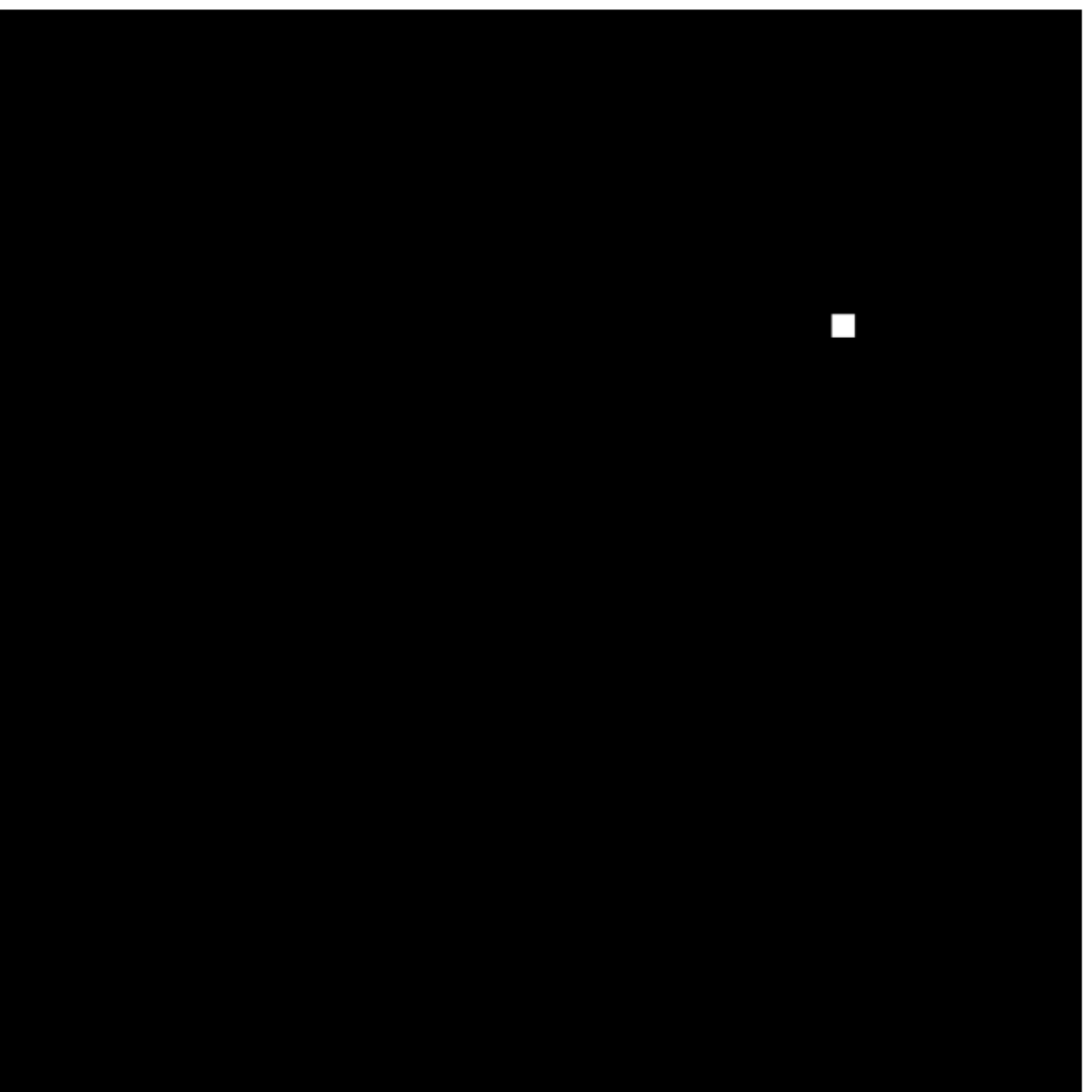}}
  {\includegraphics[scale=.19,clip]{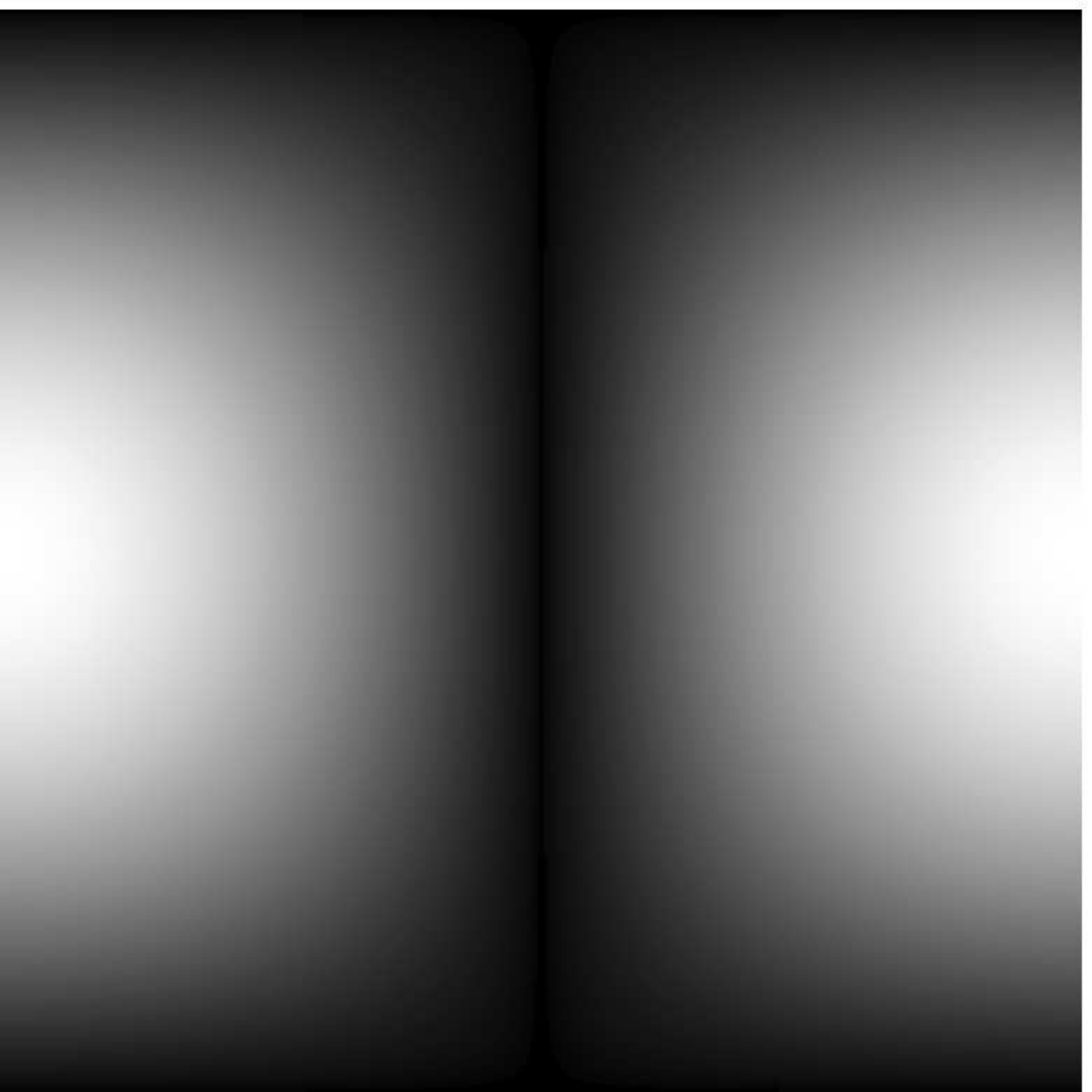}}
  {\includegraphics[scale=.19,clip]{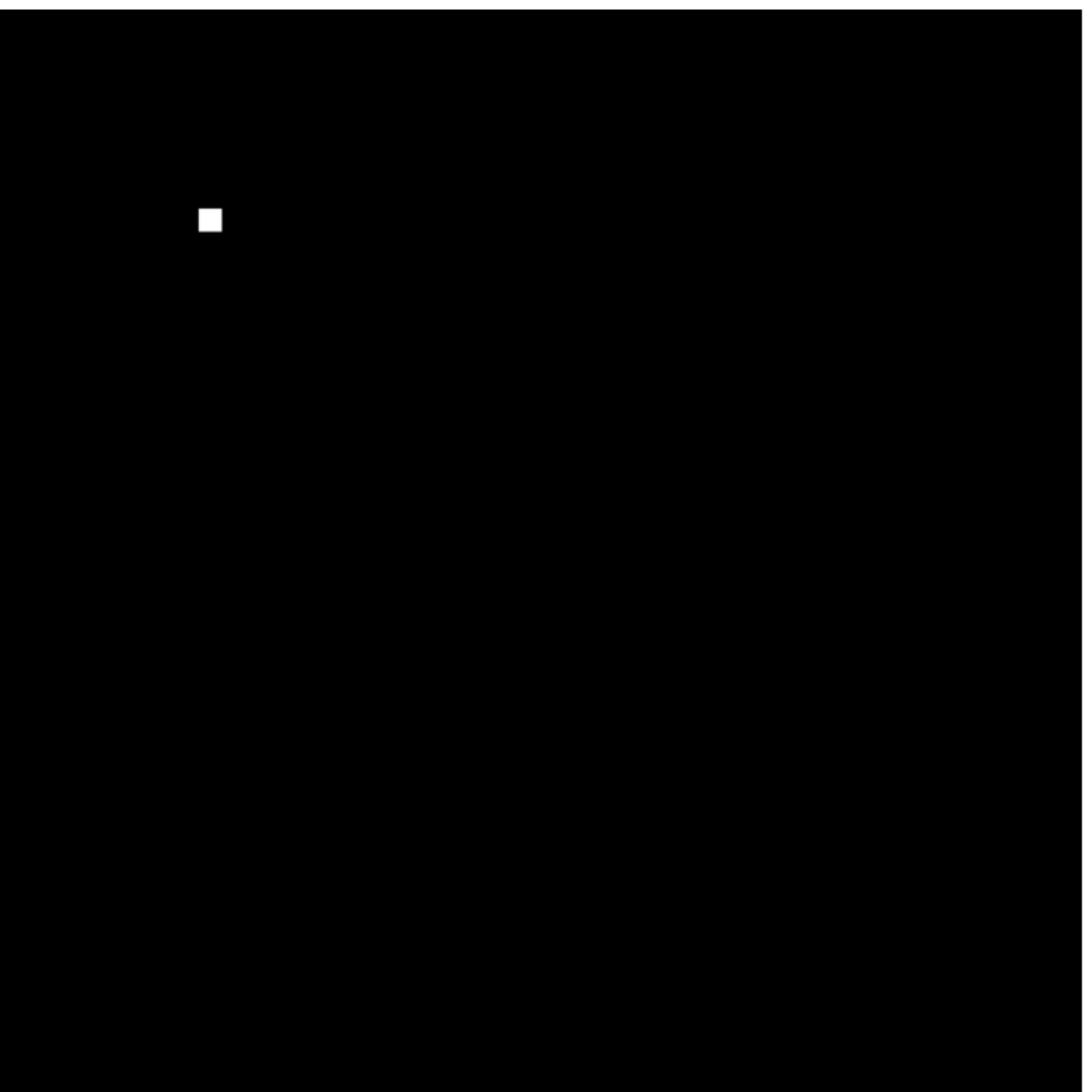}}
  {\includegraphics[scale=.19,clip]{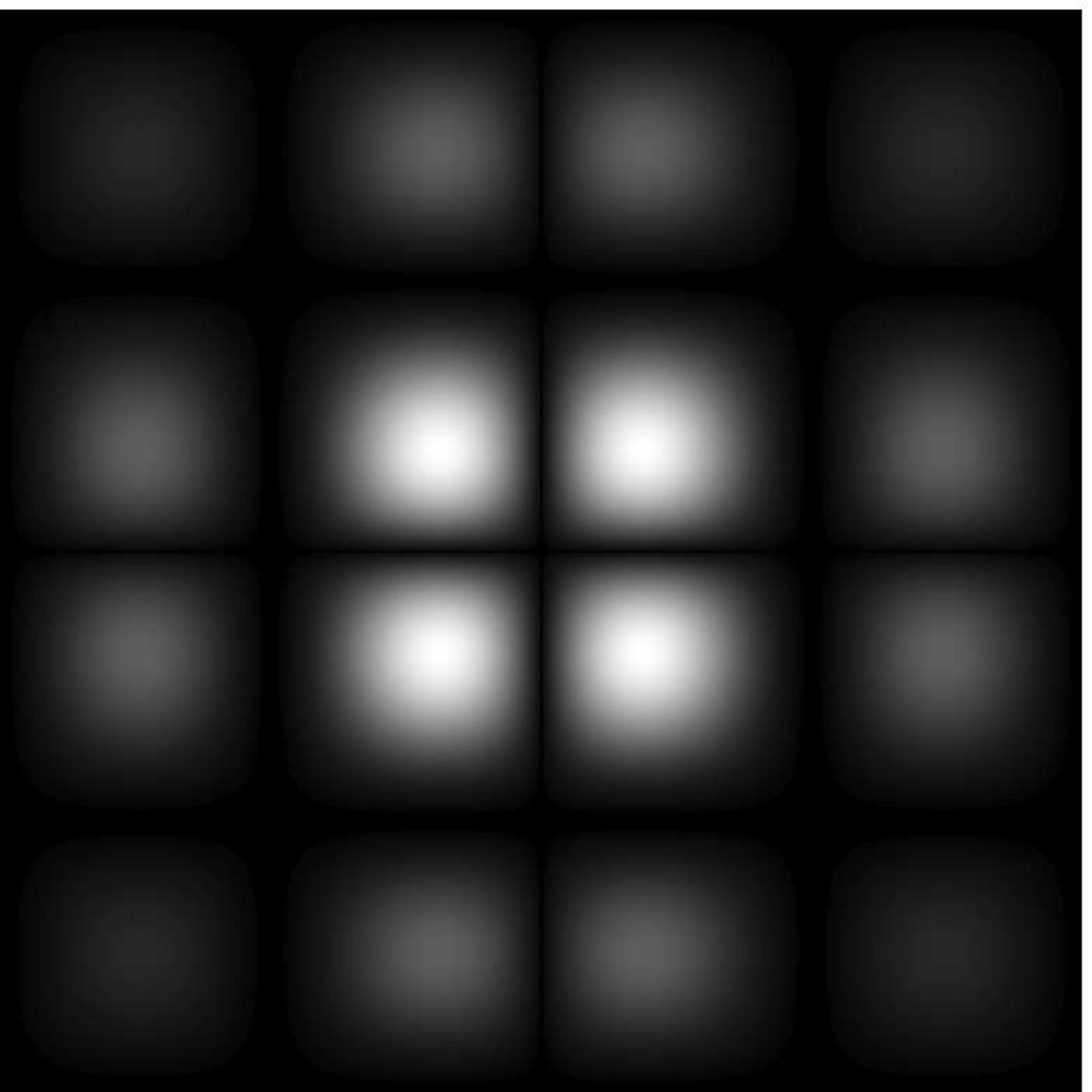}}\\
  {\includegraphics[scale=.19,clip]{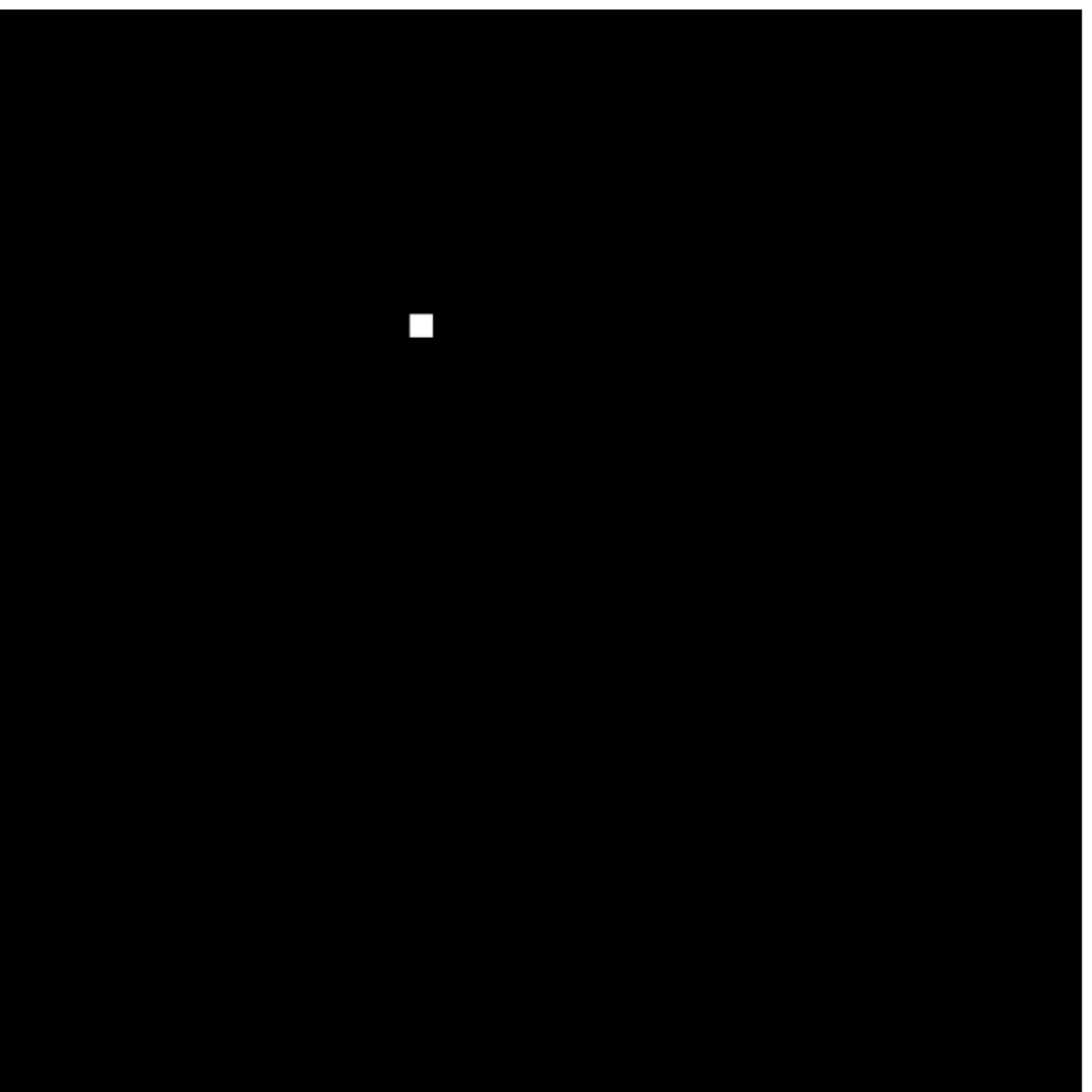}}
  {\includegraphics[scale=.19,clip]{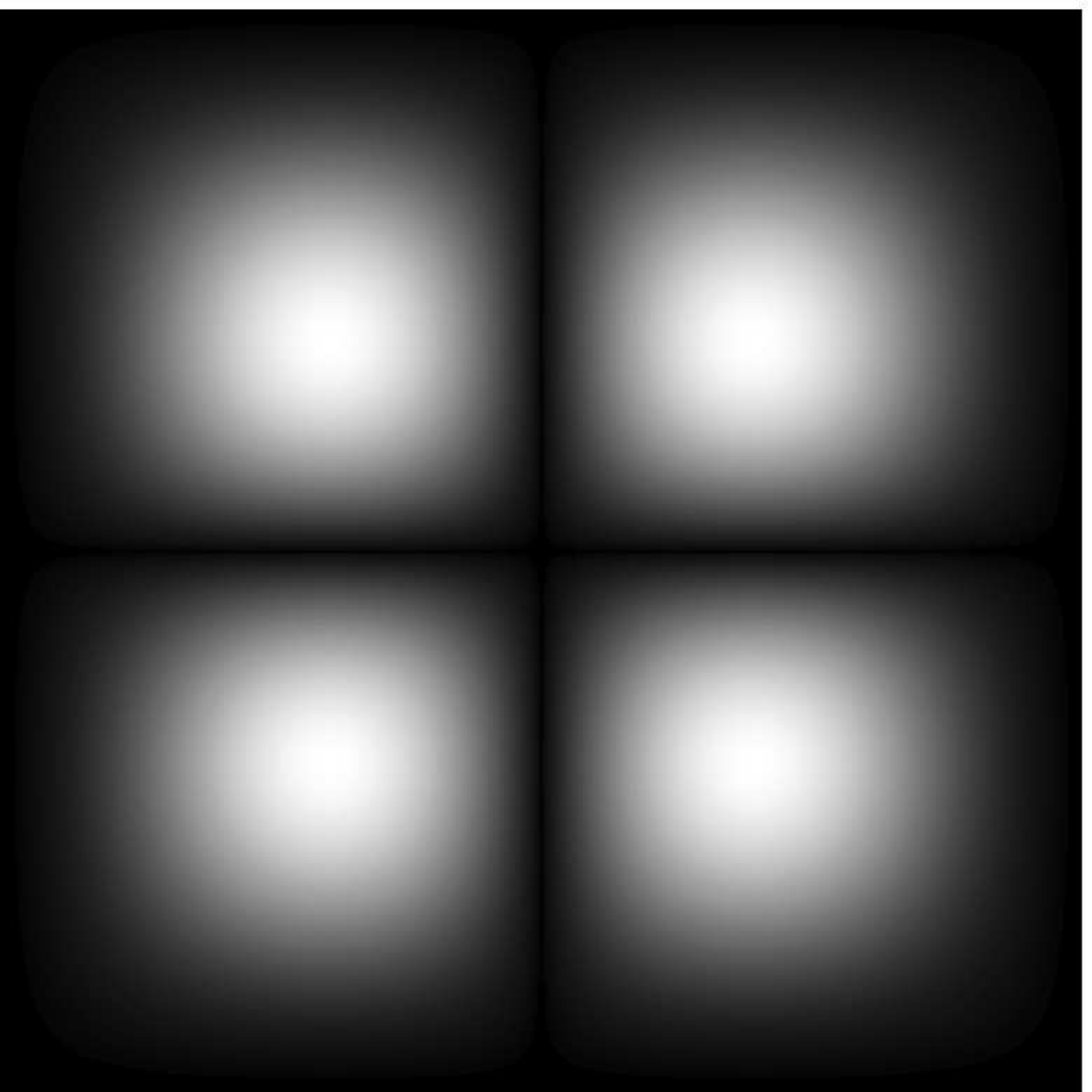}}
  {\includegraphics[scale=.19,clip]{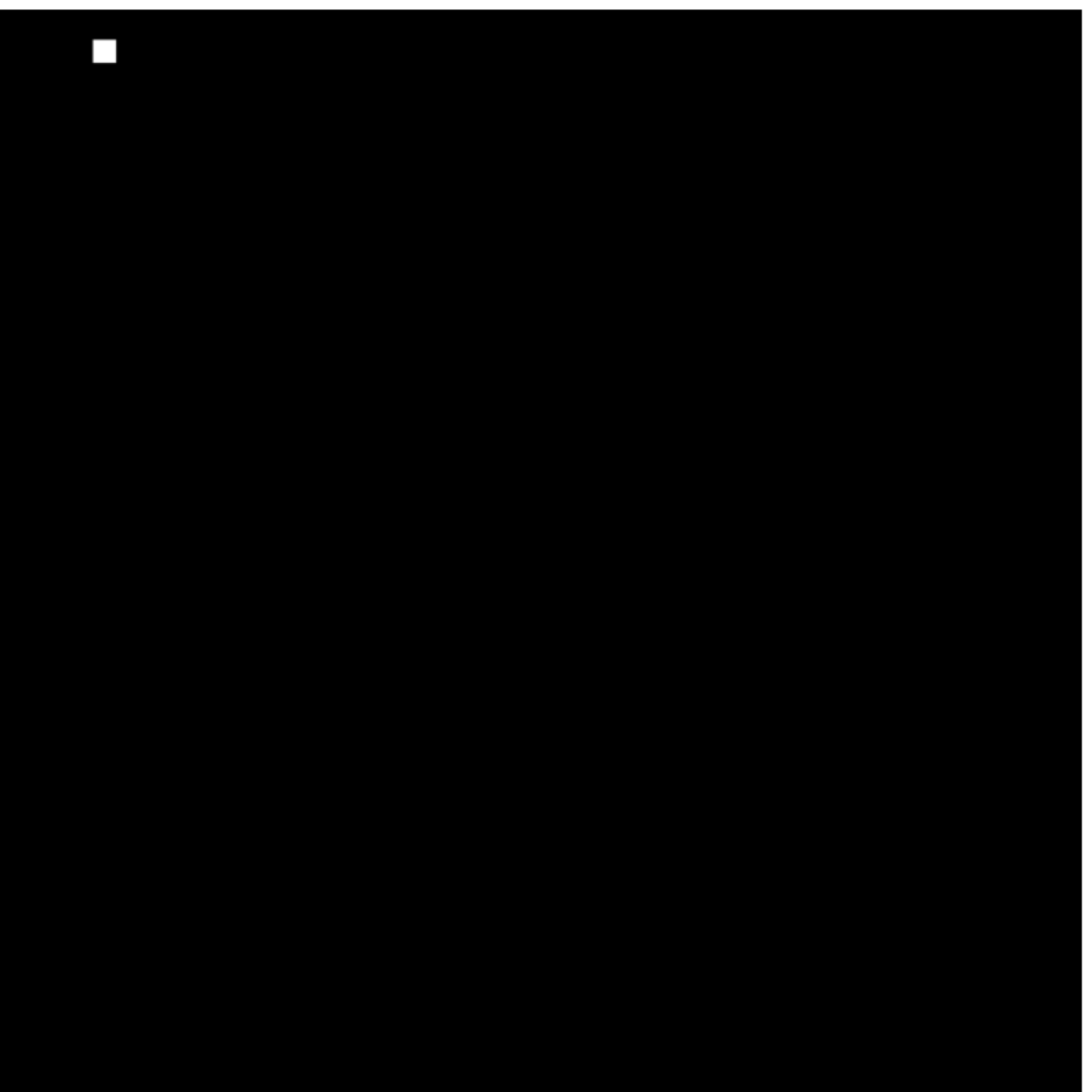}}
  {\includegraphics[scale=.19,clip]{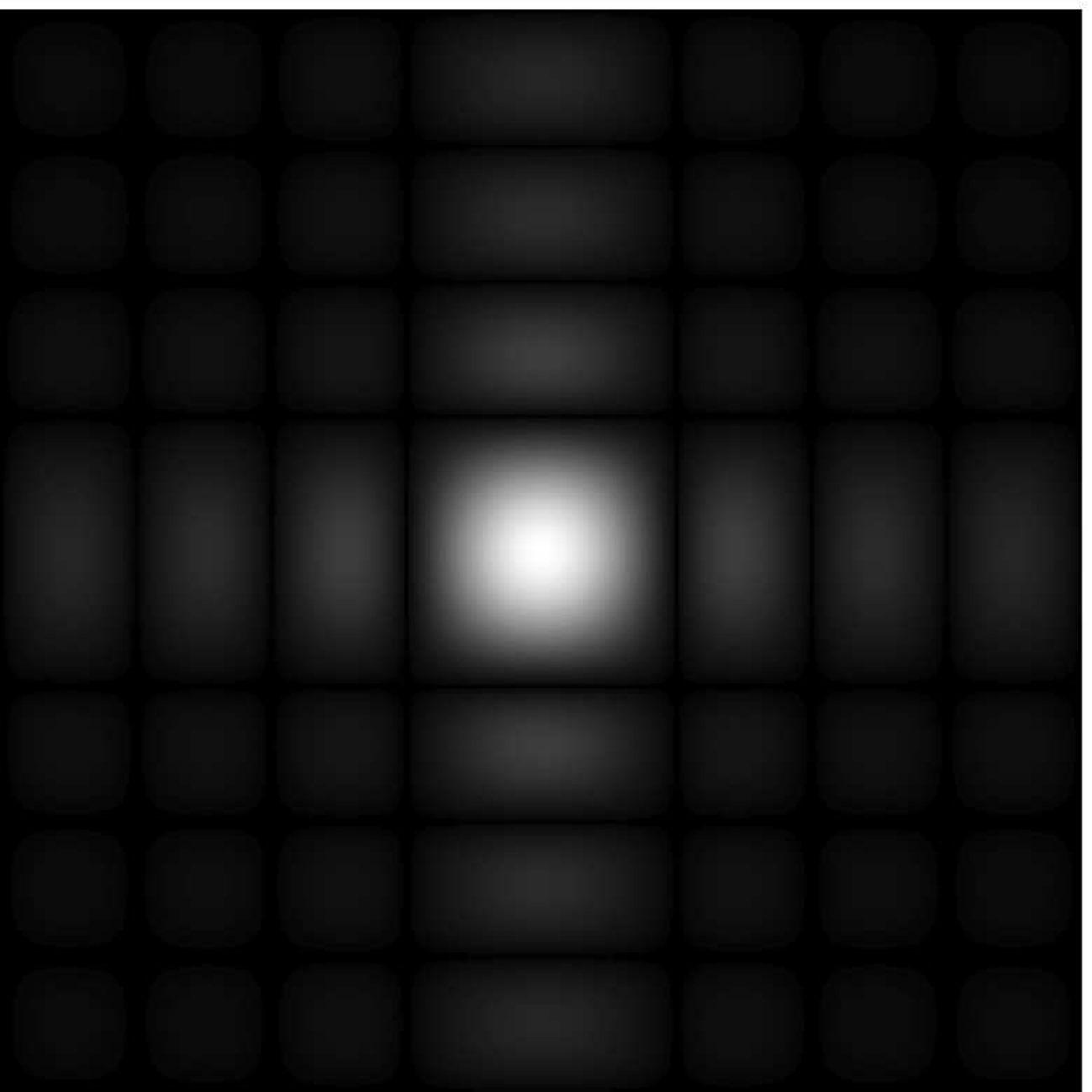}}
  \caption{Examples of two-dimensional Haar single coefficients at different resolutions and their frequency spectra (in modulus): note that lower resolution wavelet coefficients excite lower frequencies.}
\label{fig:WtoF}
\end{figure}

Vice versa, if $y=\delta_{j,\bar{j}}\cdot\delta_{k,\bar{k}}$ then:
$$(\Mca{F}\circ\Mca{W}_\psi^{-1}y)(\omega) = \hat{\psi}_{\bar{j},\bar{k}}(\omega) = 2^{-\bar{j}/2} e^{-i2^{1-\bar{j}}\pi\omega \bar{k}} \hat{\psi}(2^{-\bar{j}}\omega),$$
so \textit{high resolutions} have a frequency spectrum scattered on \textit{high frequencies} and \textit{low resolutions} have a frequency spectrum peaked in \textit{low frequencies}. Moreover, fixed a resolution $j$ but varying the position $k$, the frequency spectrum is the same in modulus (figure \ref{fig:WtoF}).

\begin{figure}
 \centering
  {\includegraphics[scale=.28,clip]{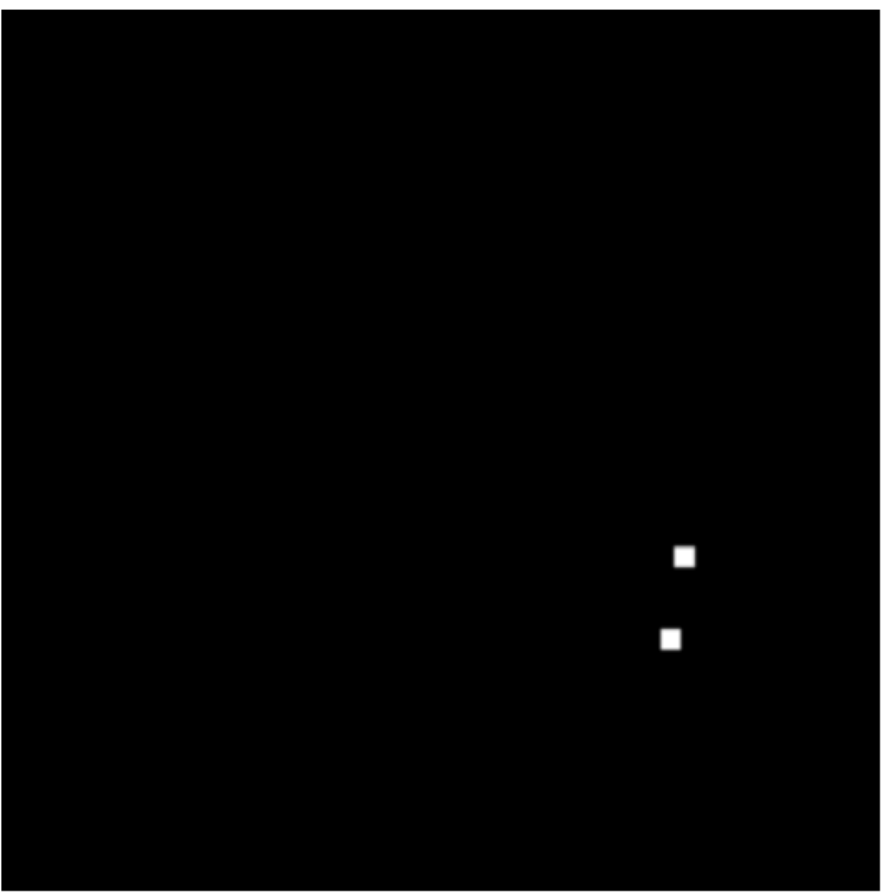}}
  {\includegraphics[scale=.28,clip]{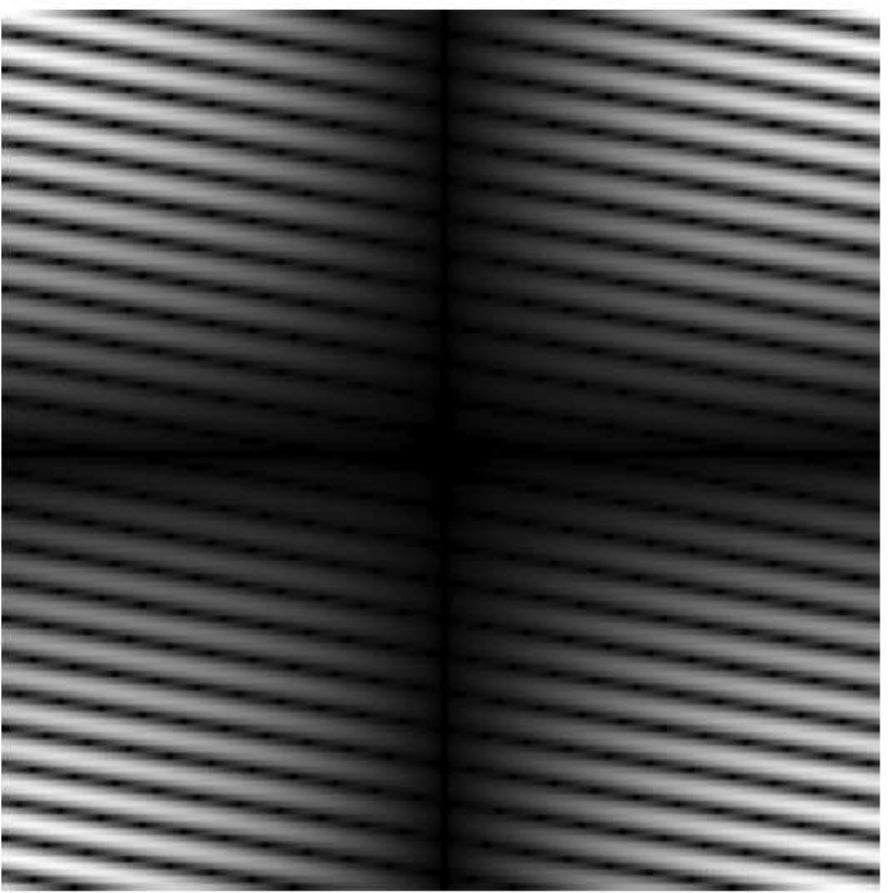}}
  {\includegraphics[scale=.28,clip]{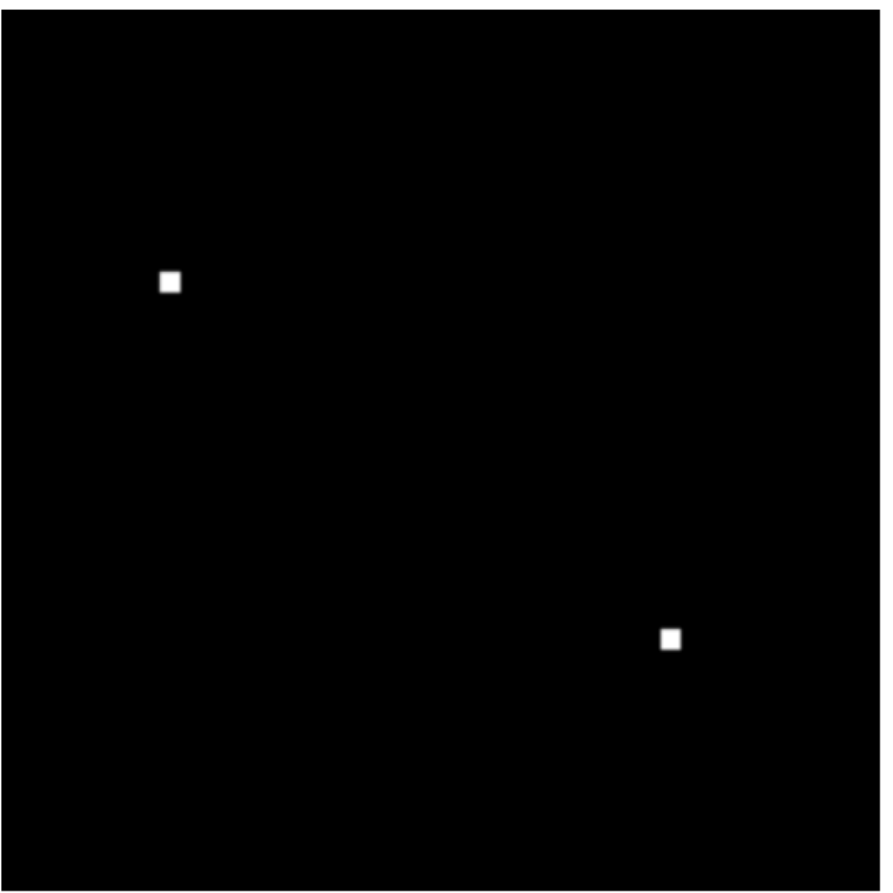}}
  {\includegraphics[scale=.28,clip]{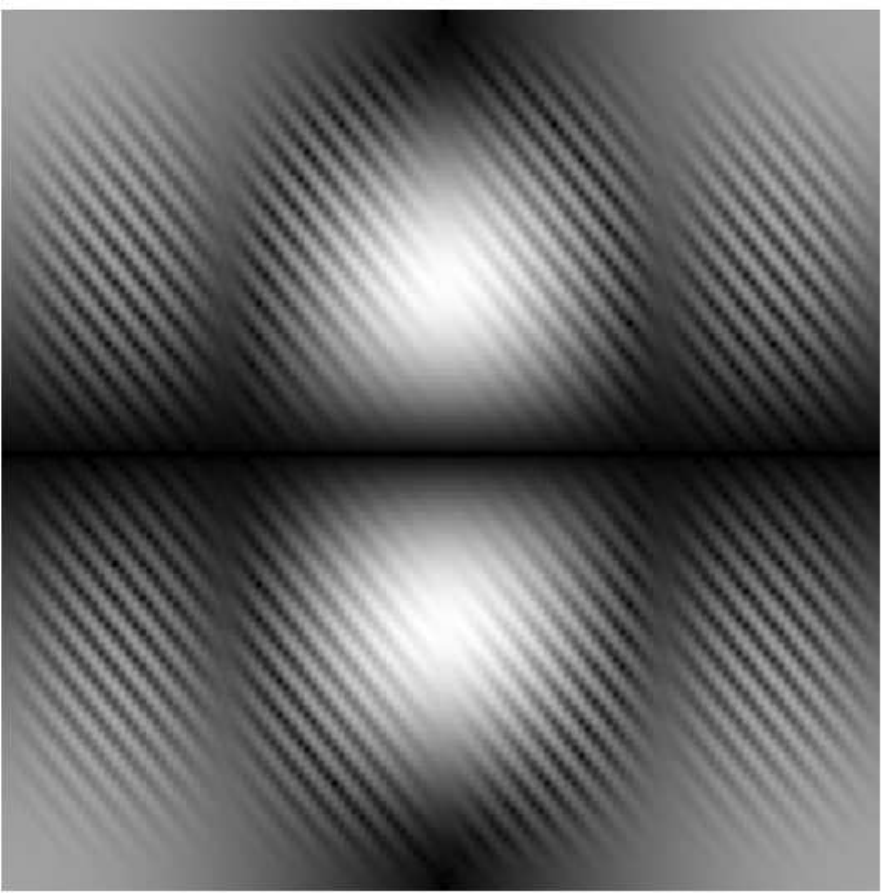}}\\
  {\includegraphics[scale=.28,clip]{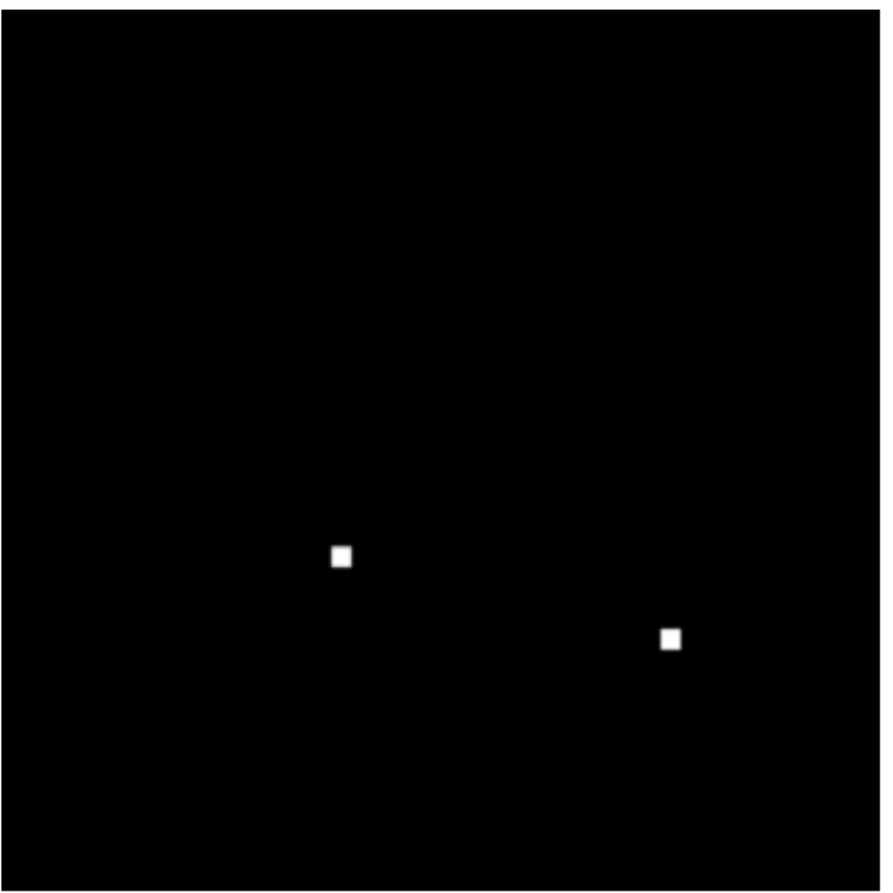}}
  {\includegraphics[scale=.28,clip]{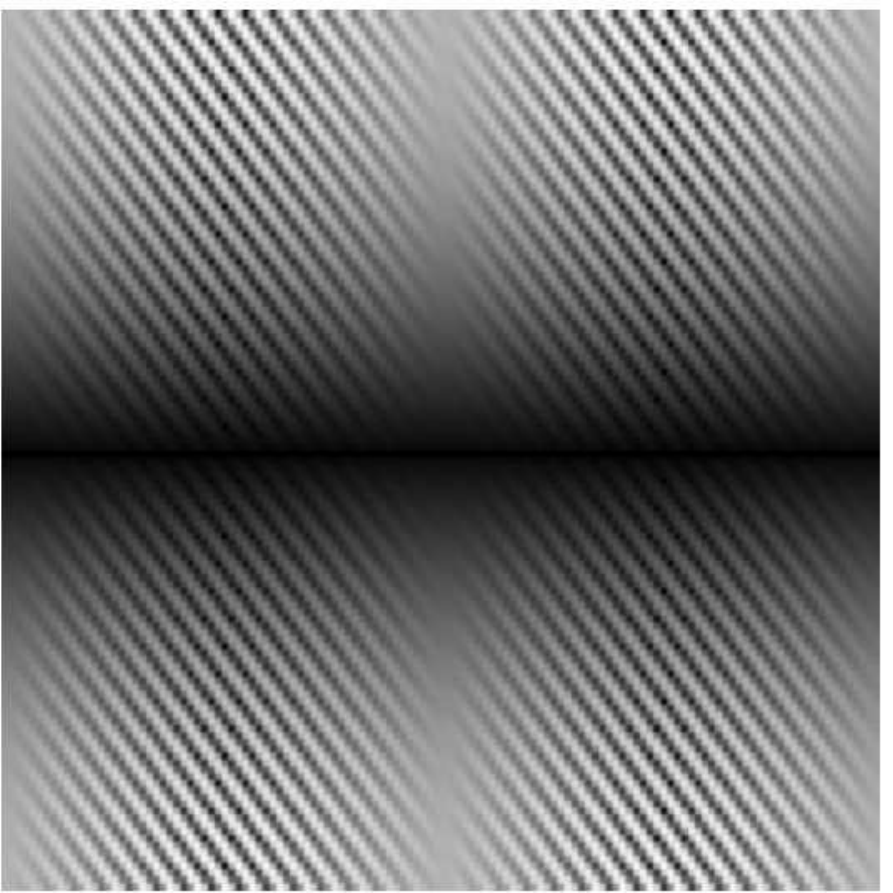}}
  {\includegraphics[scale=.28,clip]{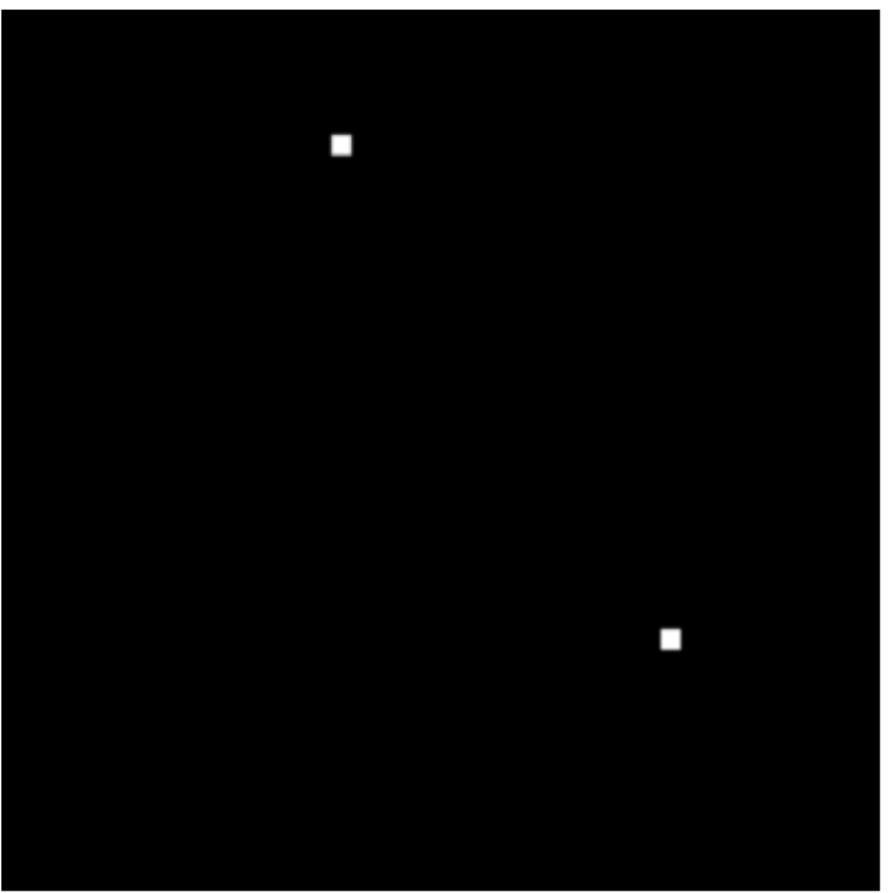}}
  {\includegraphics[scale=.28,clip]{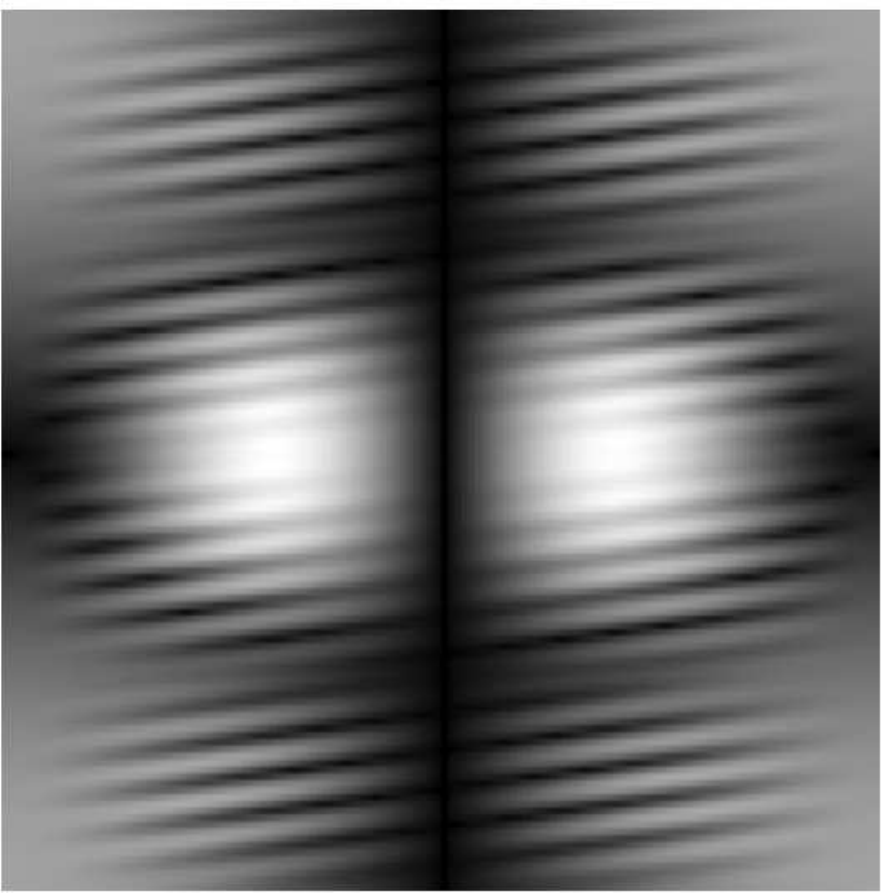}}\\
  {\includegraphics[scale=.28,clip]{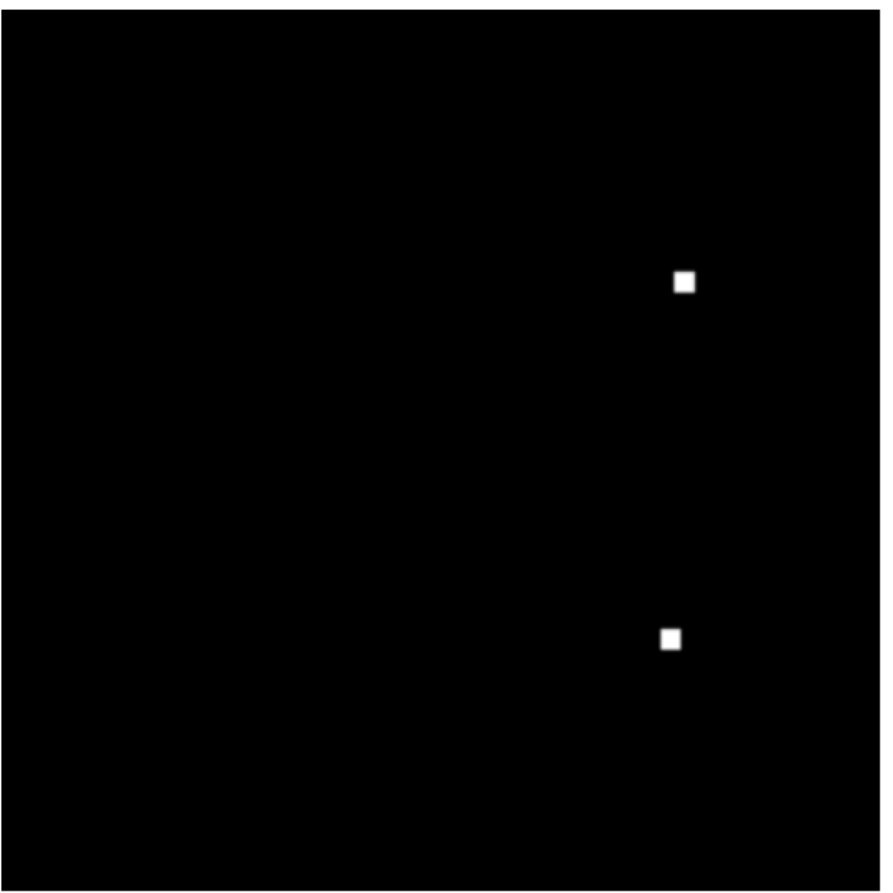}}
  {\includegraphics[scale=.28,clip]{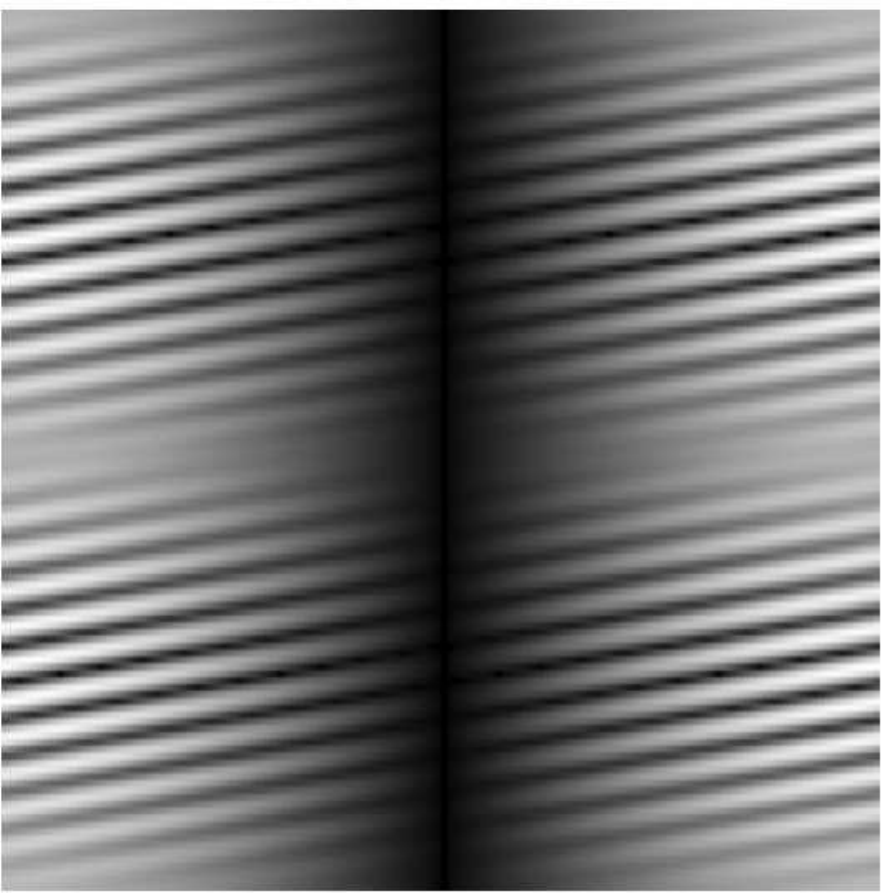}}
  {\includegraphics[scale=.28,clip]{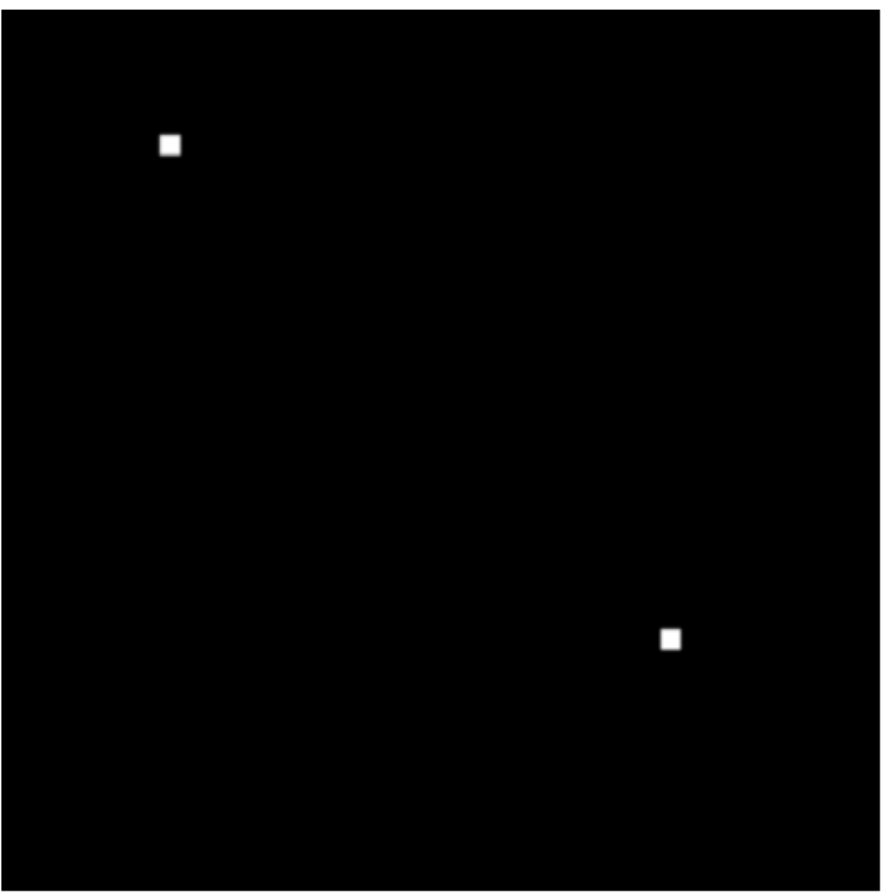}}
  {\includegraphics[scale=.28,clip]{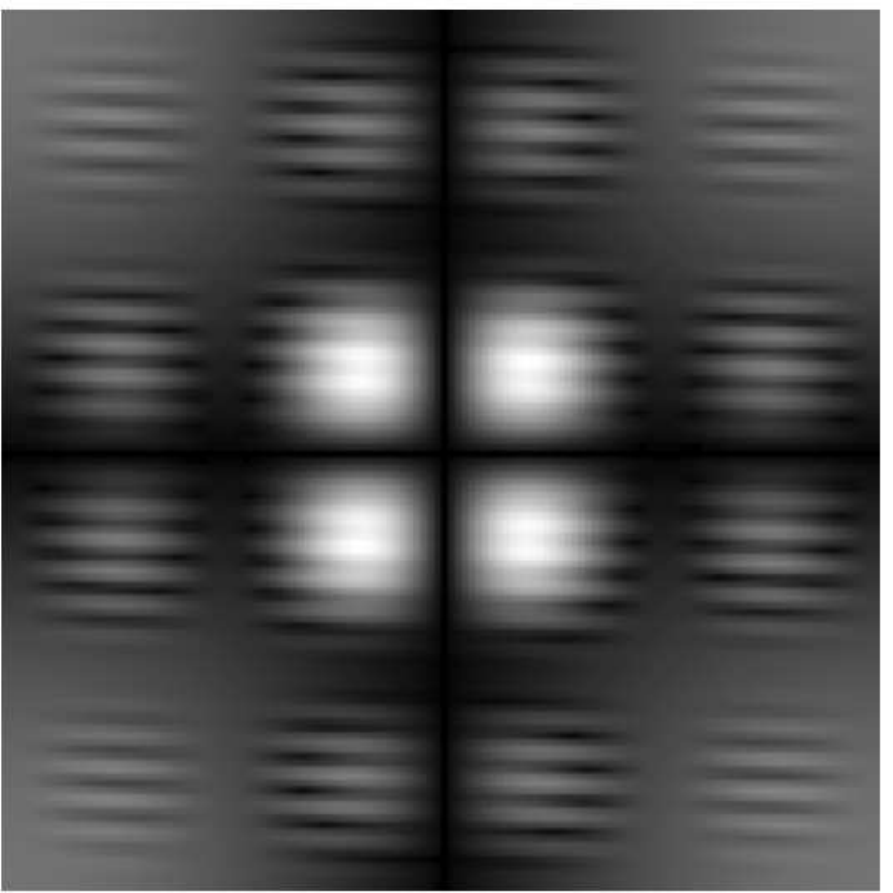}}\\
  {\includegraphics[scale=.28,clip]{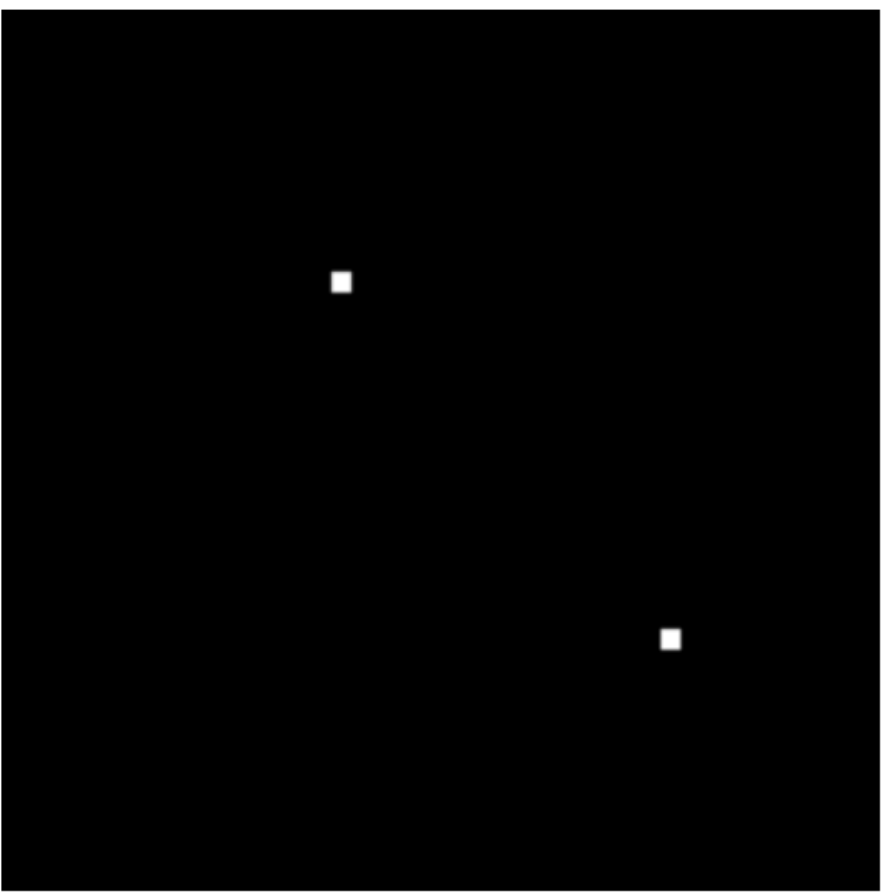}}
  {\includegraphics[scale=.28,clip]{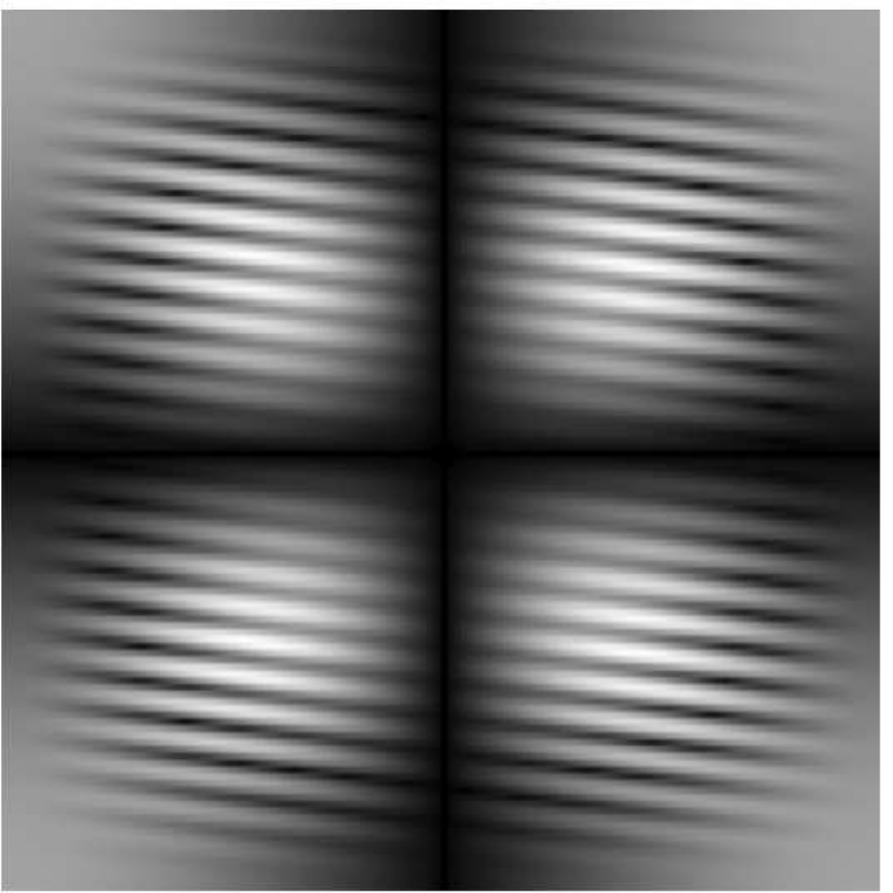}}
  {\includegraphics[scale=.28,clip]{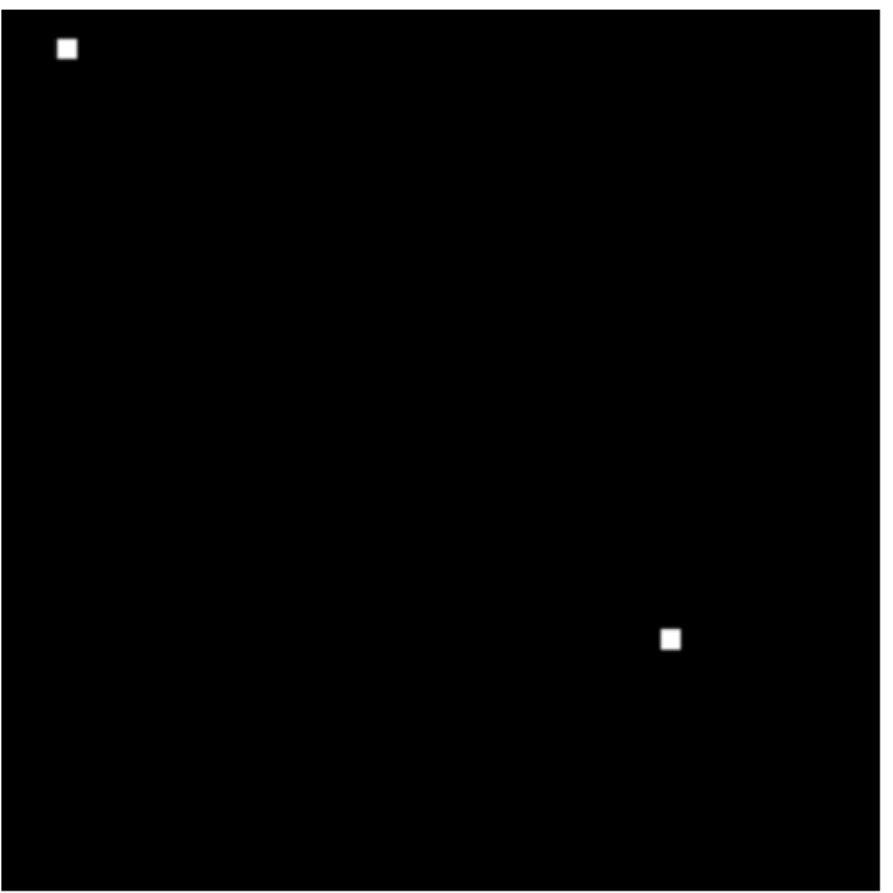}}
  {\includegraphics[scale=.28,clip]{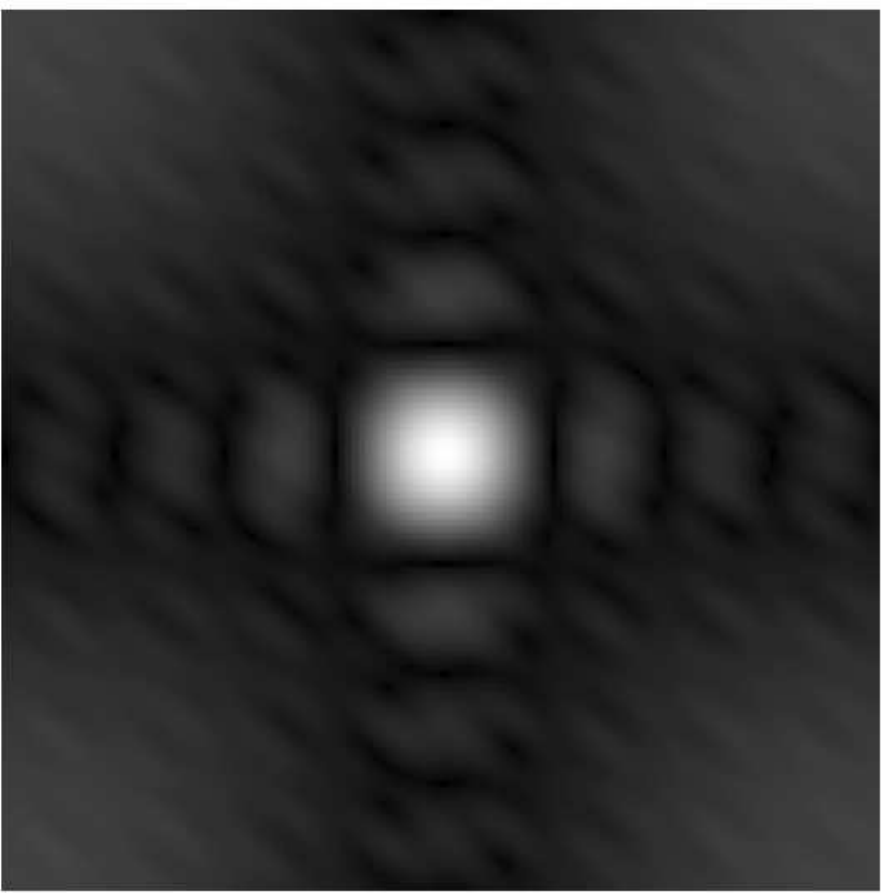}}
  \caption{Examples of pairs of two-dimensional Haar coefficients at different resolutions and their frequency spectra (in modulus): note how the two frequency spectra interfere each other.}
\label{fig:interf}
\end{figure}

When we transform more then one non-zero wavelet coefficients, the modulus of the frequency spectrum shows the typical phenomenon of wave interference (figure \ref{fig:interf}): if $y=\delta_{j,j_1}\cdot\delta_{k,k_1}+\delta_{j,j_2}\cdot\delta_{k,k_2}$, then
$$\begin{array}{ccl}
(\Mca{F}\circ\Mca{W}_\psi^{-1}y)(\omega) &=& e^{-i2^{1-j_1}\pi\omega k_1} [2^{-j_1/2} \hat{\psi}(2^{-j_1}\omega)+\\
& &+ 2^{-j_2/2} \hat{\psi}(2^{-j_2}\omega) e^{-i2\pi\omega(2^{-j_2}k_2-2^{-j_1}k_1)}].\end{array}$$

\section{Heuristic algorithms for determining subsets of measurements}
\label{algorithms}

We present four algorithms that determine good subsets $J$ of measurements (undersampling). They all implement some possible heuristics, since the optimal solution cannot be computed analytically.

The algorithms receive as input some informations about the spatial vector $x$ to recover, at least the indexes of its non-zero wavelet coefficients $I$, and at most also its values. As output the algorithms compute a set $J$ of frequency indexes to be sampled among the whole set $\{1,...,N\}$.

In a DSC-MRI exam, the time-series frames $x^{(t)}$ to acquire , $t=1,...,T$, are strongly spatio-temporal correlated. So we give as input to the algorithm the data relative to the image $\bar{x} = \mean_{t=1,...,{\tau}}{x^{(t)}}$ calculated as the mean of the first $\tau\ll T$ frames recovered in a complete way (i.e. with as much as possible frequencies measured). The set $J$ determined are then used to sample every other frame to acquire.

We also considered a time-variable mask in order to evolve the set $J$ in accordance with the evolution of the physical process. 
In particular, we update $\bar{x}^{(k)}$ as follows:
\begin{equation}
\bar{x}^{(0)}=\bar{x}, \quad \bar{x}^{(k+1)}=a\bar{x}^{(k)}+(1-a)\tilde{x}
\label{aggiornamento_mask}
\end{equation}where $\tilde{x}$ is the last reconstructed frame.
The results in section \ref{comparison} show that this is meaningful in the DSC-MRI context.

Moreover, the non-measured frequencies of a frame (elements of $R=\{1,...,N\}\setminus J$), instead of being set to zero, are set to their value in the spectrum of the image mean of the first frames.

\subsection{Algorithm 1}

Let us calculate the $m\in\Bbb N$ frequencies with maximum modulus of the mean recovered image $\bar{x}\in \Bbb{R}^N$ (a priori $F\bar{x} \in \Bbb{C}^N$):\\
\begin{algorithm}[H]
 \KwData{$\bar{x}\in \Bbb{R}^N$}
 $f = F\bar{x}$\;
 $J = \emptyset$\;
 \For{$i = 1 \to m$}{
   $h = \arg\max|f|$\;
   $J \leftarrow h$\;
   $f_h = 0$\;}
 \label{algo1}
\end{algorithm}

\begin{figure}
 \centering
  {\includegraphics[scale=.2,clip]{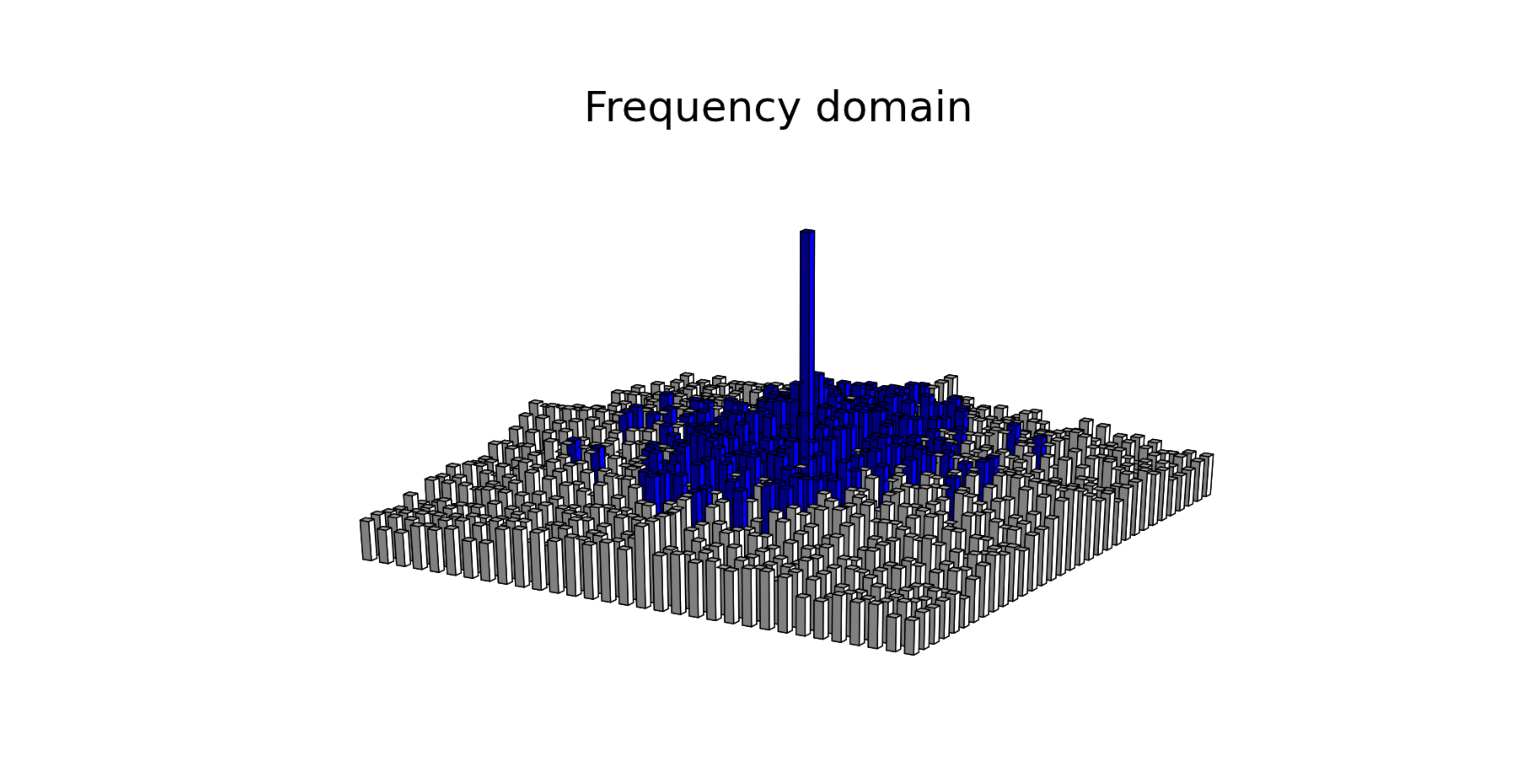}}
  \caption{Only the frequencies with the biggest modulus are selected to be measured (added to $J$).}
\label{fig:algo1}
\end{figure}

This algorithm computes the set $J$ of frequencies to measure by thresholding the ones with maximum modulus (figure \ref{fig:algo1}).

Such an under-sampling $J$ obtains an approximation $\hat{y} = WF^T \left( \begin{array}{c}
f_J \\
0 \end{array} \right)$ for $y=W\bar{x}$ such that the errors $|f_k| WF^T_{\cdot,k}$ for every \textit{single} excluded frequency $k \in R$ are minimal, since the frequencies $f_k$ set to zero are the ones with minimum modulus. This is true for excluding single frequencies but does not hold necessarily when considering the exclusion of groups of frequencies, so the error could not be globally optimal, but it is \textit{locally} optimal, in the sense that given $\mu+1$ non-zero frequencies, if you want measure $\mu$ of them (excluding just one of them), the frequency to be discarded is precisely the one with the minimum modulus.

\subsection{Algorithm 2}

Suppose to know the distribution of the wavelet non-zero coefficients of $y=W\bar{x}$ among the different resolution orders, then we can calculate the frequencies that are expected to have the maximum modulus and add them to the set of measurements $J$.

Let us exploit the analysis of the transforms $\Mca{W}$ and $\Mca{F}$. We found the following formula for the frequency spectrum of a single wavelet coefficient:
$$(\Mca{F}\circ\Mca{W}^{-1}y)(\omega) = 2^{-\bar{j}/2} e^{-i2^{1-\bar{j}}\pi\omega \bar{k}} \hat{\psi}(2^{-\bar{j}}\omega), \hspace{.2cm} \textnormal{if}\hspace{.2cm} y=\delta_{j,\bar{j}}\cdot\delta_{k,\bar{k}}.$$

So it is clear that wavelet coefficients of the same resolution order $\bar{j}$ have the same frequency spectrum in modulus. The same formula highlights that varying the order of resolution the frequency spectrum is subjected to dilatation and contraction effects: the maximum modulus frequency of a frequency spectrum of a wavelet coefficient of order $j$ has modulus proportional to $2^{-j/2}$, while the support of the frequency spectrum dilates with the factor $2^j$. These observations are useful to define sets of measures for every single wavelet coefficients: given a single wavelet coefficient $y=\delta_{j,\bar{j}}\cdot\delta_{k,\bar{k}}$, and fixed $\bar{j}, \bar{k}$, the cardinality $m' = |J'|$ of the set $J'=\{i:|(FW^Ty)_i|>threshold\}$ of frequencies with modulus greater then a fixed threshold is proportional to $2^{\bar{j}}\cdot 2^{-\bar{j}/2} = 2^{\bar{j}/2}$.

\begin{figure}
 \centering
  \raisebox{-0.5\height}{\includegraphics[scale=.145,clip]{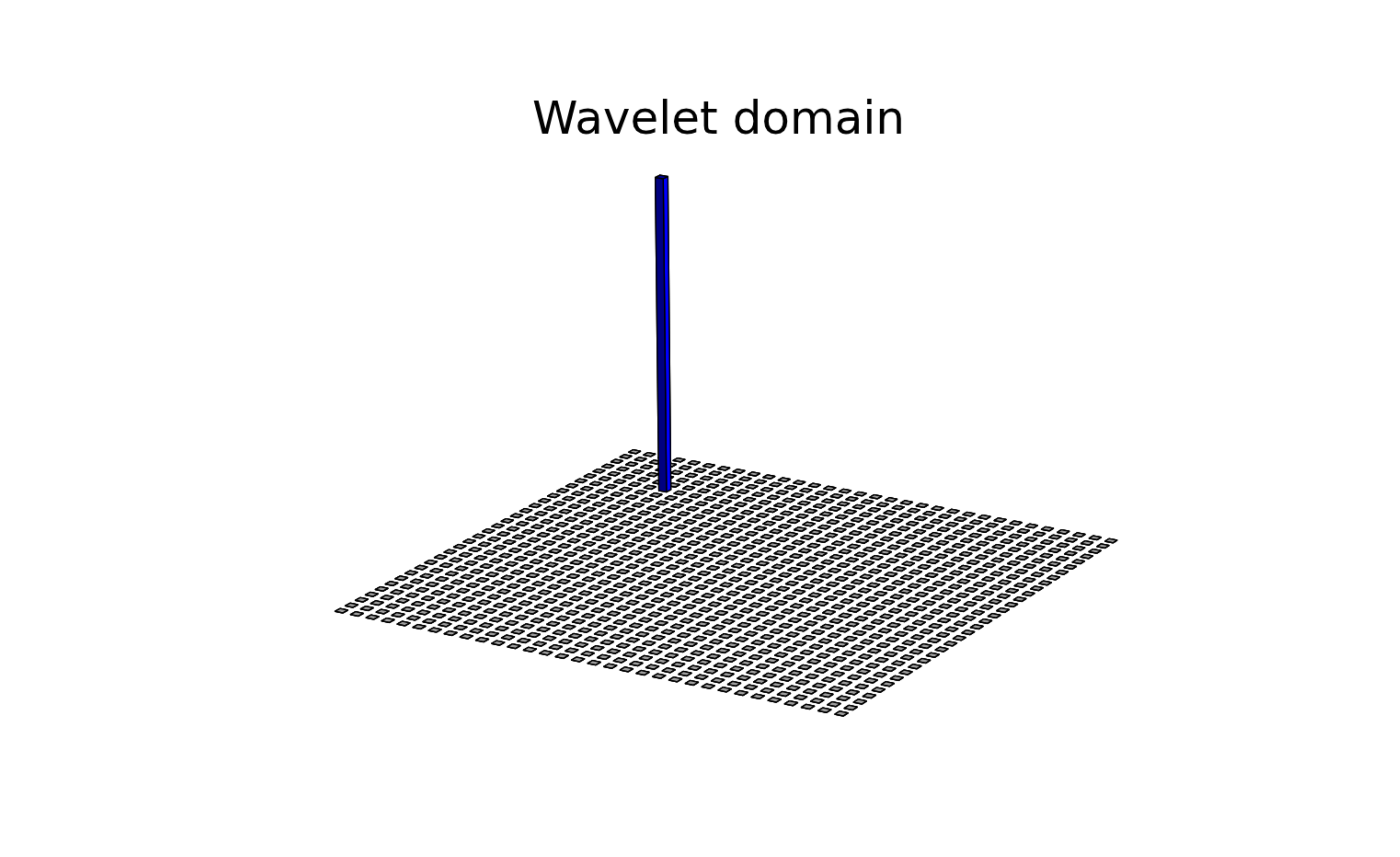}}
  \raisebox{-0.5\height}{\includegraphics[scale=.07,clip]{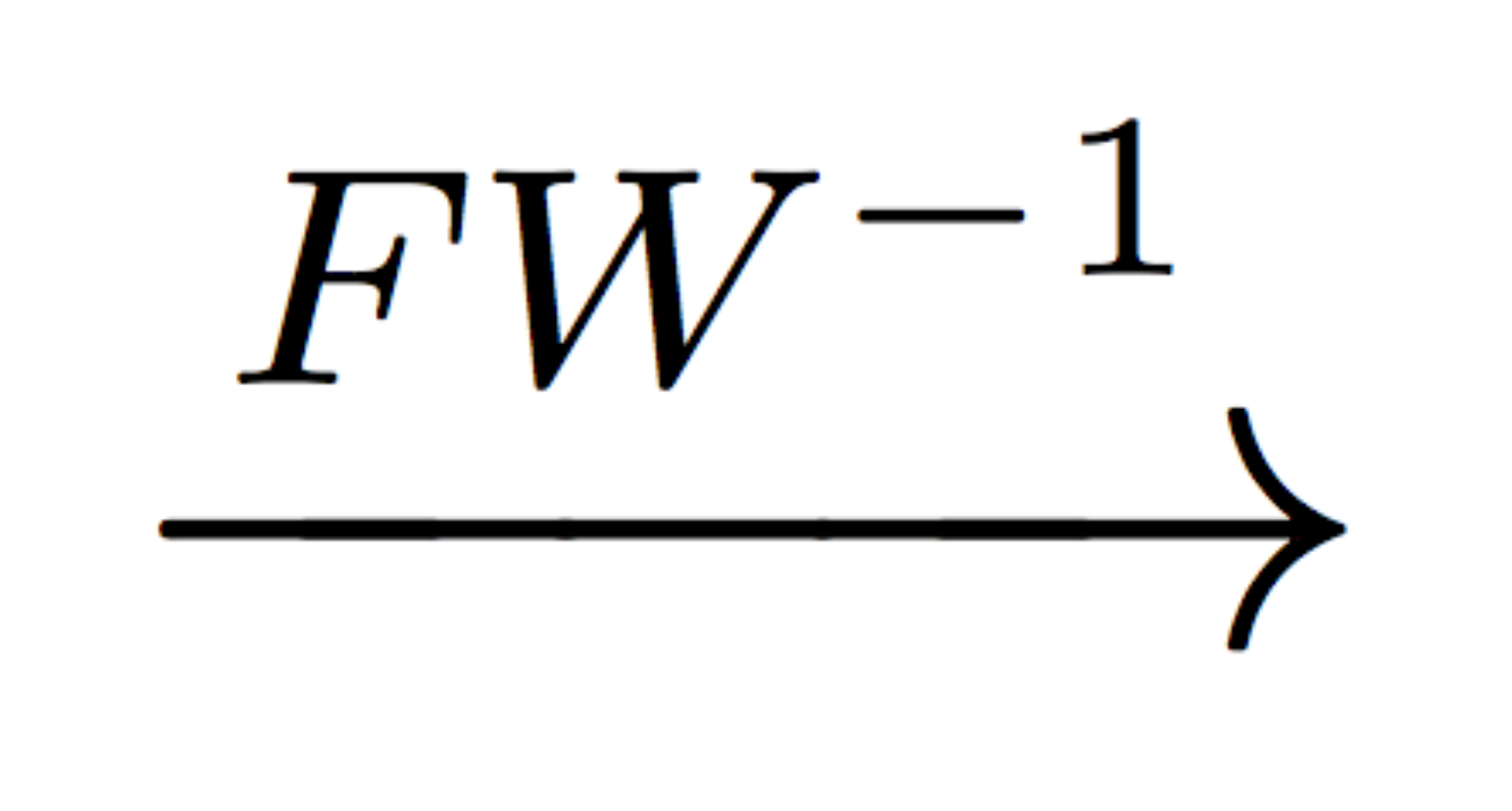}}
  \raisebox{-0.5\height}{\includegraphics[scale=.145,clip]{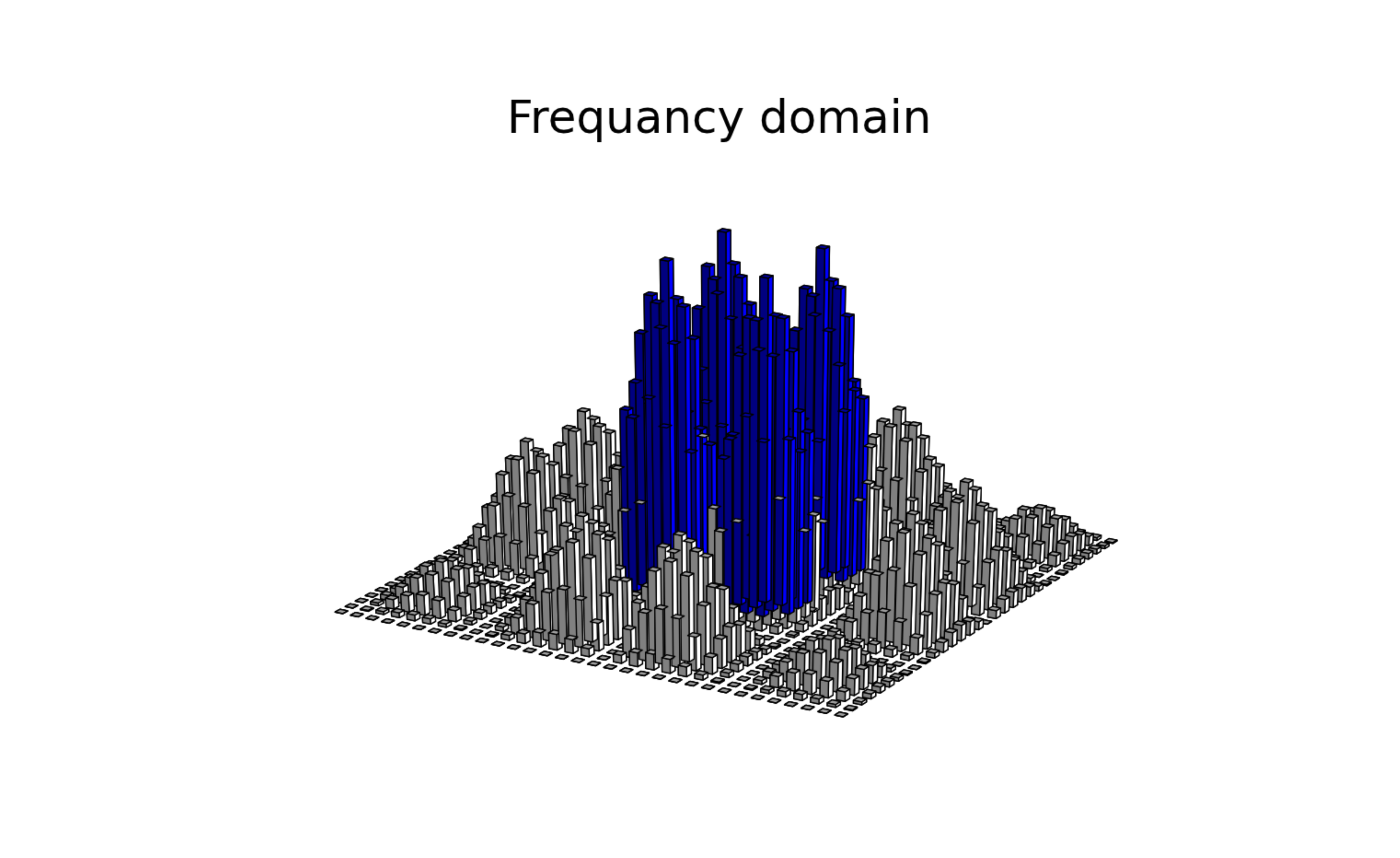}}\\
  \raisebox{-0.5\height}{\includegraphics[scale=.145,clip]{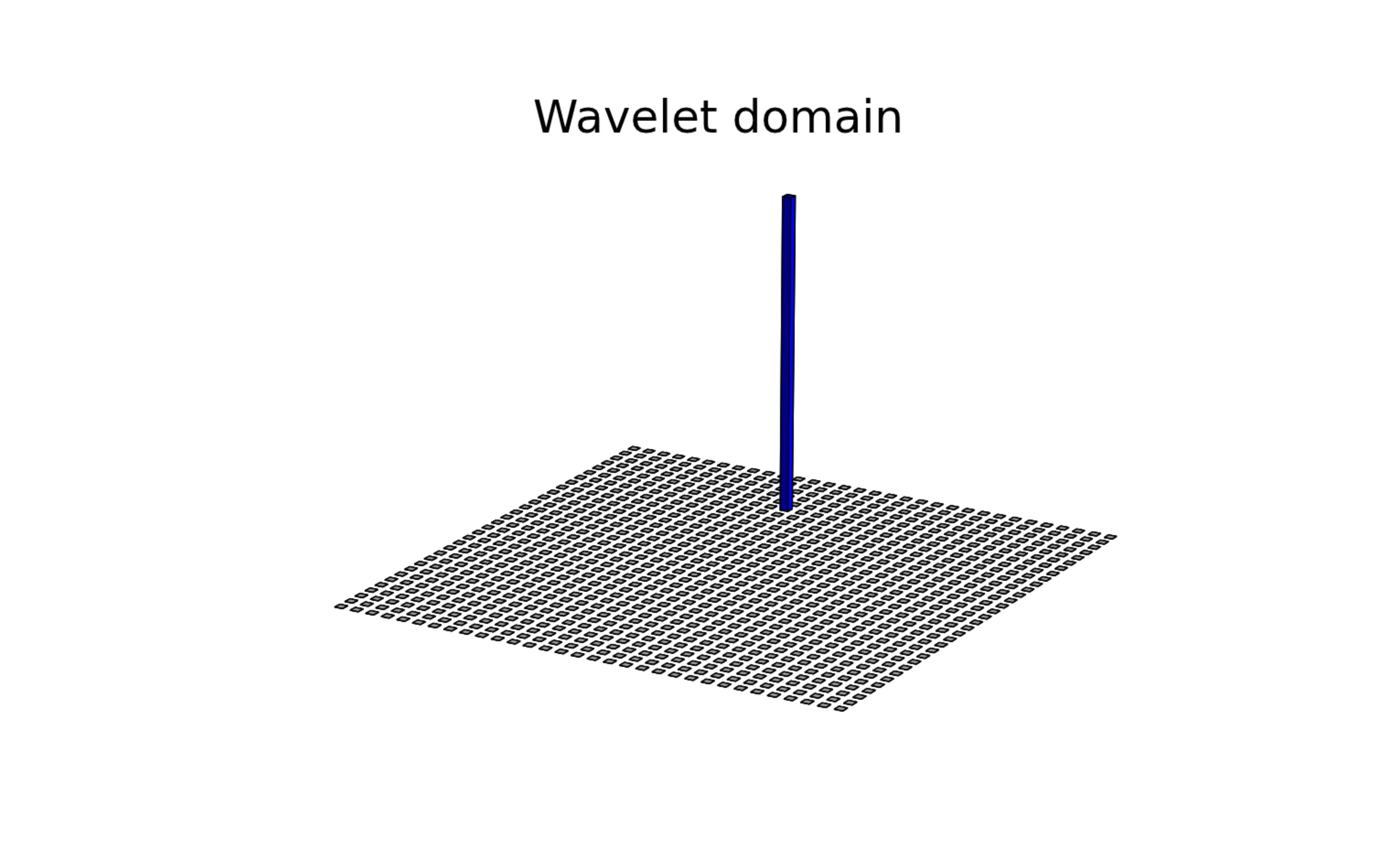}}
  \raisebox{-0.5\height}{\includegraphics[scale=.07,clip]{plot_rightarrow-eps-converted-to.pdf}}
  \raisebox{-0.5\height}{\includegraphics[scale=.145,clip]{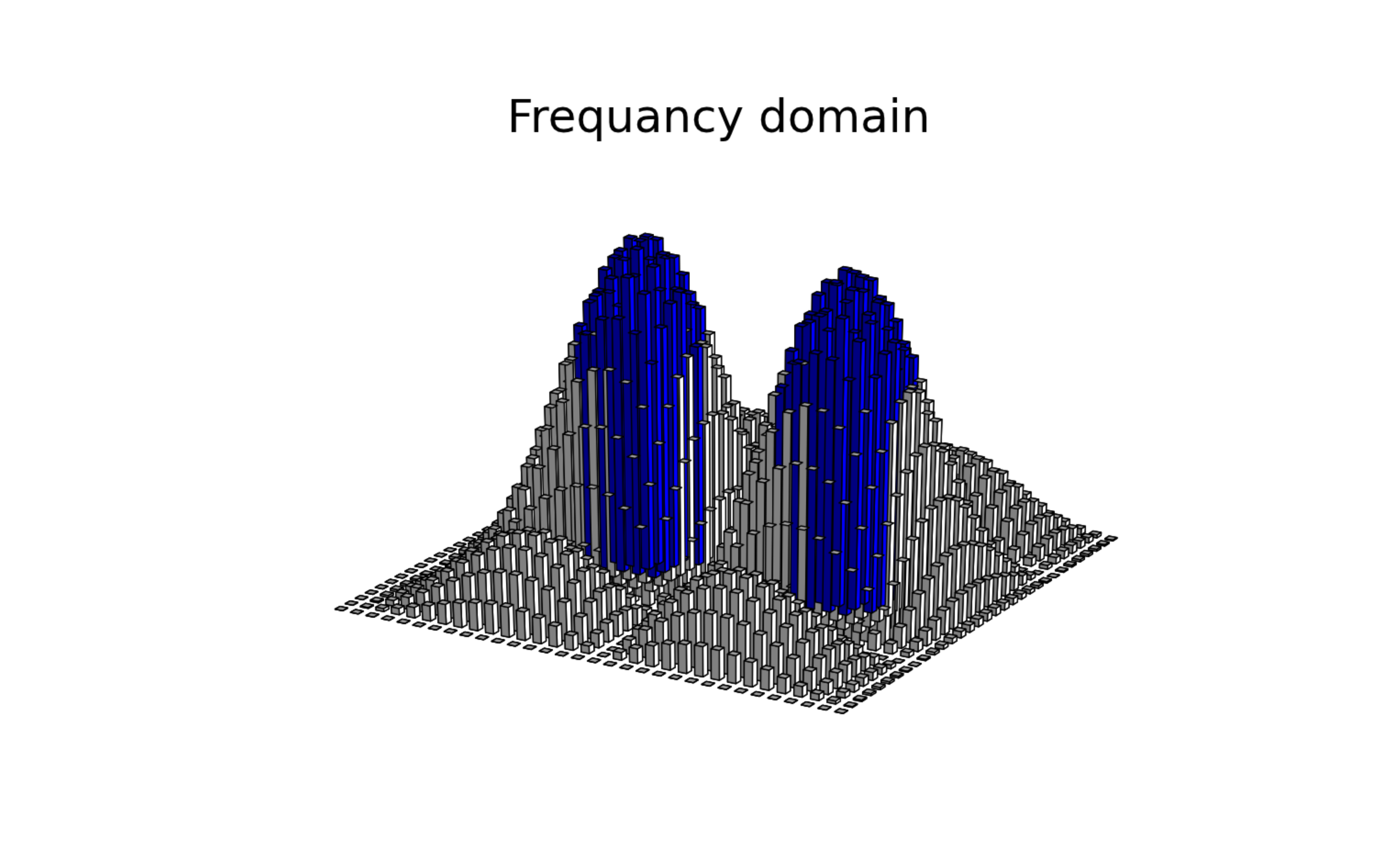}}
  \caption{For all the orders of wavelet resolution it is calculated the frequency spectrum of a single wavelet coefficient, then the frequencies with the biggest modulus are added to $J$.}
\label{fig:algo2}
\end{figure}

Suppose that the wavelet representation $y=W\bar{x}$ of the signal $\bar{x}$ has only a set $I,\hspace{.2cm} |I|=n$ of non-zero coefficients, and let $n_j, j=0,...,ord_{max}$ the numbers of the non-zero coefficients in every subsequence of different resolution (let the minimum order trend subsequence have index $0$, as the maximum order fluctuation sequence has index $ord_{max}$). We can now exploit the facts just described to compute the set $J$: for every resolution order of the wavelet representation $y$ we add to $J$ a group of frequencies with maximum modulus such that their cardinality is proportional to the number of non-zero wavelet coefficients, to the size of the spectrum support, and to the maximum modulus of the spectrum (figure \ref{fig:algo2}).

The algorithm is then:\\
\begin{algorithm}[H]
 \KwData{$n=|\{i:y_i\neq0\}|,\hspace{.2cm} n_j,\hspace{.2cm} j=0,...,ord_{max}$}
 $J = \emptyset$\;
 \For{$\bar{j} = 0 \to ord_{max}$}{
   $c = \delta_{j,\bar{j}}\cdot\delta_{k,0}$\;
   \Comment{only the coefficient corresponding to $k = 0$ and $j=\bar{j}$ is non-zero}\;
   $f = FW^Tc$\;
   $m' = 2^{j/2}\cdot n_{\bar{j}}/n \cdot \textnormal{param}$\;
   \Comment{param is adjustable if different weights are needed}\;
   \For{$i = 1 \to m'$}{
     $h = \arg\max|f|$\;
     $J \leftarrow h$\;
     $f_h = 0$\;}
   }
 \label{algo2}
\end{algorithm}

\subsection{Algorithm 3}

The analysis of the transforms $\Mca{W}$ and $\Mca{F}$ has however pointed out an important event that the previous algorithm does not consider: wavelet representations in which there are more than one non-zero coefficient exhibit frequency spectra with interference-like effects. So we defined a more refined algorithm that computes better measurements sets in respect of this occurrence. The idea is to iteratively compute frequency spectra of an increasing number of wavelet coefficients, in which we can select a new frequency to be added to the set $J$.

\begin{figure}
 \centering
  \raisebox{-0.5\height}{\includegraphics[scale=.145,clip]{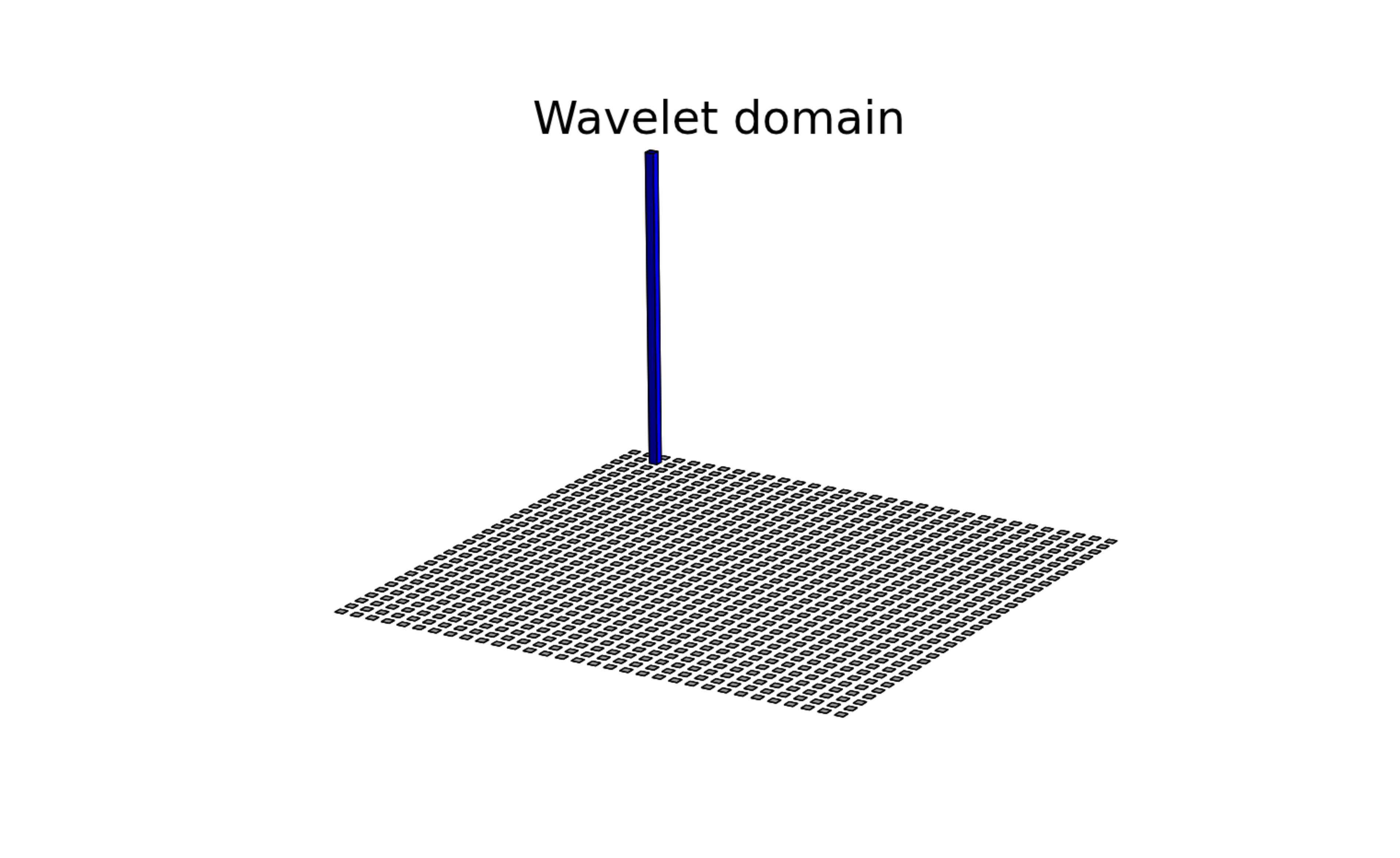}}
  \raisebox{-0.5\height}{\includegraphics[scale=.07,clip]{plot_rightarrow-eps-converted-to.pdf}}
  \raisebox{-0.5\height}{\includegraphics[scale=.145,clip]{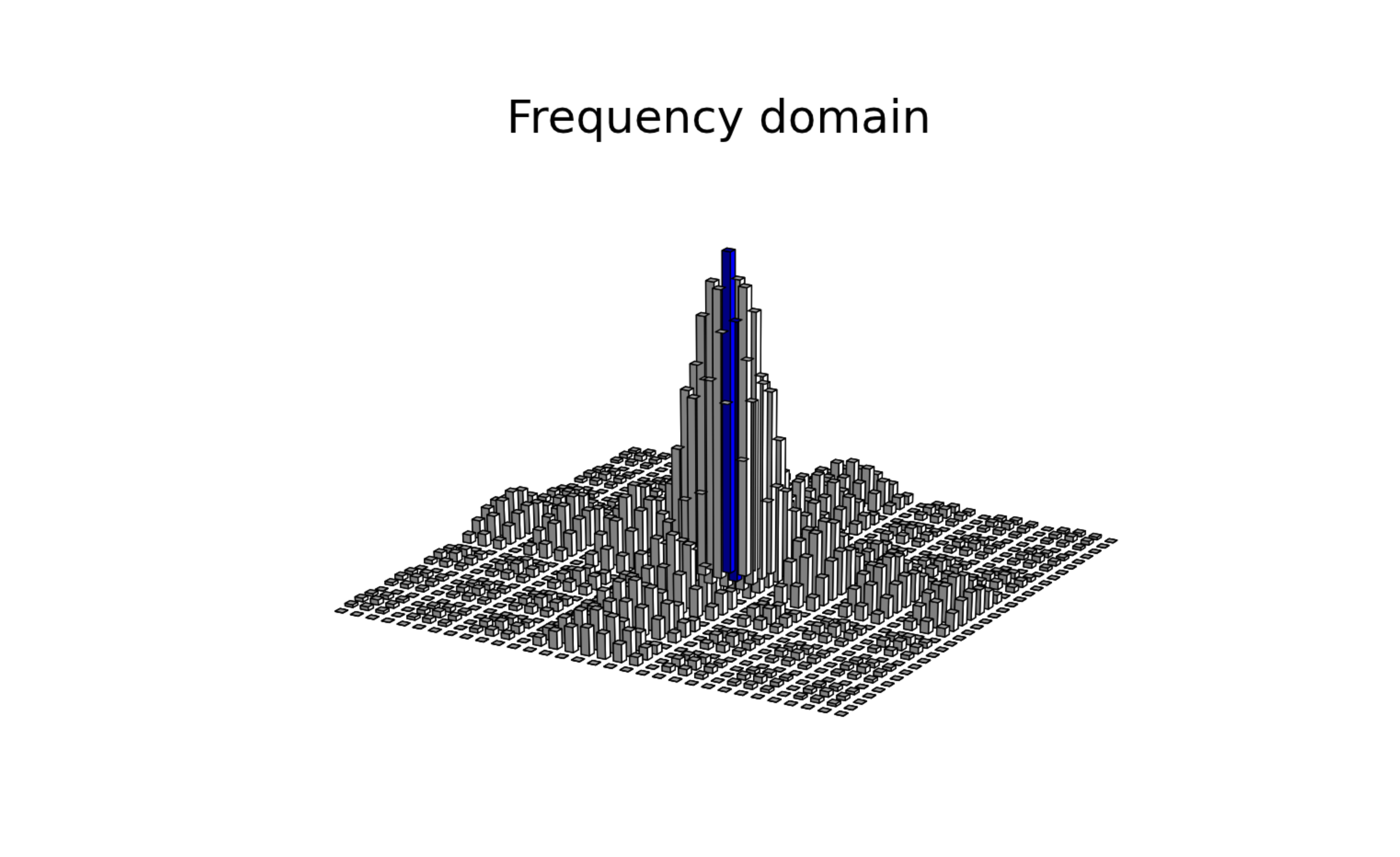}}\\
  \raisebox{-0.5\height}{\includegraphics[scale=.145,clip]{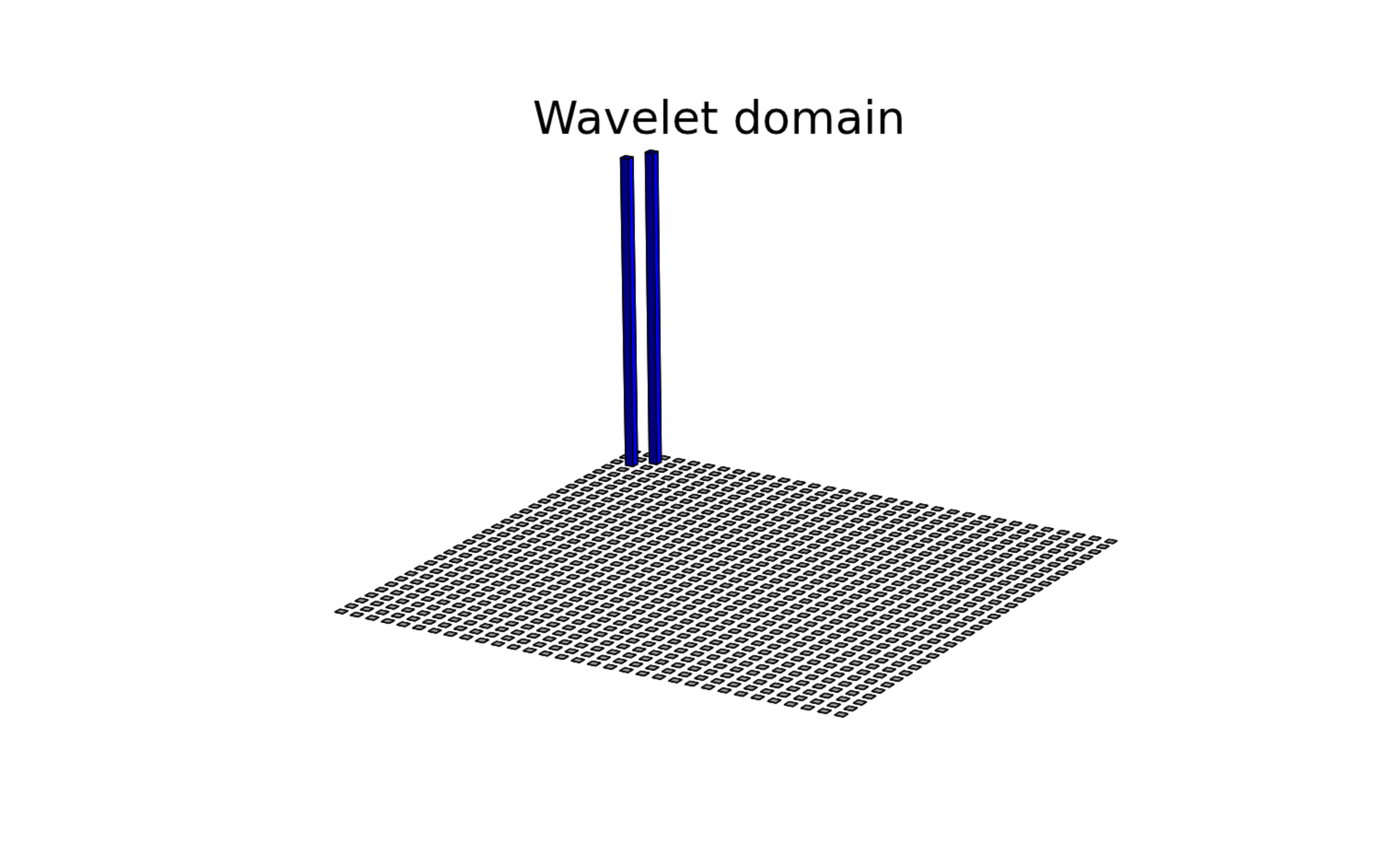}}
  \raisebox{-0.5\height}{\includegraphics[scale=.07,clip]{plot_rightarrow-eps-converted-to.pdf}}
  \raisebox{-0.5\height}{\includegraphics[scale=.145,clip]{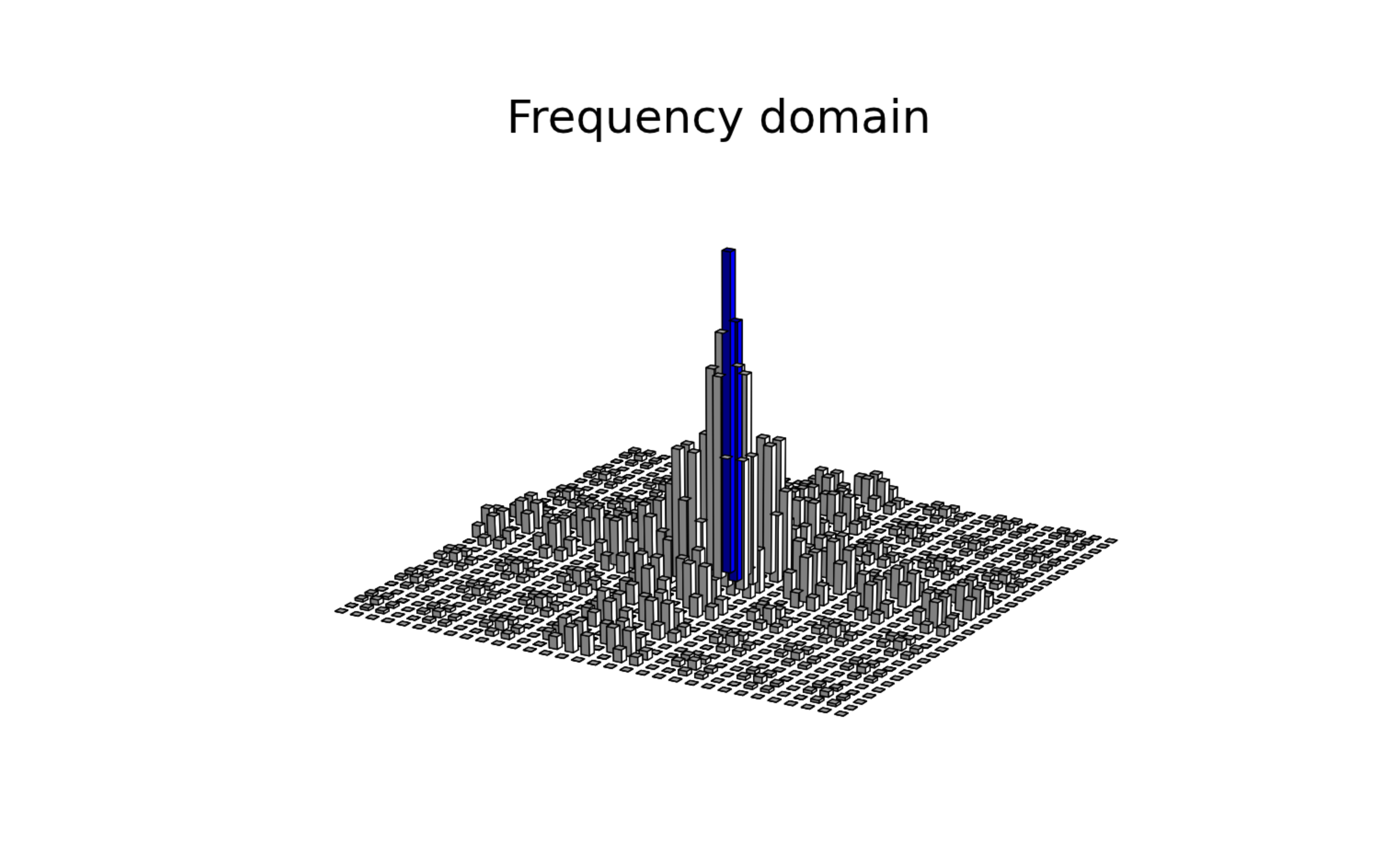}}
  \caption{For a single non-zero wavelet coefficient of $x$, the frequency that has the biggest modulus is added to $J$. A second non-zero wavelet coefficient of $x$ is considered with the first one and the frequency that has the biggest modulus and is not already in $J$ is added to $J$. The process continues for all the non-zero coefficients of $x$.}
\label{fig:algo3}
\end{figure}

Let $y = W\bar{x}$ be the wavelet representation of the signal $\bar{x}$. Let $I=\{i:y_i\neq0\}$ be the set of indices of the non-zero wavelet coefficients, $|I|=n$. Let $I$ be sorted with respect to the descending values $|y_i|,\hspace{.2cm} i\in I$. Let $y_{(l)} = \sum_{i\in I, i<l}y_i e_i$ (i.e. $y$ with every element smaller then the $l$ elements of maximum modulus set to zero). Then for every $l = 1,...,n$ we calculate the spectrum $f=FW^Ty_{(l)}$, and we add $\arg\max|f|$ to the set $J$; while this frequency is already an element of $J$, we set it to zero and add $\arg\max|f|$ to $J$ again (figure \ref{fig:algo3}).

The algorithm is the following:\\
\begin{algorithm}[H]
 \KwData{$y=W\bar{x},\hspace{.2cm} I=\{i:y_i\neq0\},\hspace{.2cm} |I|=n$}
 $J = \emptyset$\;
 Sort $I$ in decreasing order of $|y_i|,\hspace{.2cm} i\in I$\;
 $\bar{y}=0$\;
 \For{$l = 1 \to n$}{
   $\bar{y} = \bar{y} + y_l\cdot e_l$\;
   \Comment{next element of $y$ in decreasing order of modulus value is added}\;
   $f=FW^T\bar{y}$\;
   $h = \arg\max|f|$\;
   \While{$h \in J$}{
     $f_h = 0$\;
     $h = \arg\max|f|$\;}
   $J \leftarrow h$\;}
 \label{algo3}
\end{algorithm}

\subsection{Algorithm 4}

Suppose that the positions of the non-zero wavelet coefficients of the signal $\bar{x}$ are known. Let us find the frequencies that ``influence more'' such coefficients: with this expression we mean the frequencies that have a wavelet spectrum (by the transform $\Mca{W}\circ\Mca{F}^{-1}$) with values over a fixed threshold in the same positions of the non-zero coefficients of $y=W\bar{x}$.

\begin{figure}
 \centering
  \raisebox{-0.5\height}{\includegraphics[scale=.145,clip]{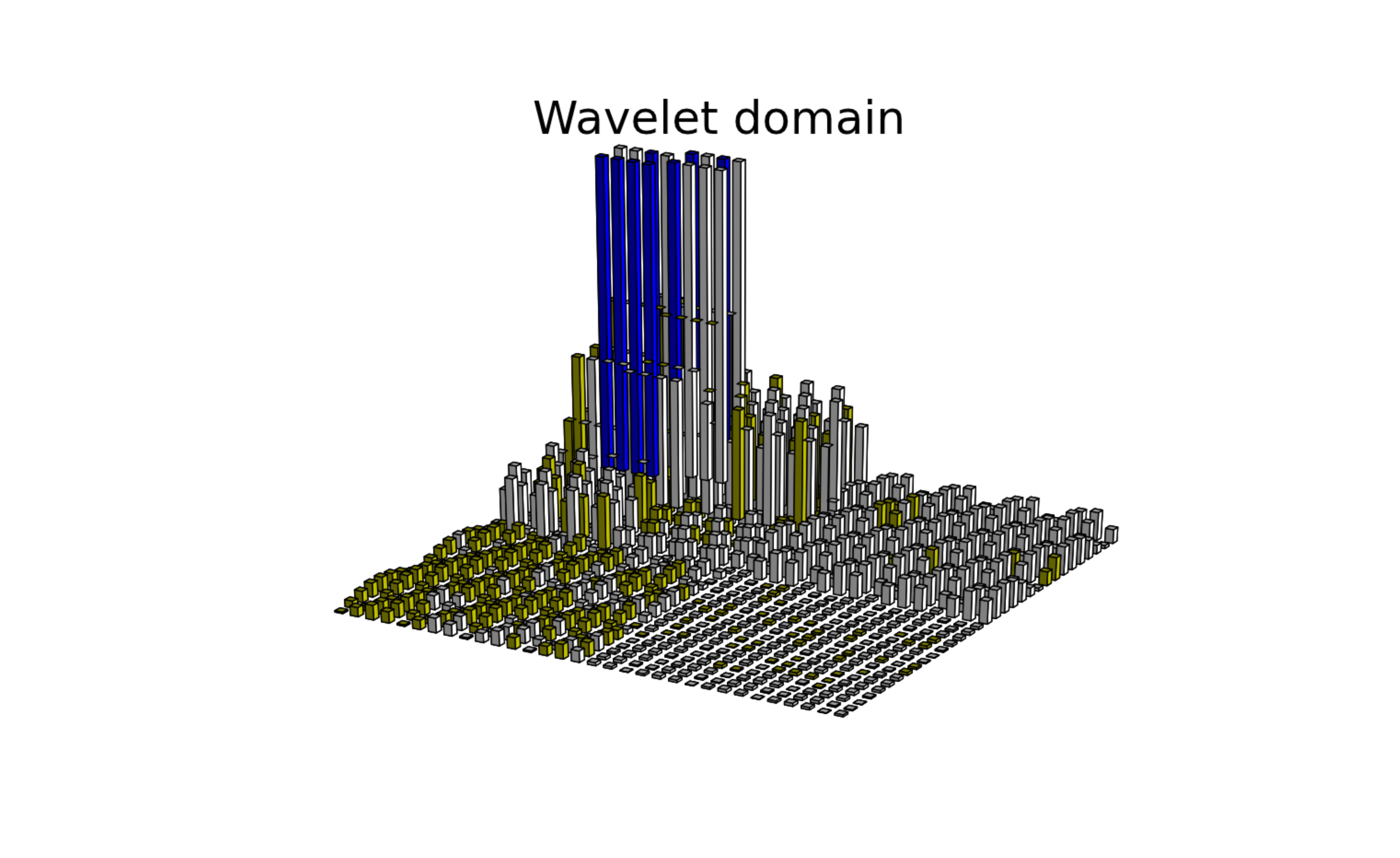}}
  \raisebox{-0.5\height}{\includegraphics[scale=.07,clip]{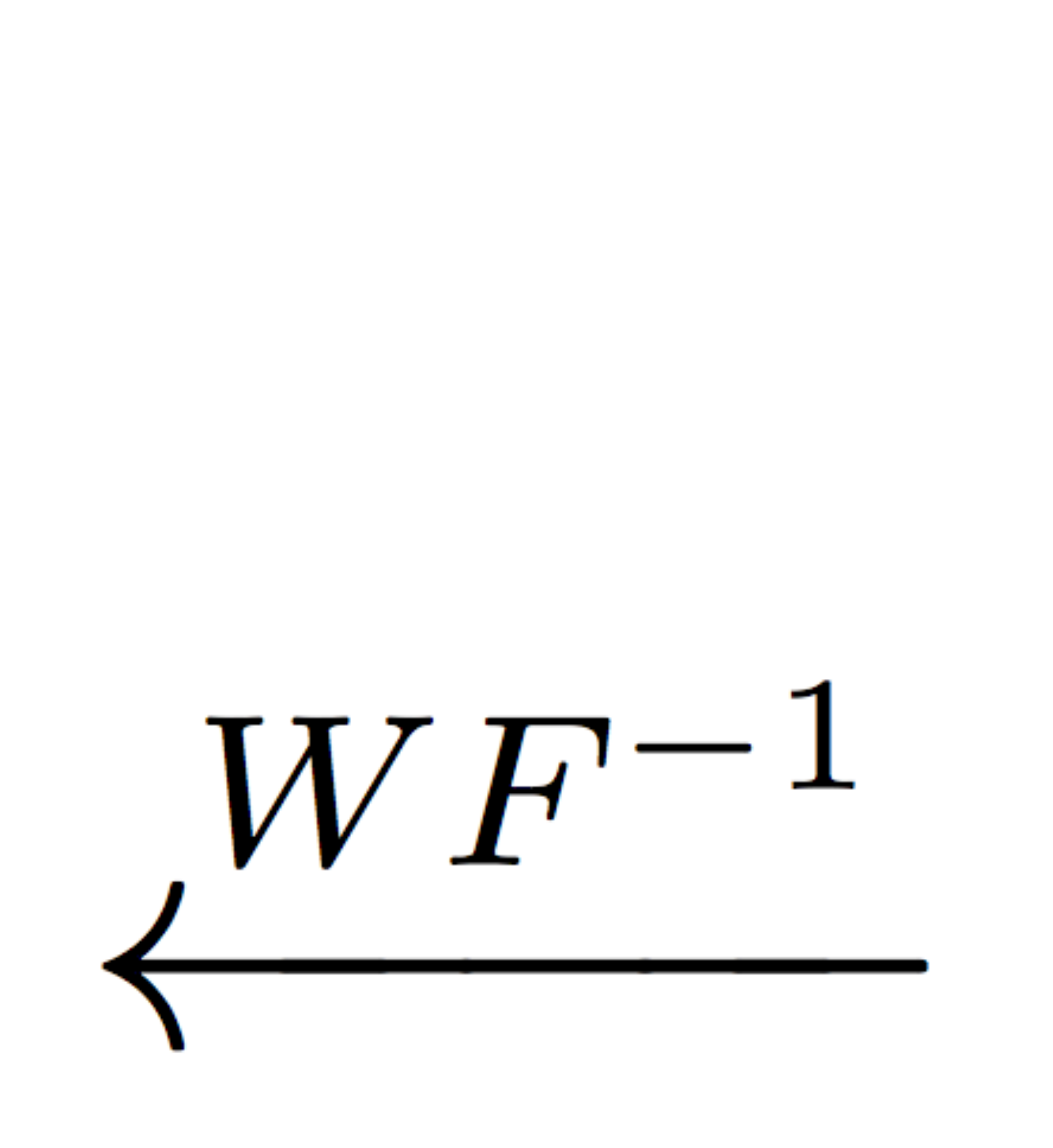}}
  \raisebox{-0.5\height}{\includegraphics[scale=.145,clip]{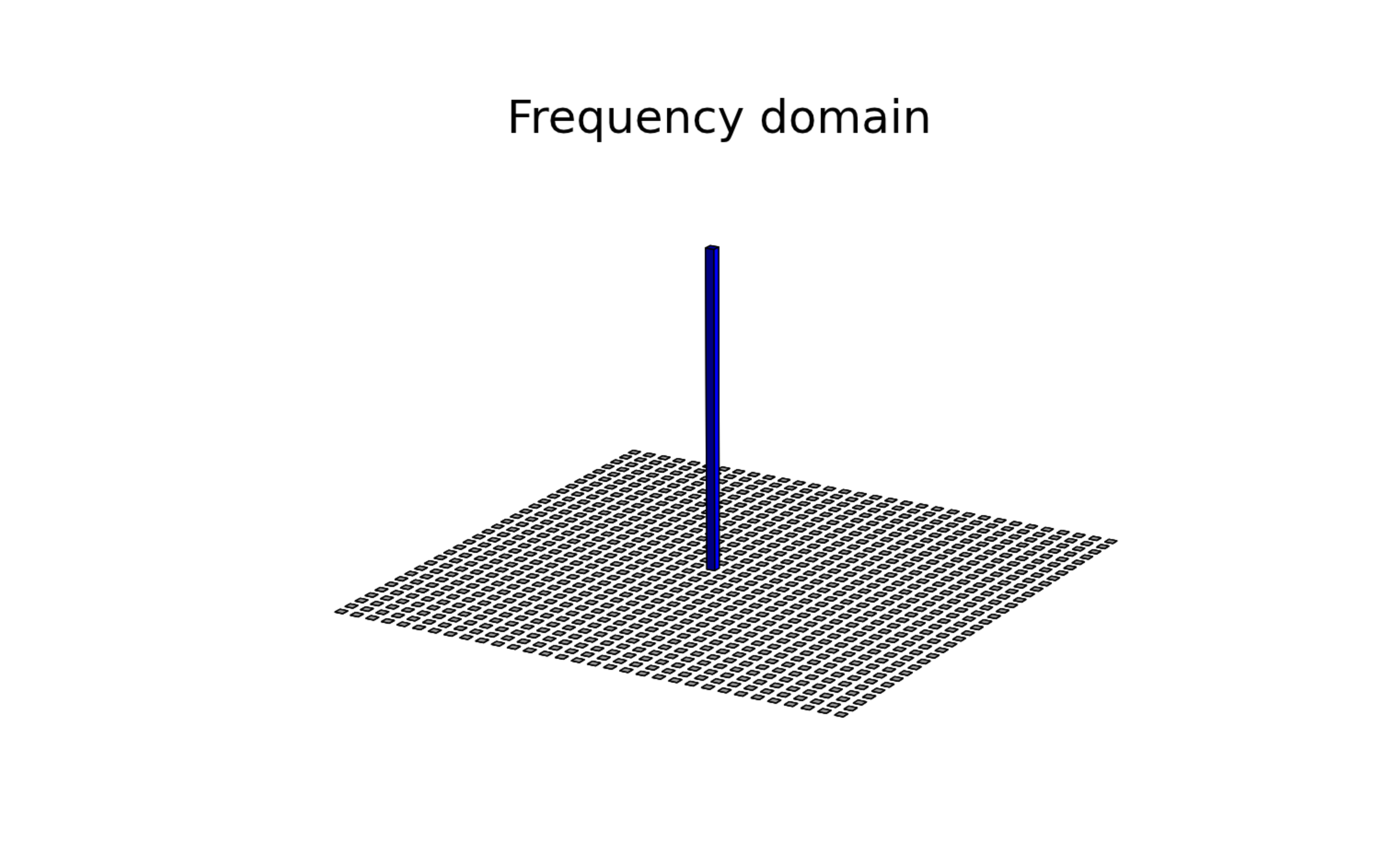}}
  \caption{A single frequency and its wavelet spectrum. The highlighted coefficients correspond to the non-zero wavelet coefficients of $x$. Between all frequencies, those that have at least one highlighted wavelet coefficient over some fixed threshold are added to $J$.}
\label{fig:algo4a}
\end{figure}

Let $I = \{i \in \{1,..,N\}: y_i \neq 0\}$ the positions of the non-zero coefficients of $y$, then the frequencies $f_h$ we want to measure are the ones that transformed by $WF^T$ have at least one of the components in $I$ over a fixed threshold. Since the $h$-th frequency corresponds to the $h$-th row of $FW^T$ and the $i$-th component of $y$ to the $i$-th column of $FW^T$, the problem reduces to compute the indices $h \in \{1,..,N\}$ such that $\max(WF^Te_h)_{i \in I}>s_1$, with $s_1$ a fixed threshold (figure \ref{fig:algo4a}).

\begin{figure}
 \centering
  \raisebox{-0.5\height}{\includegraphics[scale=.145,clip]{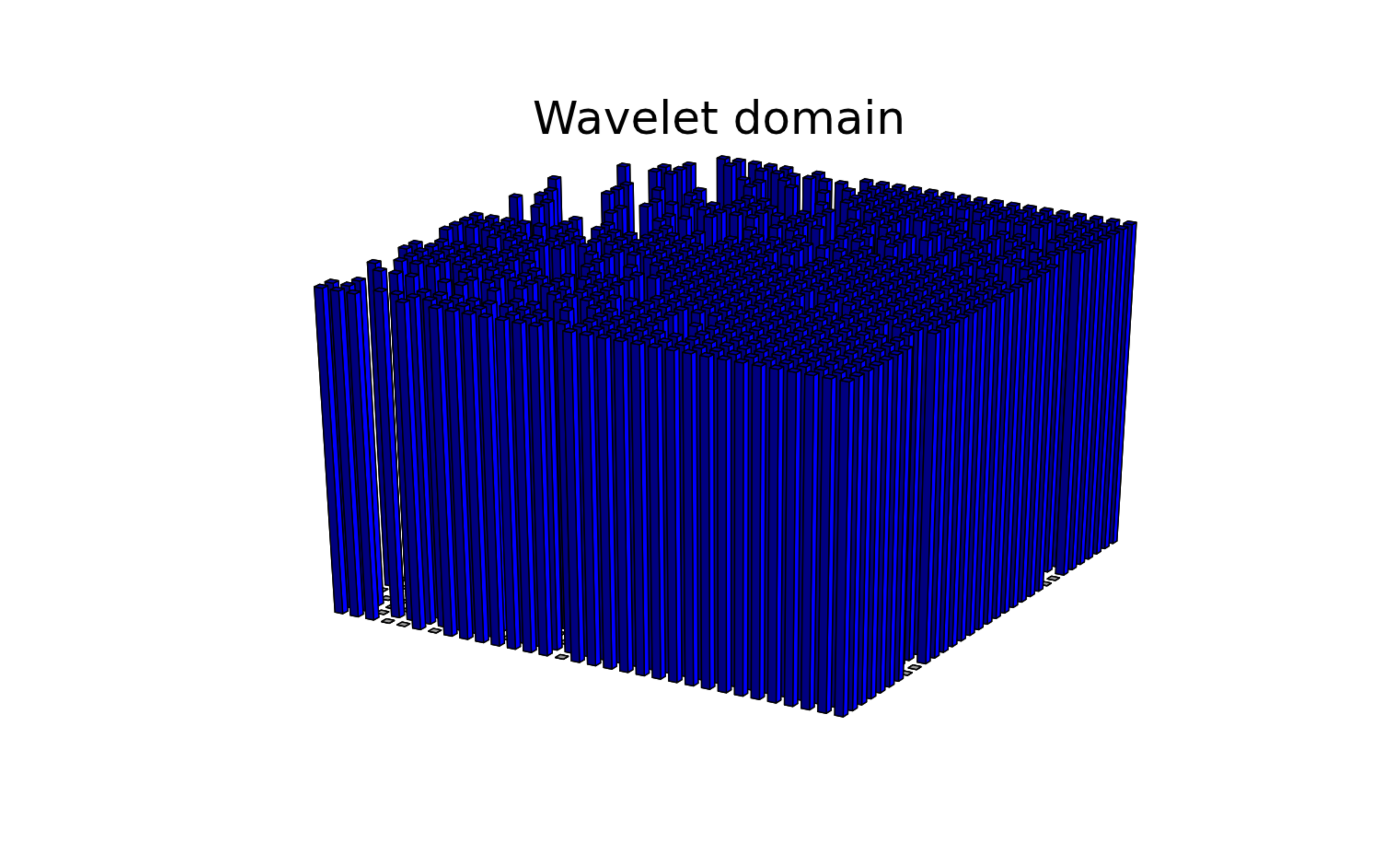}}
  \raisebox{-0.5\height}{\includegraphics[scale=.07,clip]{plot_rightarrow-eps-converted-to.pdf}}
  \raisebox{-0.5\height}{\includegraphics[scale=.145,clip]{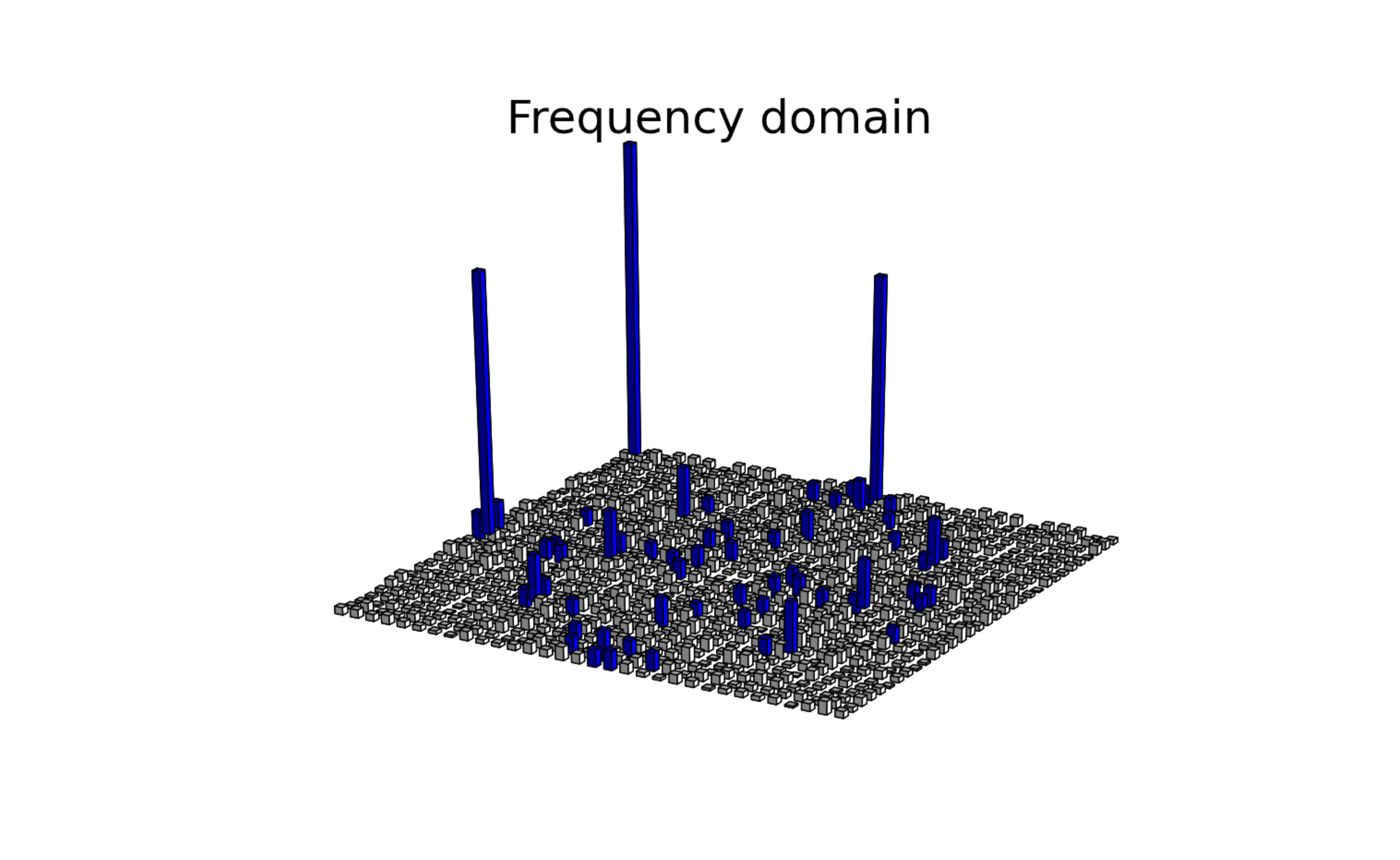}}
  \caption{The zero wavelet coefficients of $x$ (set to value $1$) and their frequency spectrum. The frequencies that are over fixed threshold are discarded.}
\label{fig:algo4b}
\end{figure}

Dually, we discard the frequencies that would influence more the zero wavelet coefficients of the image. We compute the frequency spectrum of the zero wavelet coefficient set to the unitary value (but they could also be weighted differently) and we exclude the frequencies whose modulus exceeds a fixed value. Since $S = \{i \in \{1,..,N\}: y_i = 0\}$, then we discard the frequencies correspondent to the indices $h \in \{1,..,N\}$ such that $(FW^Ty_S)_h>s_2$, with $s_2$ a fixed value (figure \ref{fig:algo4b}).

The algorithm is as follows:\\
\begin{algorithm}[H]
 \KwData{$I=\{i:y_i\neq0\}, |I|=n$}
 \For{$h = 1 \to N$}{
 	 $v(h) = \max_{i \in I}|WF^Te_h|$\;
 	 \Comment{maximum element of the $h$-th column of $WF^T$ among the non-zero ones}\;}
 $\bar{y} = \sum_{i\in S}e_i$\;
 $\bar{f} = |FW^T\bar{y}|$\; 
 $J = \{h:v(h)>\textnormal{s}_1, \bar{f}(h)<\textnormal{s}_2\}.$
 \label{algo4}
\end{algorithm}

\section{Iterative algorithms}
\label{iterative algorithms}

In order to compare the results of the previous algorithms with other sparse recovery methods already studied in literature, we considered two iterative algorithms from the compressed sensing theory: the Iterative Hard Thresholding (IHT) algorithm \cite{IHT} and the Location Constraint Approximate Message Passing (LCAMP) algorithm \cite{LCAMP}. Both algorithms use $F_{J}W^T$ as the sensing matrix and an undersampling set $J\subset \{1,...,N\}$ that has a casual distribution concentrated in low frequency.

The IHT algorithm:\\
\begin{algorithm}[H]
 \KwData{$F_{J}W^T$ as CS matrix, $f\in\Bbb{C}^m$ measure vector, $n$ sparsity level}
 $\hat{x}_0=0$\;
 \For{$i = 1; i = i+1; \textnormal{until a stop criterion is met} $}{
   $\hat{x}_i = H_n(\hat{x}_{i-1}+WF_{J}^T(f-F_{J}W^T\hat{x}_{i-1}))$\;
   }
 \label{algoIHT}
\end{algorithm}

where $H_k(z)$ is the \textit{hard thresholding} operator on $z$ that set all the $z$ components to zero except the $k$ components with the maximum modulus.

The LCAMP algorithm:\\
\begin{algorithm}[H]
 \KwData{$M$ the mask of the non-zero wavelet coefficients of $\bar{x}$, $F_{J}W^T$ as CS matrix, $f\in\Bbb{C}^m$ measure vector, $N$ the dimension of the signal to recover}
 $\hat{x}_0=0, r_0=f, \rho=N/m, c=|\supp(M)|/N$\;
 \For{$i = 1; i = i+1\hspace{.2cm}; \textnormal{until a stop criterion is met} $}{
   $r_i = f-F_{J}W^T\hat{x}_{i-1}+\rho\cdot c\cdot r_{i-1}$\;
   $\hat{x}_{i} = (\hat{x}_{i-1}+WF_{J}^Tr_i)\cdot M$\; 
   }
   \label{algoLCAMP}
\end{algorithm}

The stop criterion in both algorithms can consist in a limit of the iteration number or a convergence criterion for $\hat{x}$.

\section{Data}
\label{data}

\subsection{Real data}
A series of $T2^*$ weighted volumes ($128\times 128\times 20$ voxels with in plane resolution of $1.8\times1.8$ mm and slice thickness of 4 mm, $TR=1500$ ms, $TE=40$ ms) were acquired from a patient affected by artero-venous malformation, after a bolus of gadolinium (Gd-TPA).

\subsection{Simulated data}
We simulated a set of a DSC-MRI sequence, using a $256\times256$ pixels Shepp-Logan phantom \cite{shepp74}, commonly used in the simulations in the field of MRI. To the phantom we added white noise to decrease the signal to noise ratio (SNR) to $15$ dB. Moreover, the  simulation of the local transit of contrast agent was obtained by adding to specific regions a monovariate gamma function $y_{gad}(t)$ with fixed parameters with a secondary gamma function $y_{rec}(t)$ representing the recirculation \cite{peruzzo12}; to the gamma function a log-normally distributed noise was added.

\section{Comparing the reconstructions}
\label{comparison}

We ran all the above algorithms on three sets of time-series images: a sequence of images from a real medical DSC-MRI exam on the encephalic region of a arterio-venous malformation (AVM) patient, and the Shepp-Logan phantom imitating an encephalic region with the addition of localized flow of a contrast agent, with and without noise.

For every sequence we show the reconstructions of one particular frame made by all the algorithms described in the previous sections and a table of the relative percent errors (2-norm) of any single frame averaged over the whole sequence. The error of a particular reconstructed (under-sampled) frame from the original frame is computed only on the central encephalic region (the background is discarded).

\subsection{Real data}

The algorithms 1 and 3 give the best reconstructions with any number of measurements we tried. The algorithms 2 and 4 performs better than the CS algorithms while the number of measures is lower than the 33\% of the total number of possible measures. Note that between the errors corresponding to the CS algorithms and the algorithms 1 or 3 there is a factor $1.5$/$2$, a little less then the factor $2$ between the necessary number of random measures and the sparsity of the signal to recover in the estimates in the CS theory \cite{DDEK12}. As we saw, the algorithms 1 and 3 do not produce the optimal set $J$, maybe such an optimal set would achieve a factor 2. LCAMP algorithm obtain good results only with a few measures, the results degrade as their number is increased: this problem is intrinsic in the AMP algorithm: it recovers a signal correctly only when the quantity $|\supp{y}|/\dim{y}$ is lesser then a certain increasing function of $\dim{y}/|J|$ \cite{Ma11}.

\begin{figure}[!htbp]
 \centering
  {\includegraphics[scale=.27,clip]{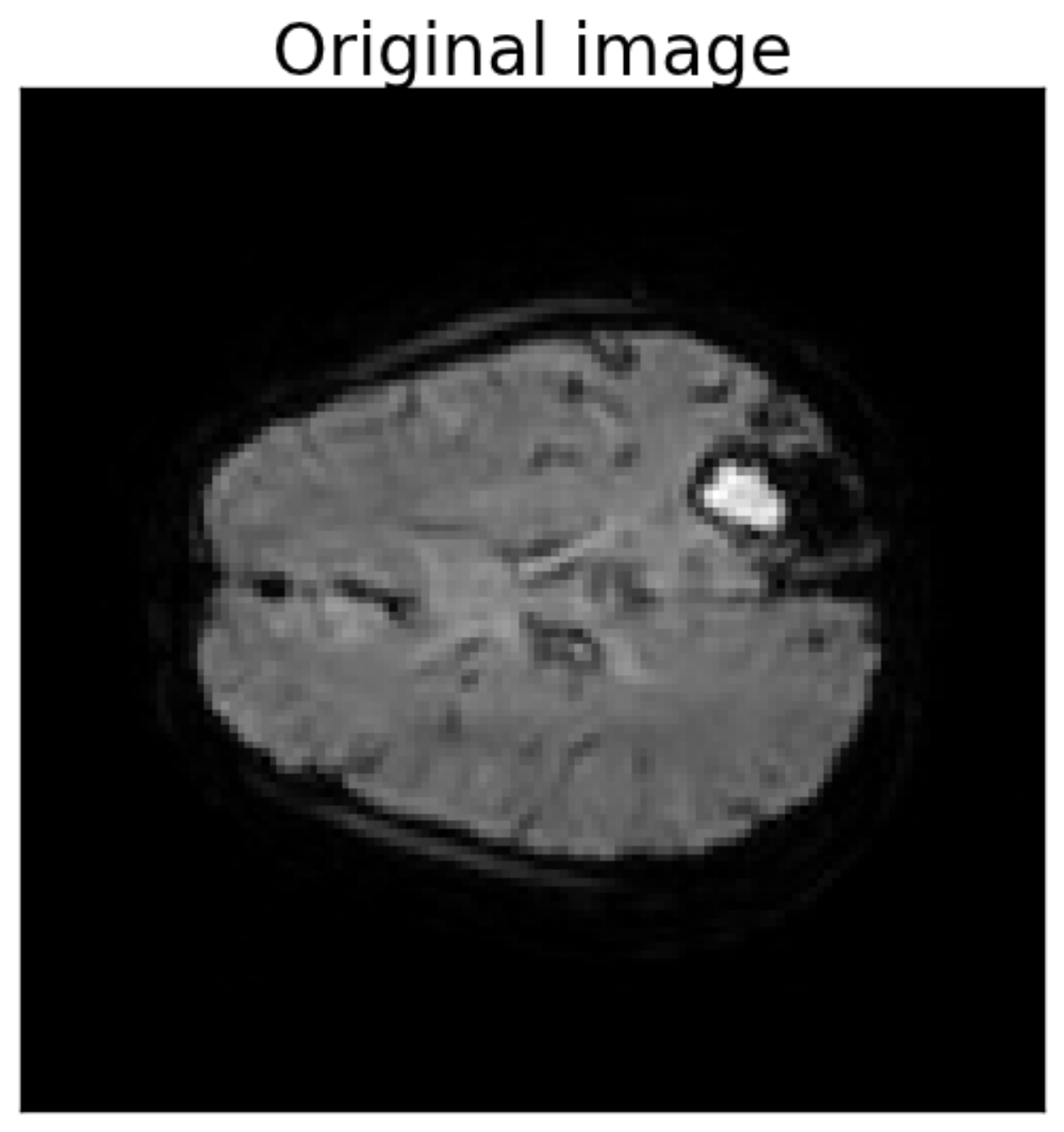}}\\
  {\includegraphics[scale=.27,clip]{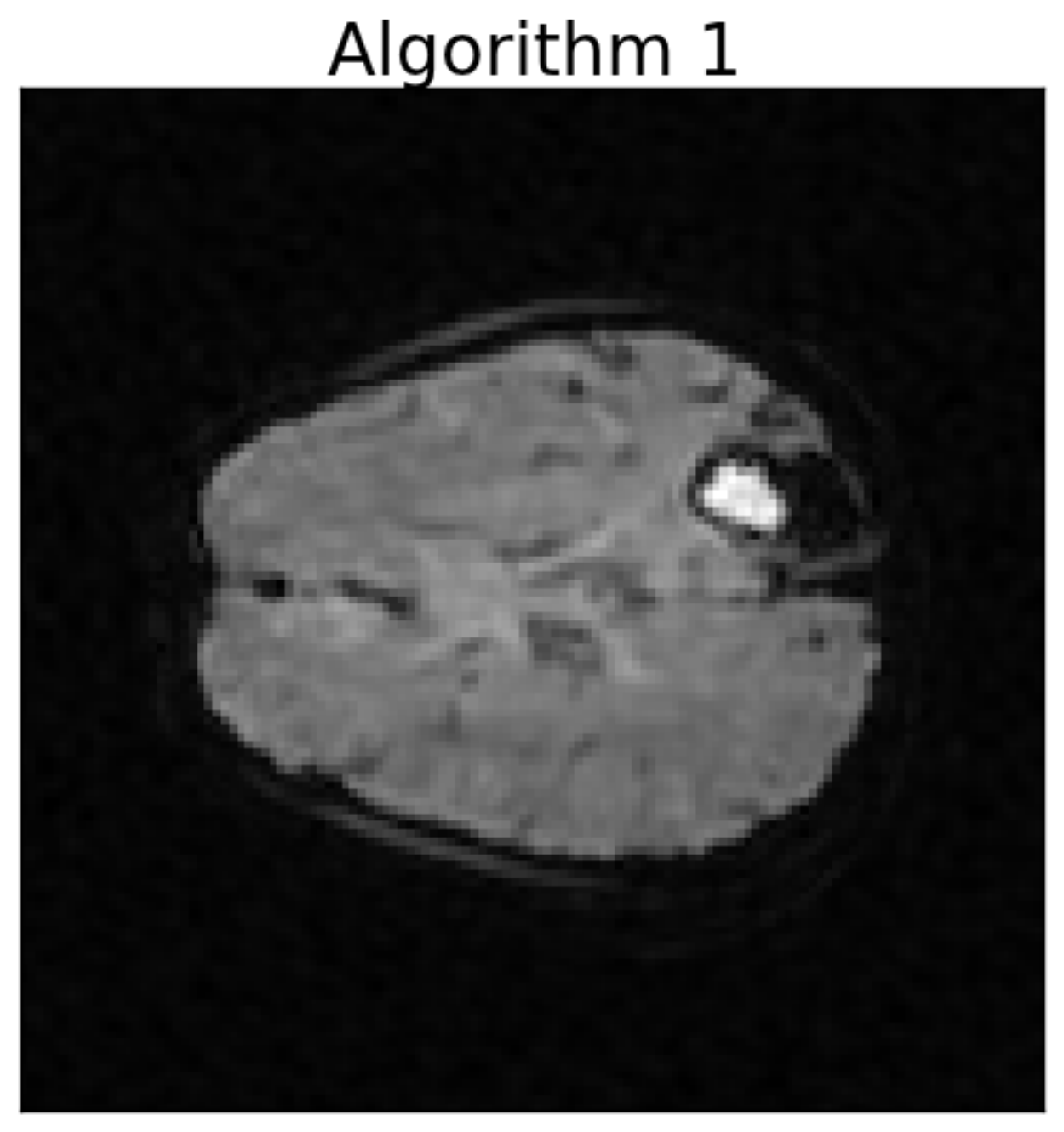}}
  {\includegraphics[scale=.27,clip]{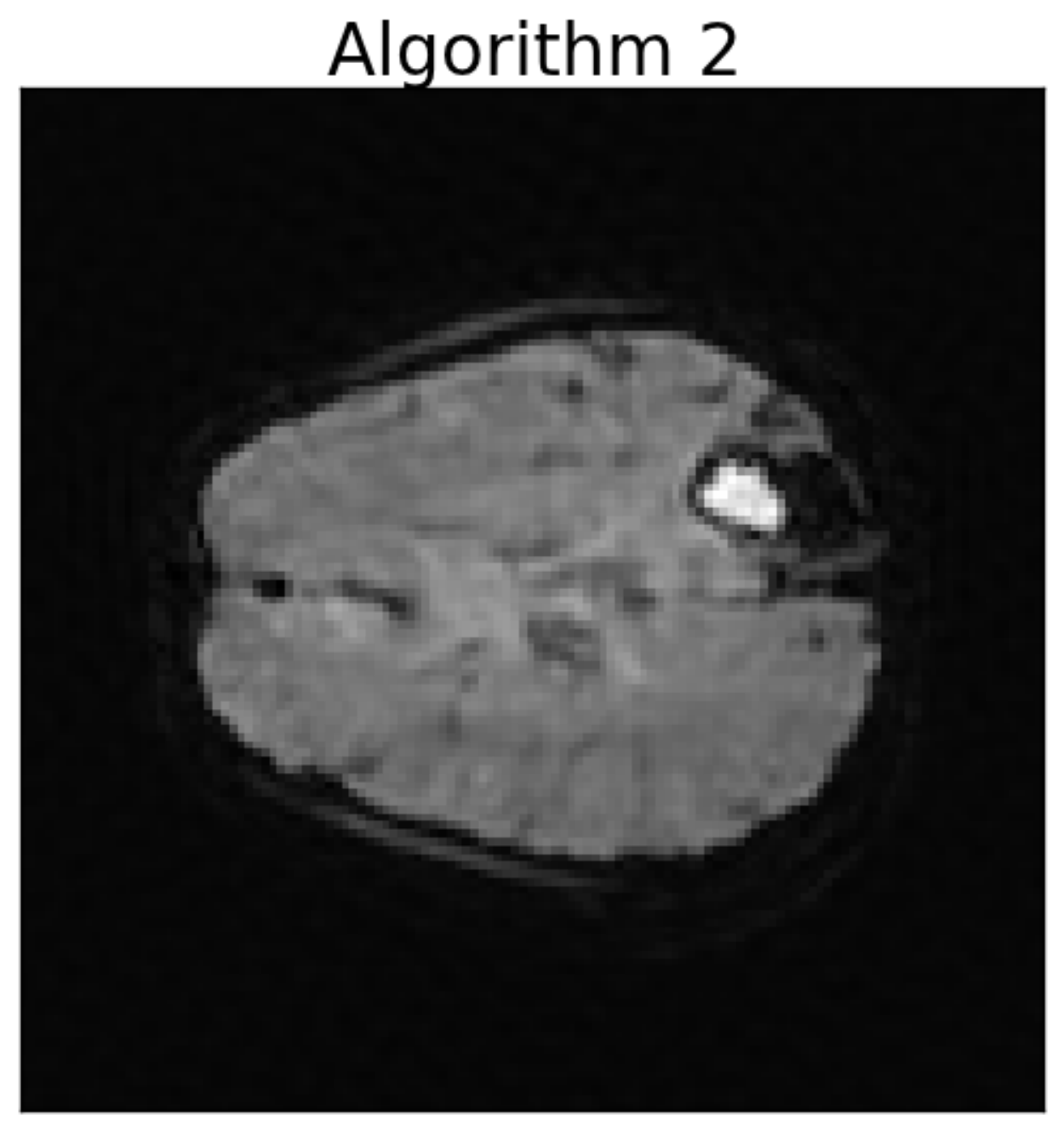}}
  {\includegraphics[scale=.27,clip]{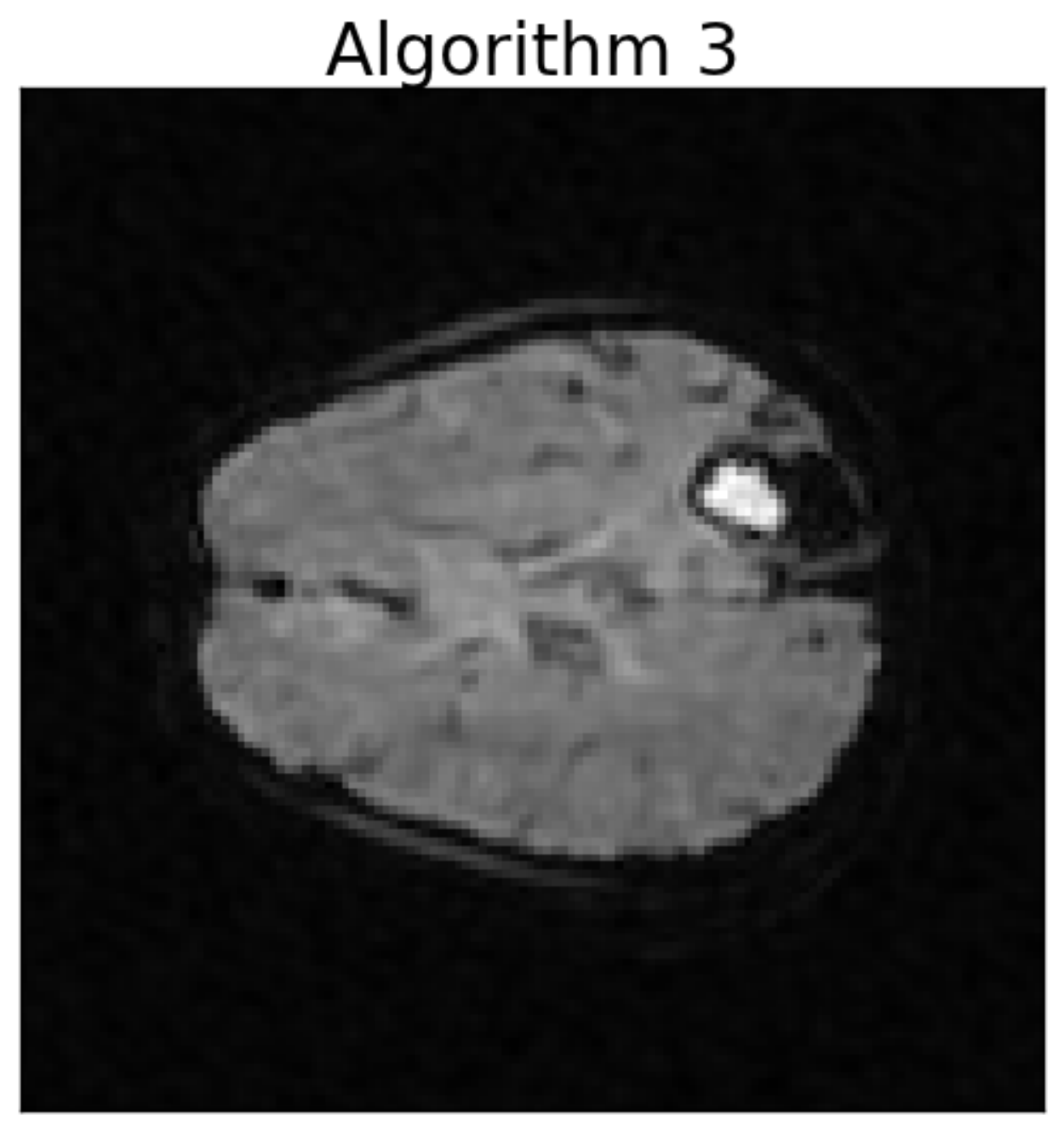}}\\
  {\includegraphics[scale=.27,clip]{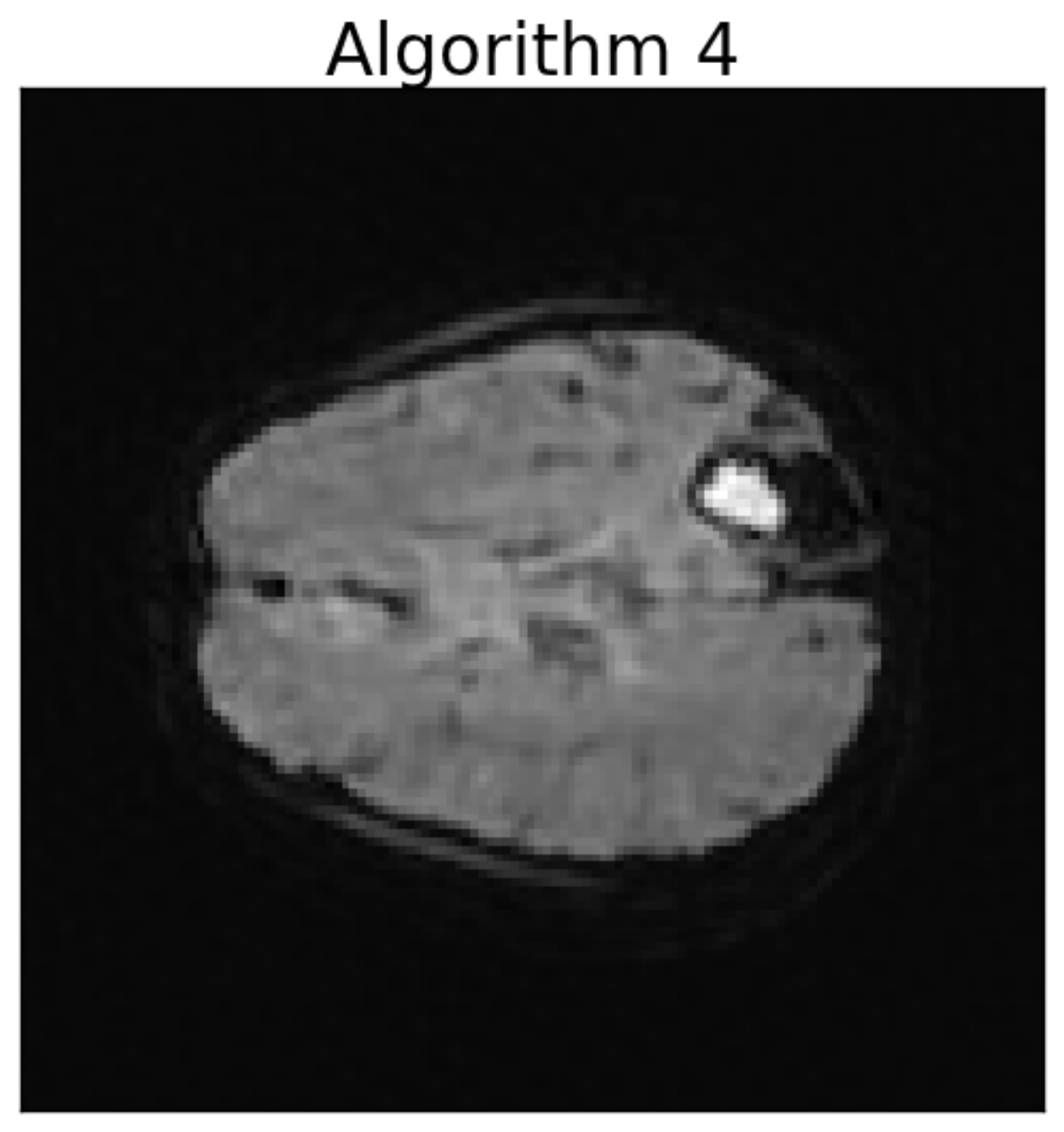}}
  {\includegraphics[scale=.27,clip]{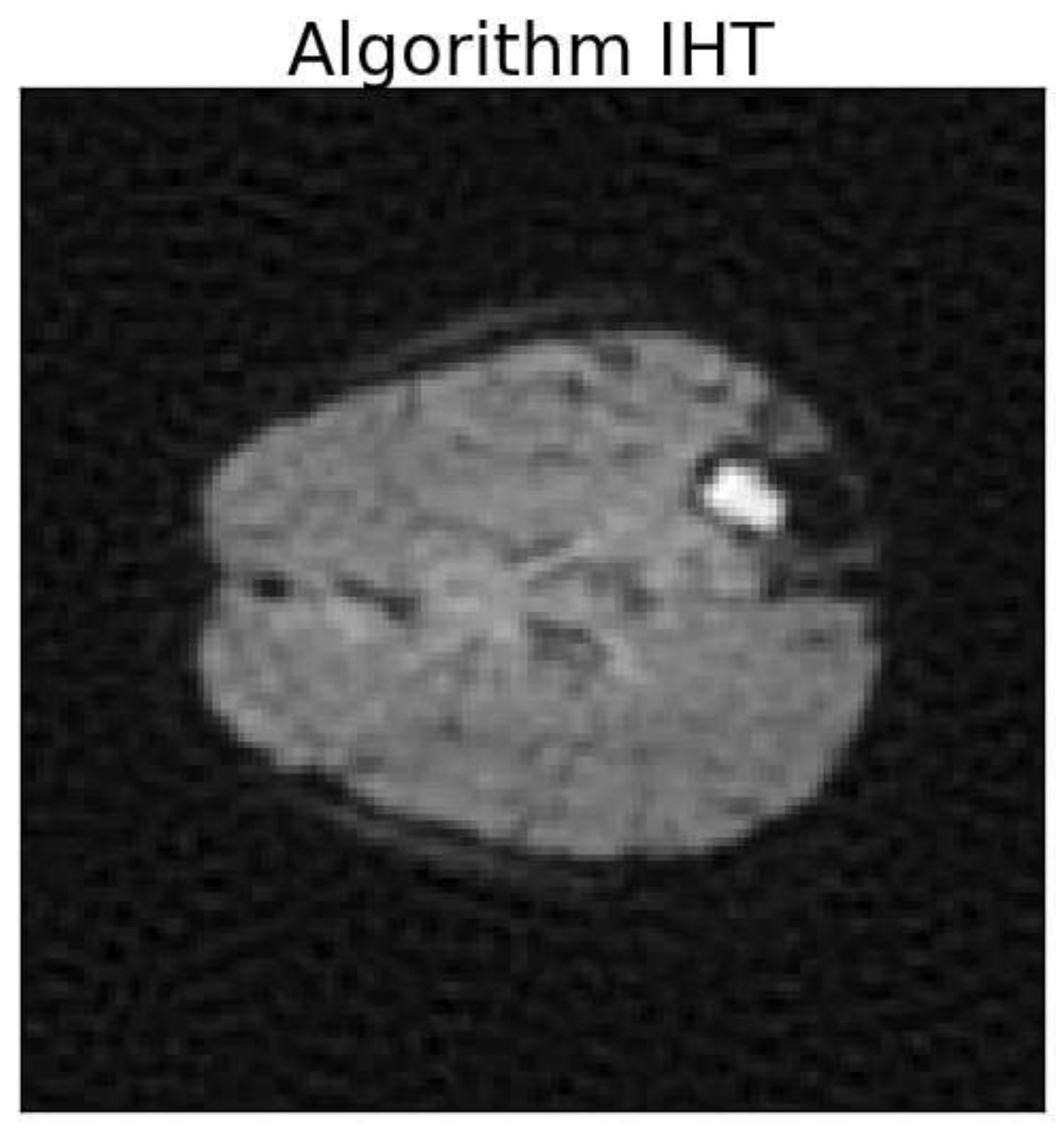}}
  {\includegraphics[scale=.27,clip]{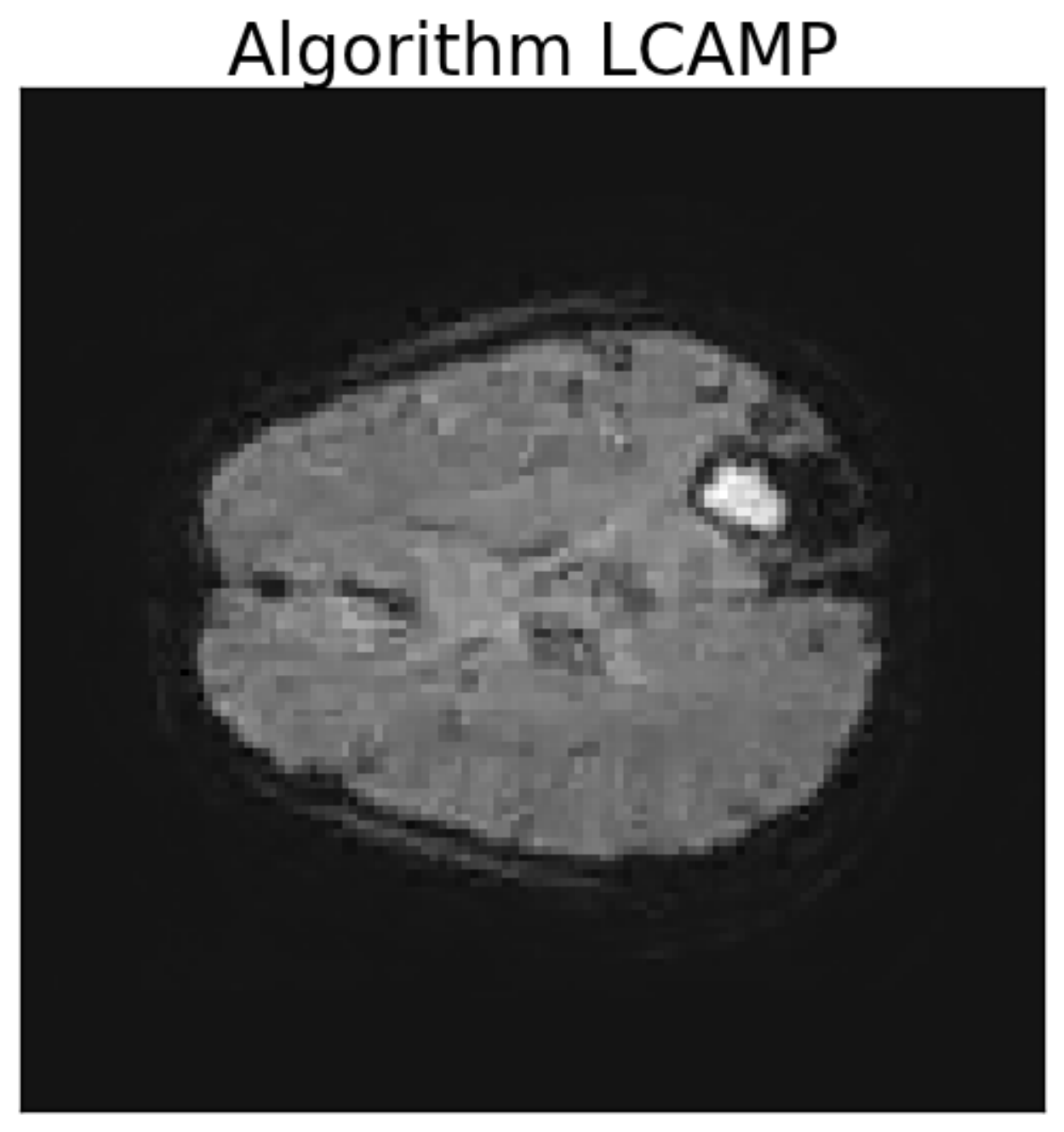}}
  \caption{Recovery of a particular frame of a real DSC-MRI (20\% of the total number of measures).}
  \label{fig:real_rec}
\end{figure}

The recovered images in the figure \ref{fig:real_rec} reflect the error table: note how the deterministic algorithms perform better then the CS algorithms recovering the higher resolution details.

\begin{table}[h]
\begin{center}
\begin{tabular}{|ccccc|}
\hline
Number of measures (on the total):    & 10\%   & 20\%   & 33\%  & 50\%  \\
\hline
1         & 1.78  & 1.20  & 0.77   & 0.45\\
2         & 2.10  & 1.66  & 1.37   & 0.99\\
3         & 1.79  & 1.22  & 0.77   & 0.45\\
4         & 1.91  & 1.44  & 0.95   & 0.82\\
IHT     & 4.18  & 2.40  & 1.41   & 0.63 \\
LCAMP     & 3.06  & 2.16  & 1.99   & 4.04 \\
\hline
\end{tabular}
\end{center}
\caption{Relative percentage errors of the real data set reconstructed by the algorithms with different number of measures.}
\label{tab:real}
\end{table}

\subsection{Phantom without noise}

The deterministic algorithms give the best reconstructions then the CS algorithms with any number of measures we tried. All of them seem to obtain similar results. Between the two CS algorithms LCAMP has the better results at a lower number of measures.

\begin{figure}[!htbp]
 \centering
  {\includegraphics[scale=.27,clip]{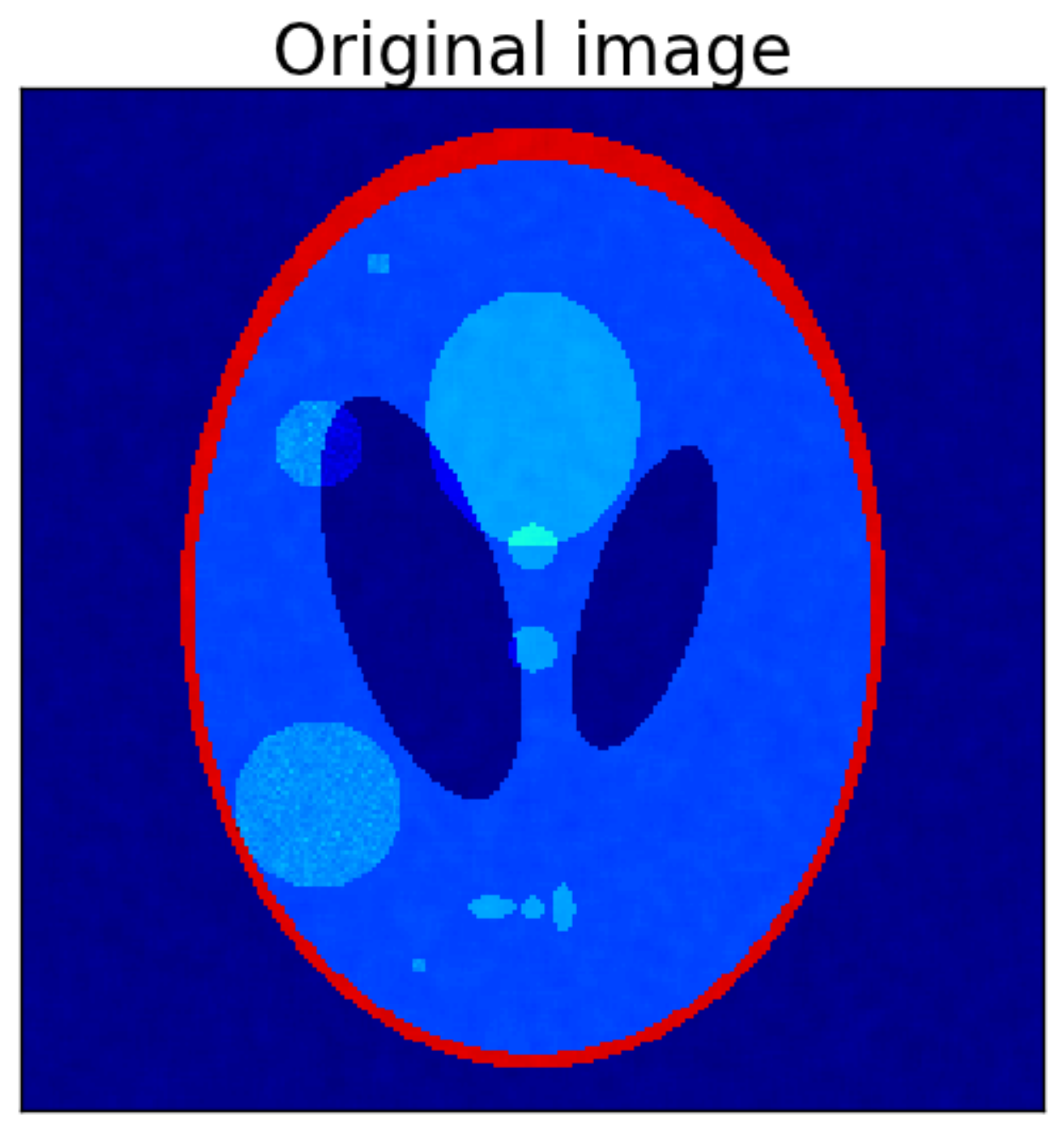}}\\
  {\includegraphics[scale=.27,clip]{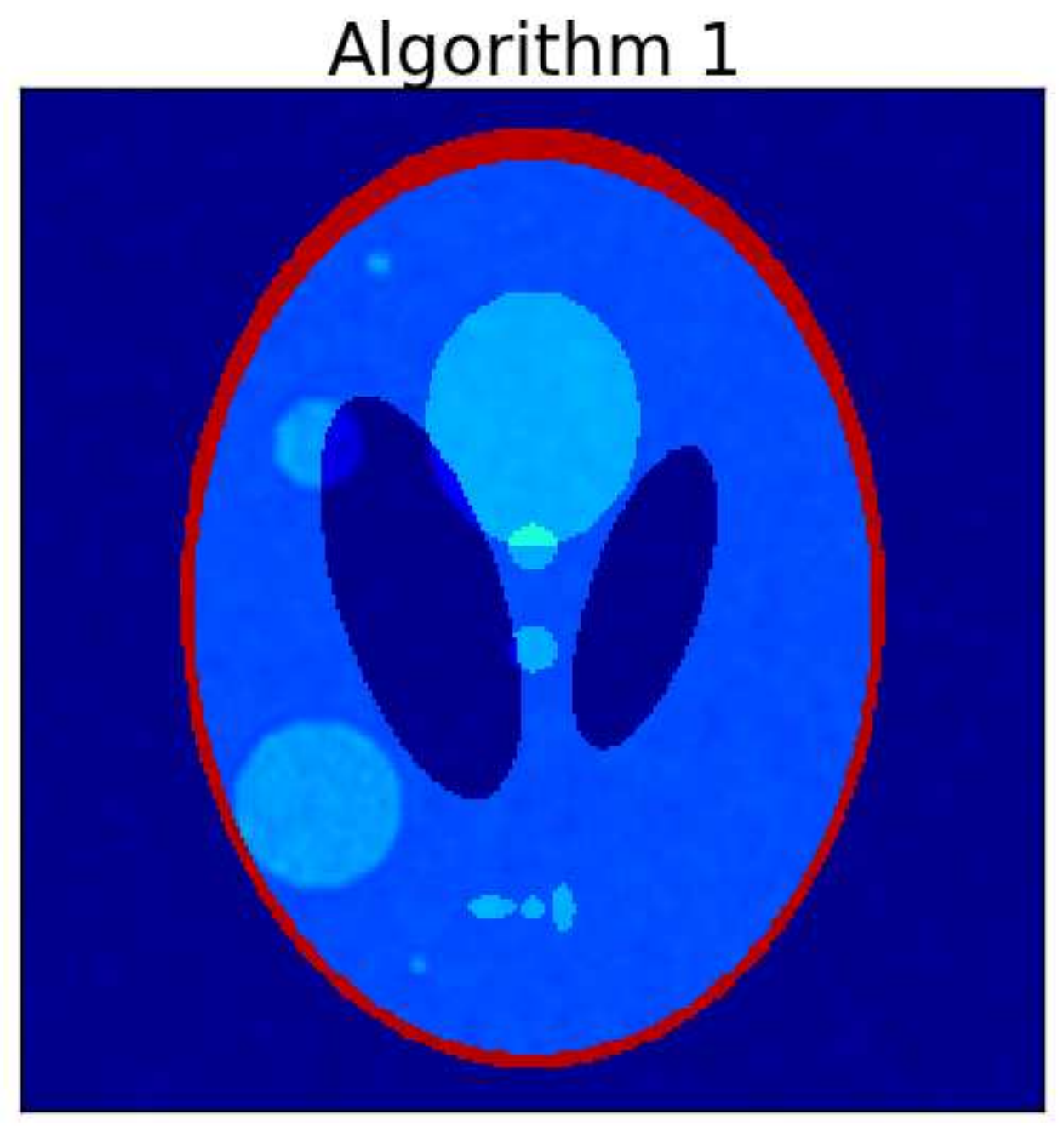}}
  {\includegraphics[scale=.27,clip]{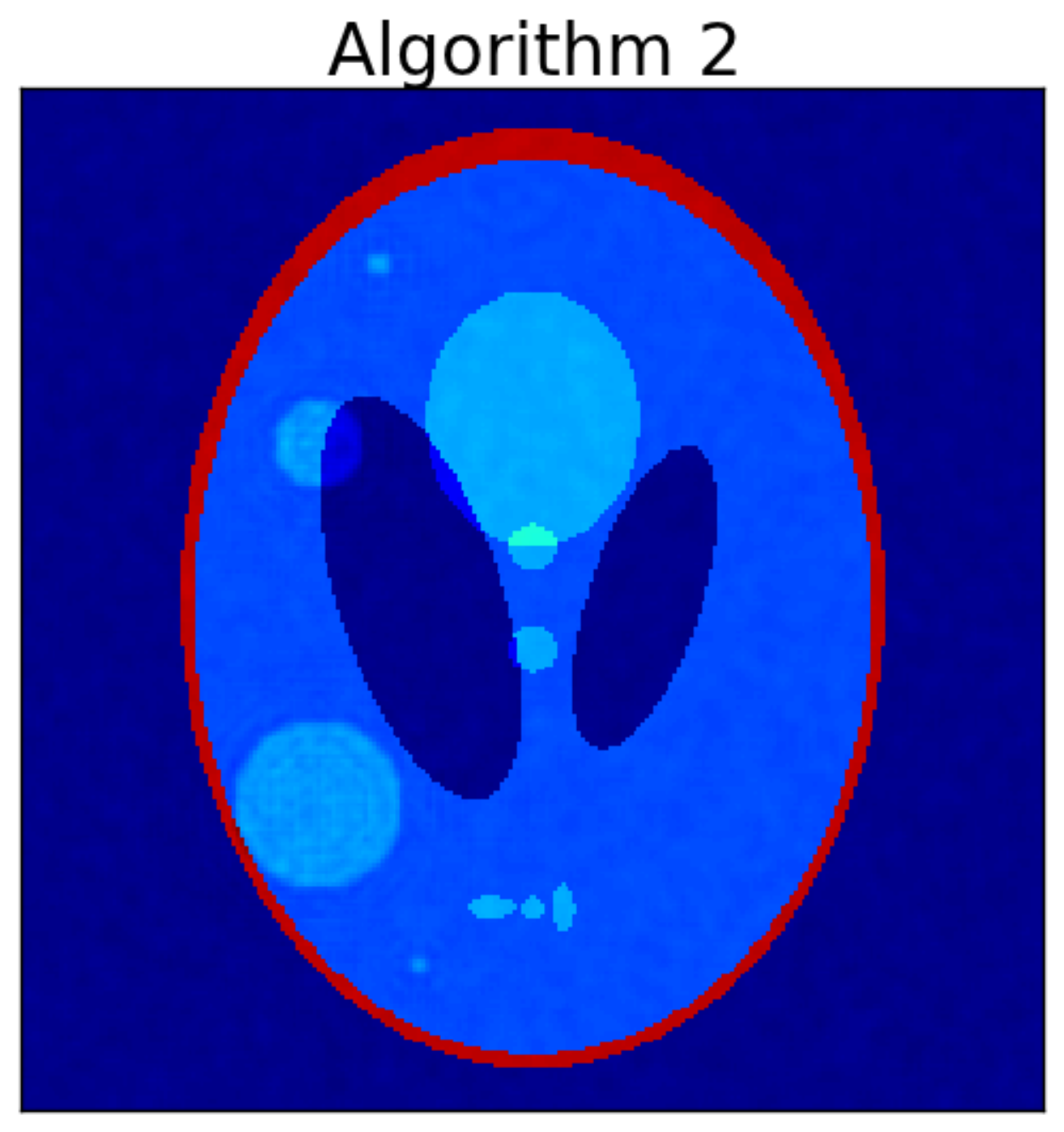}}
  {\includegraphics[scale=.27,clip]{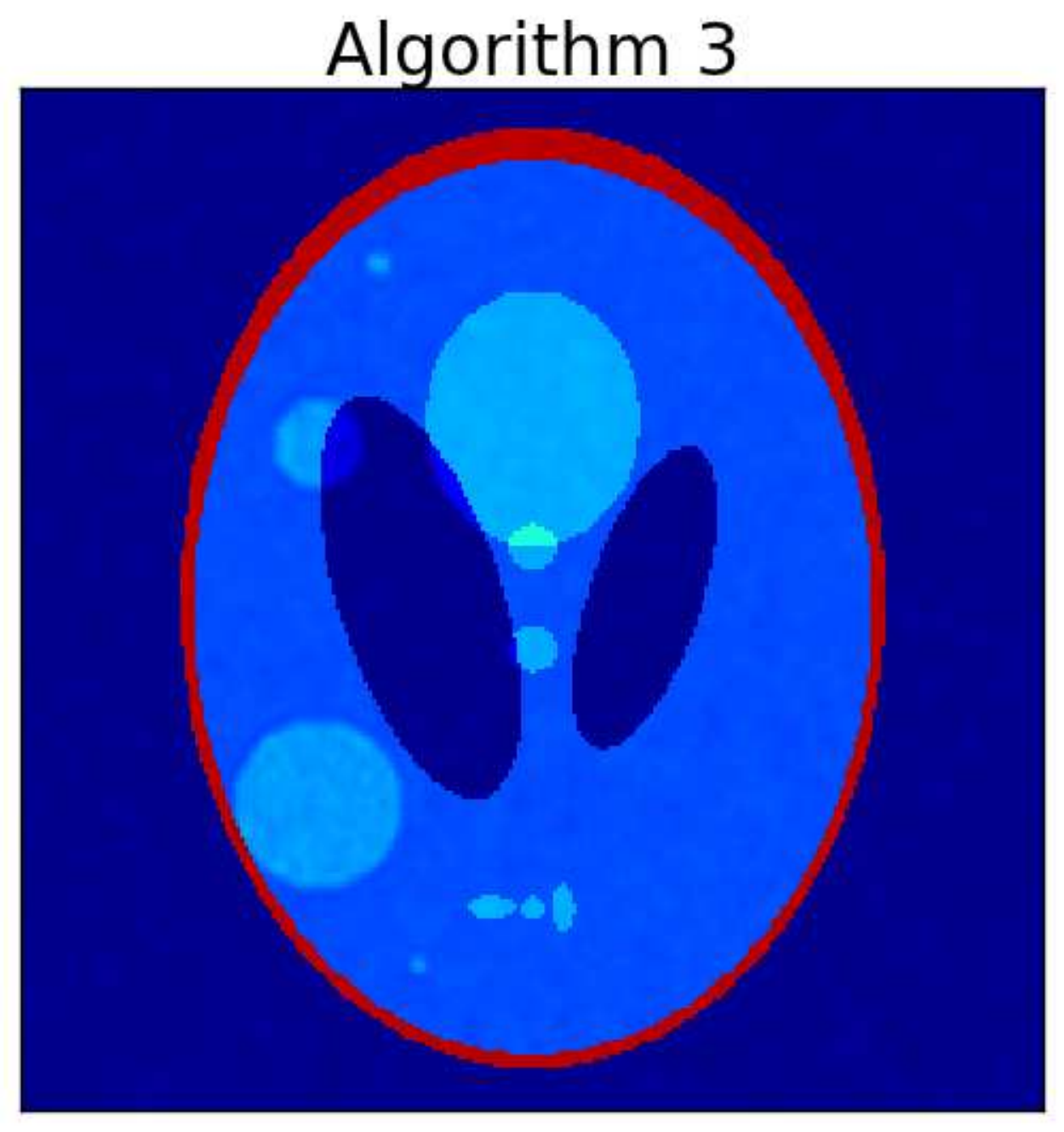}}\\
  {\includegraphics[scale=.27,clip]{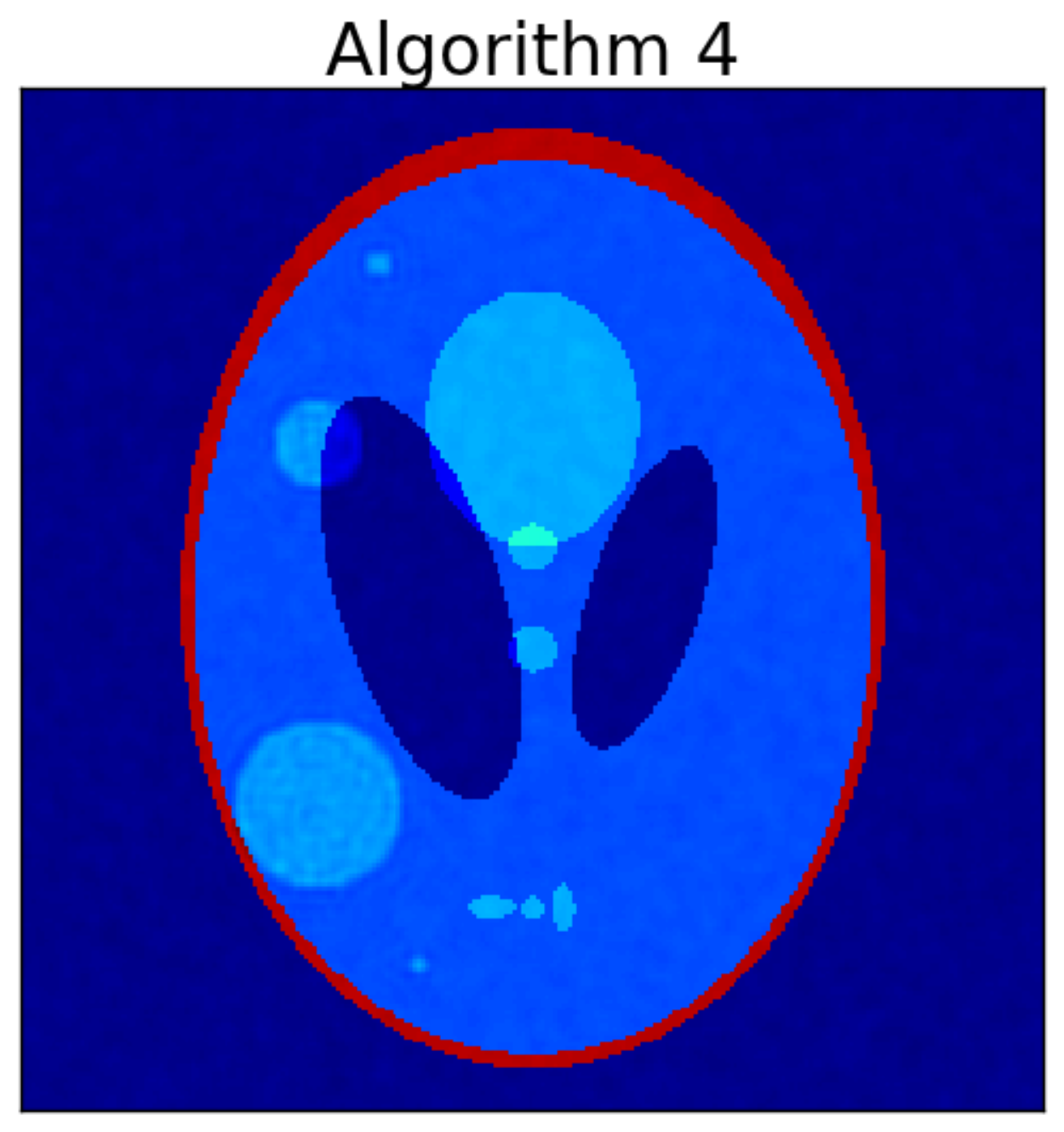}}
  {\includegraphics[scale=.27,clip]{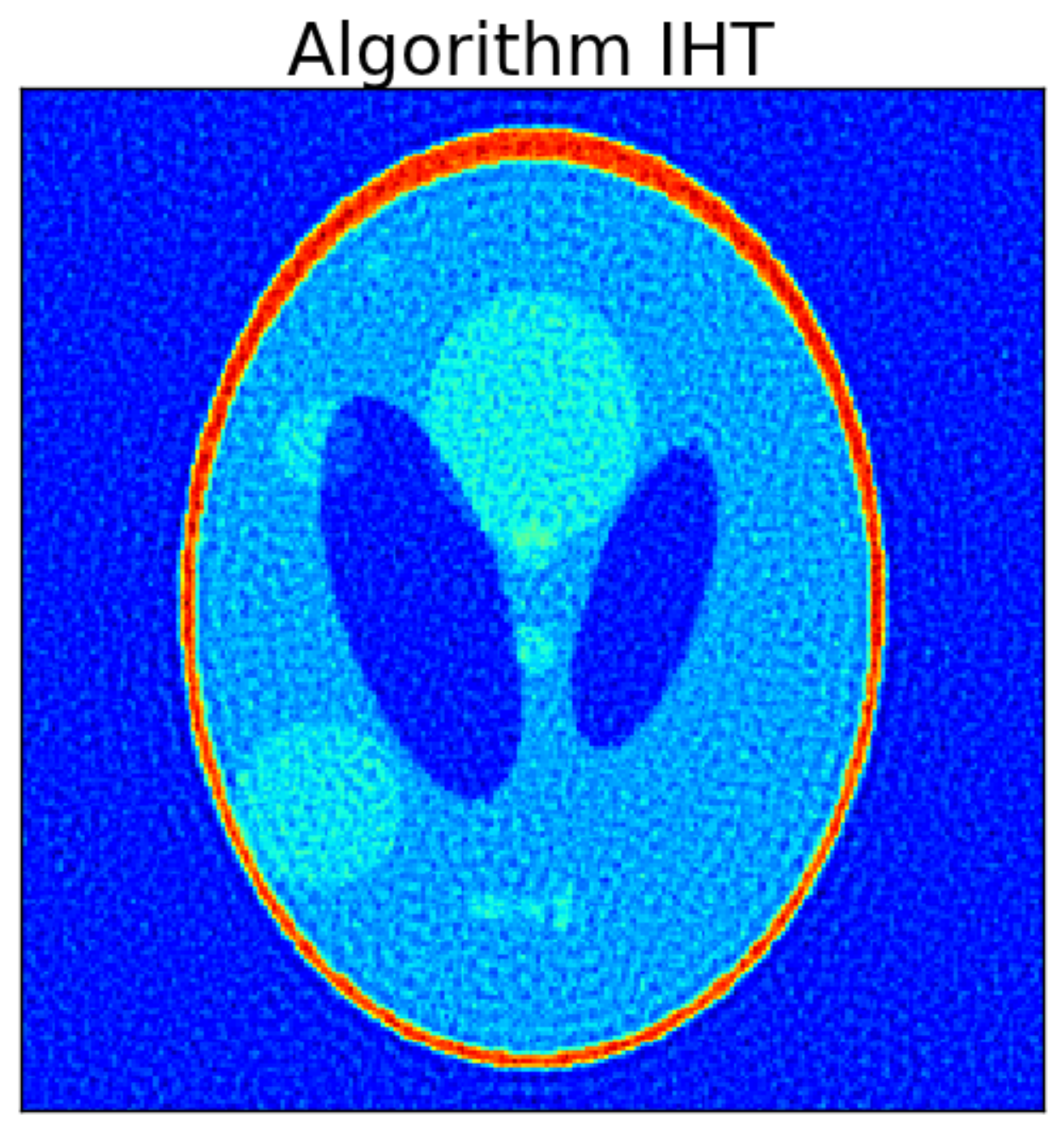}}
  {\includegraphics[scale=.27,clip]{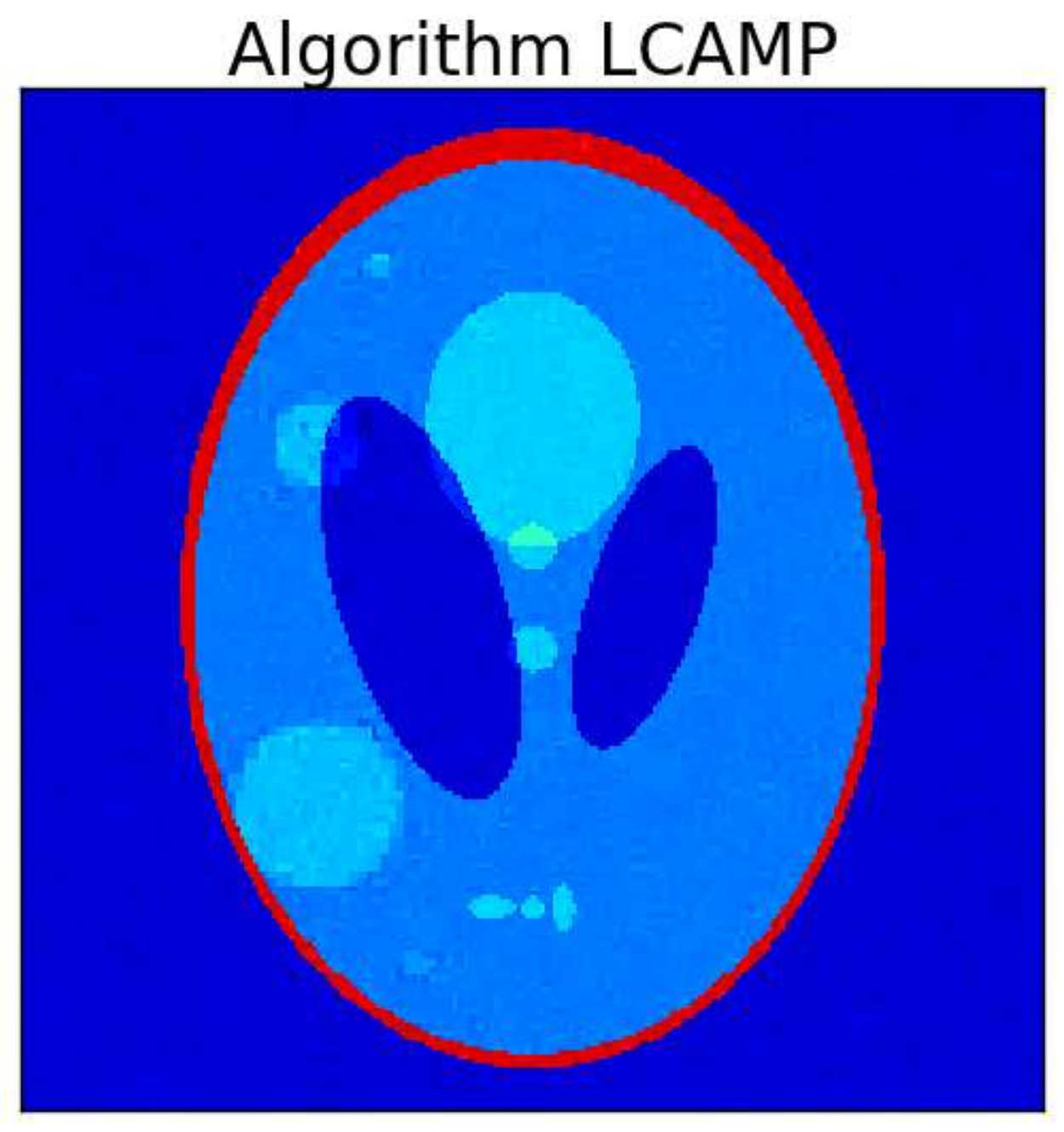}}
  \caption{Recovery of a particular frame of a simulated dataset. (20\% of the total number of measures).}
  \label{fig:phantom_rec}
\end{figure}

The recovered images in the figure \ref{fig:phantom_rec} reflect the error table. Note how the first four reconstructions are close to the original frame while the number of measures are only the 20\% of all the possible measures.

\begin{table}[h]
\begin{center}
\begin{tabular}{|ccccc|}
\hline
Number of measures (on the total):   & 10\%   & 20\%   & 33\%  & 50\%  \\
\hline
1         & 1.24  & 1.08  & 0.91   & 0.70\\
2         & 1.23  & 1.00  & 0.90   & 0.79\\
3         & 1.24  & 1.07  & 0.91   & 0.70\\
4         & 1.19  & 1.02  & 0.89   & 0.76\\
IHT     & 10.36  & 5.96  & 3.46   & 3.60\\
LCAMP     & 2.52  & 1.98  & 2.88   & - \\
\hline
\end{tabular}
\end{center}
\caption{Relative percent errors of the simulated dataset reconstructed by the algorithms with different number of measures.}
\label{tab:phantom}
\end{table}

\subsection{Phantom with noise}

The deterministic algorithms give the best reconstructions then the CS algorithms with any number of measures we tried. All of them seem to obtain similar results. Between the two CS algorithms, LCAMP has the best results at a lower number of measures. The errors are globally worse then the phantom without noise.

\begin{figure}[!htbp]
 \centering
  {\includegraphics[scale=.27,clip]{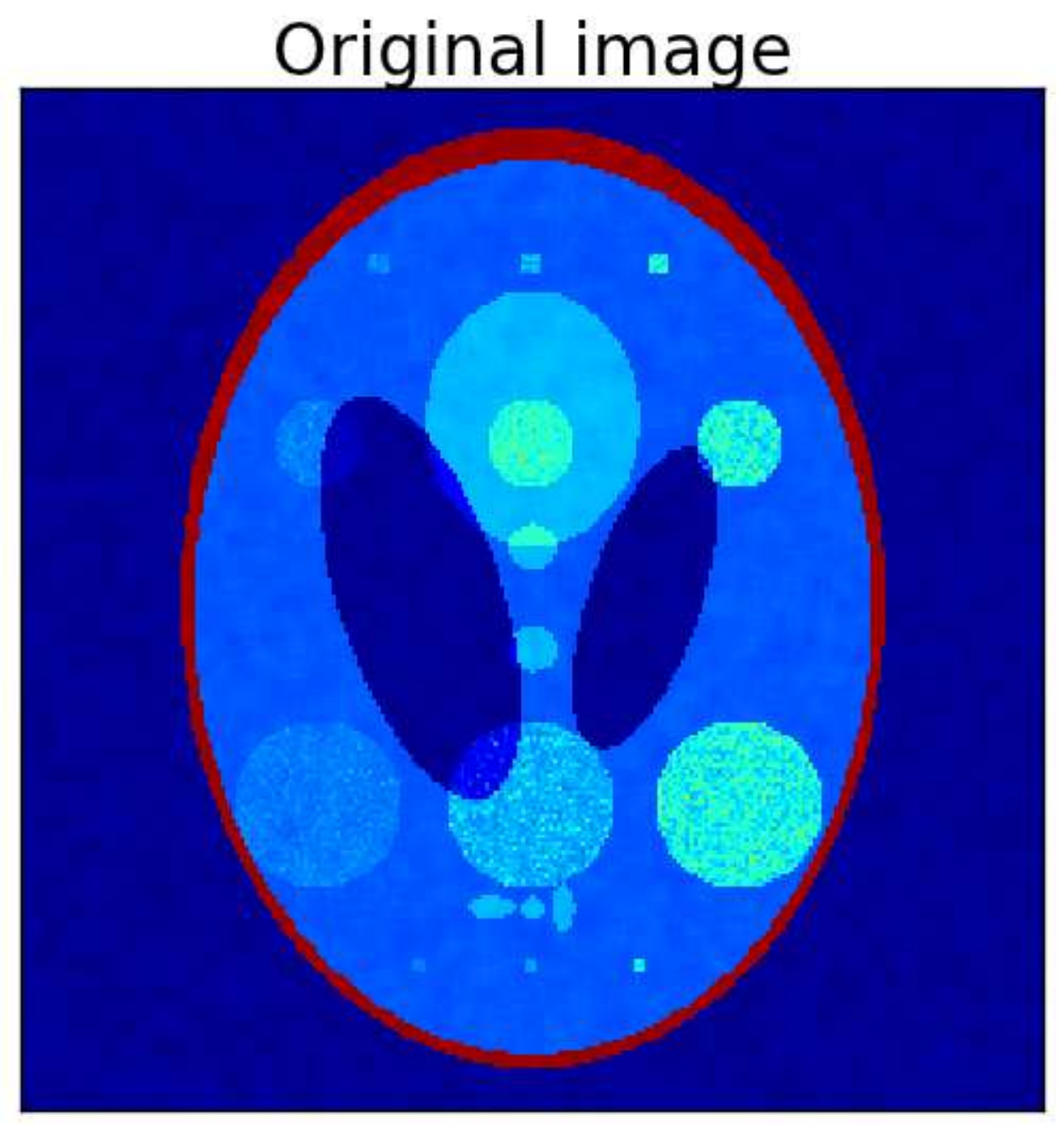}}\\
  {\includegraphics[scale=.27,clip]{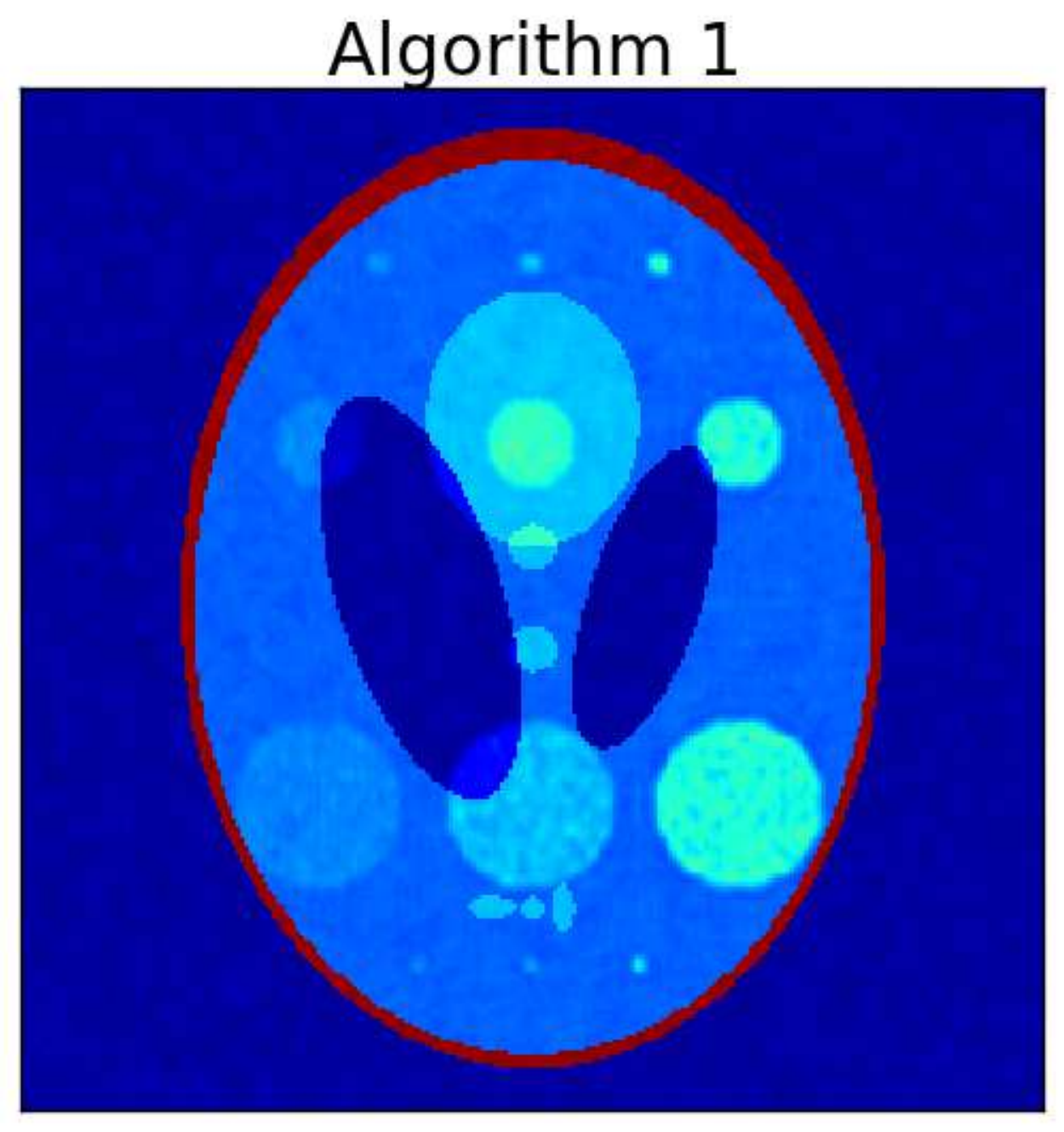}}
  {\includegraphics[scale=.27,clip]{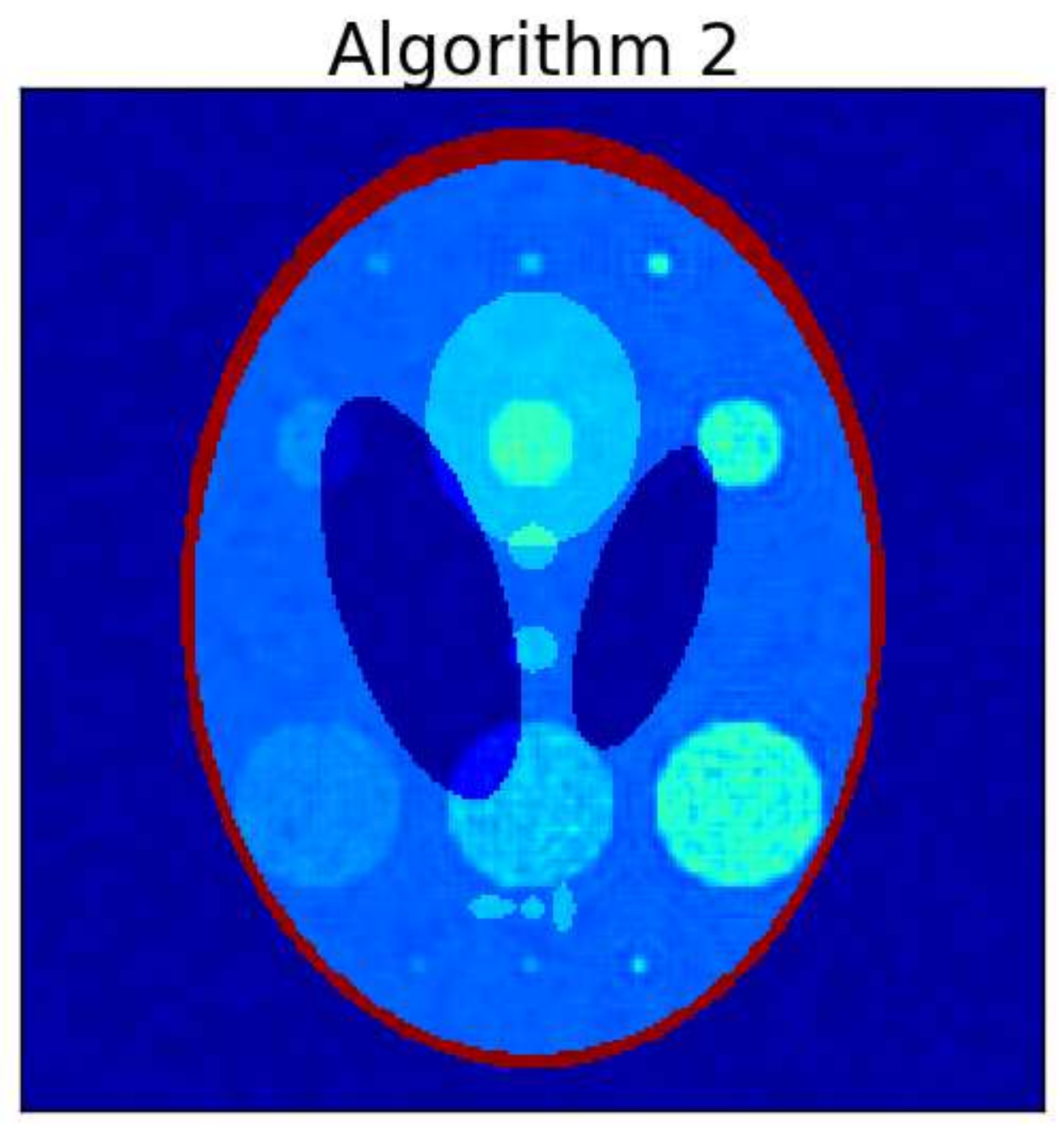}}
  {\includegraphics[scale=.27,clip]{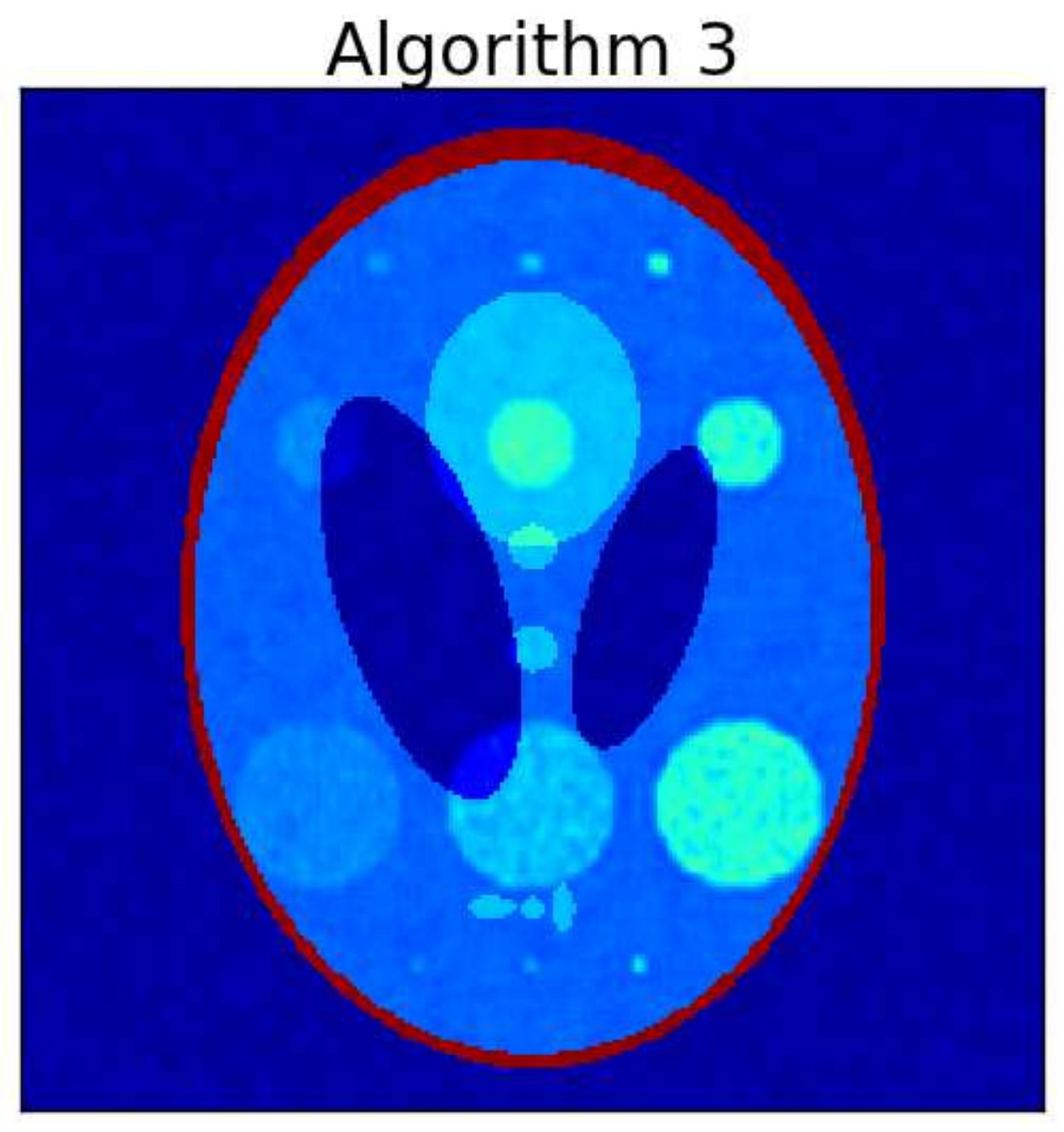}}\\
  {\includegraphics[scale=.27,clip]{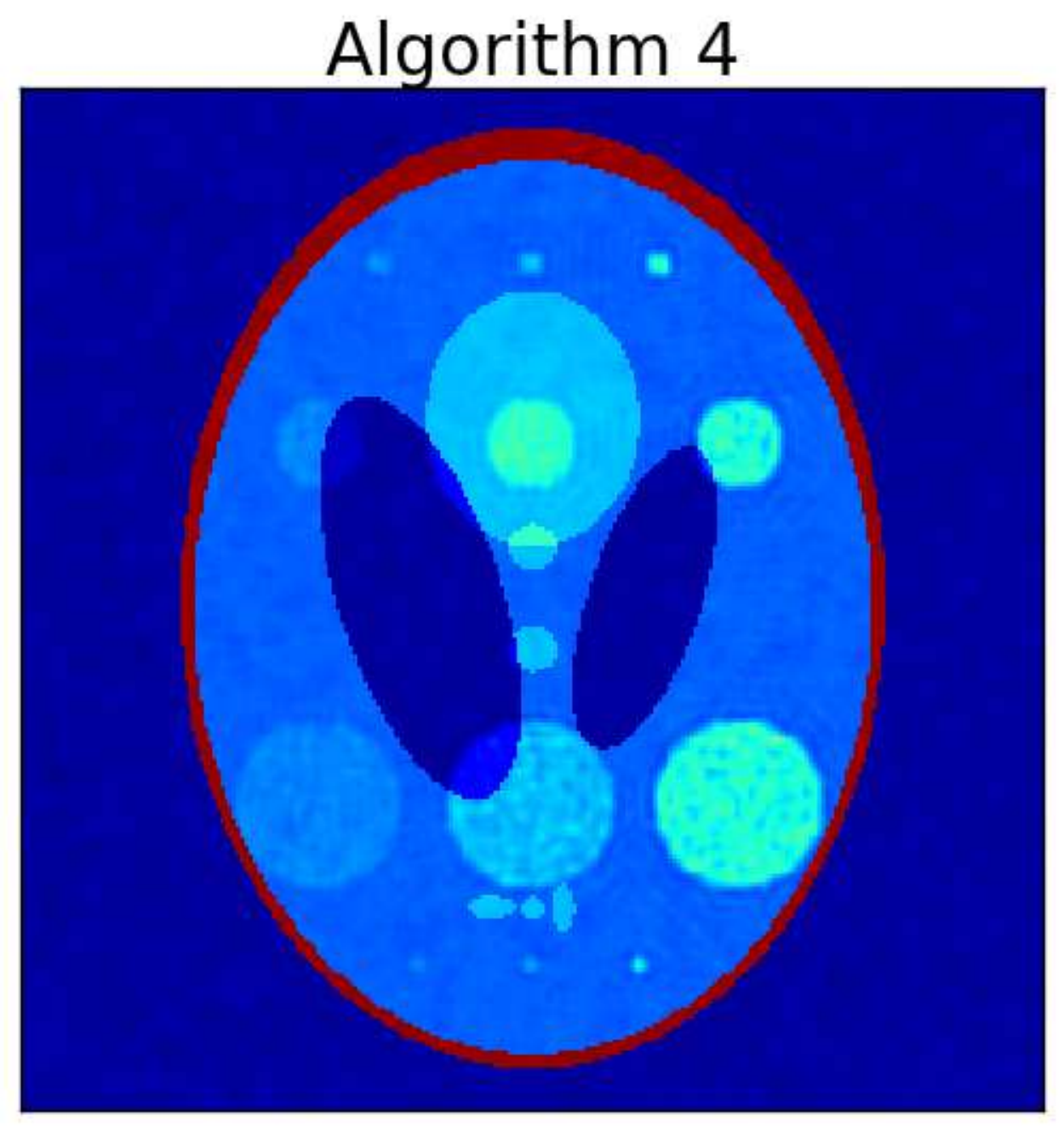}}
  {\includegraphics[scale=.27,clip]{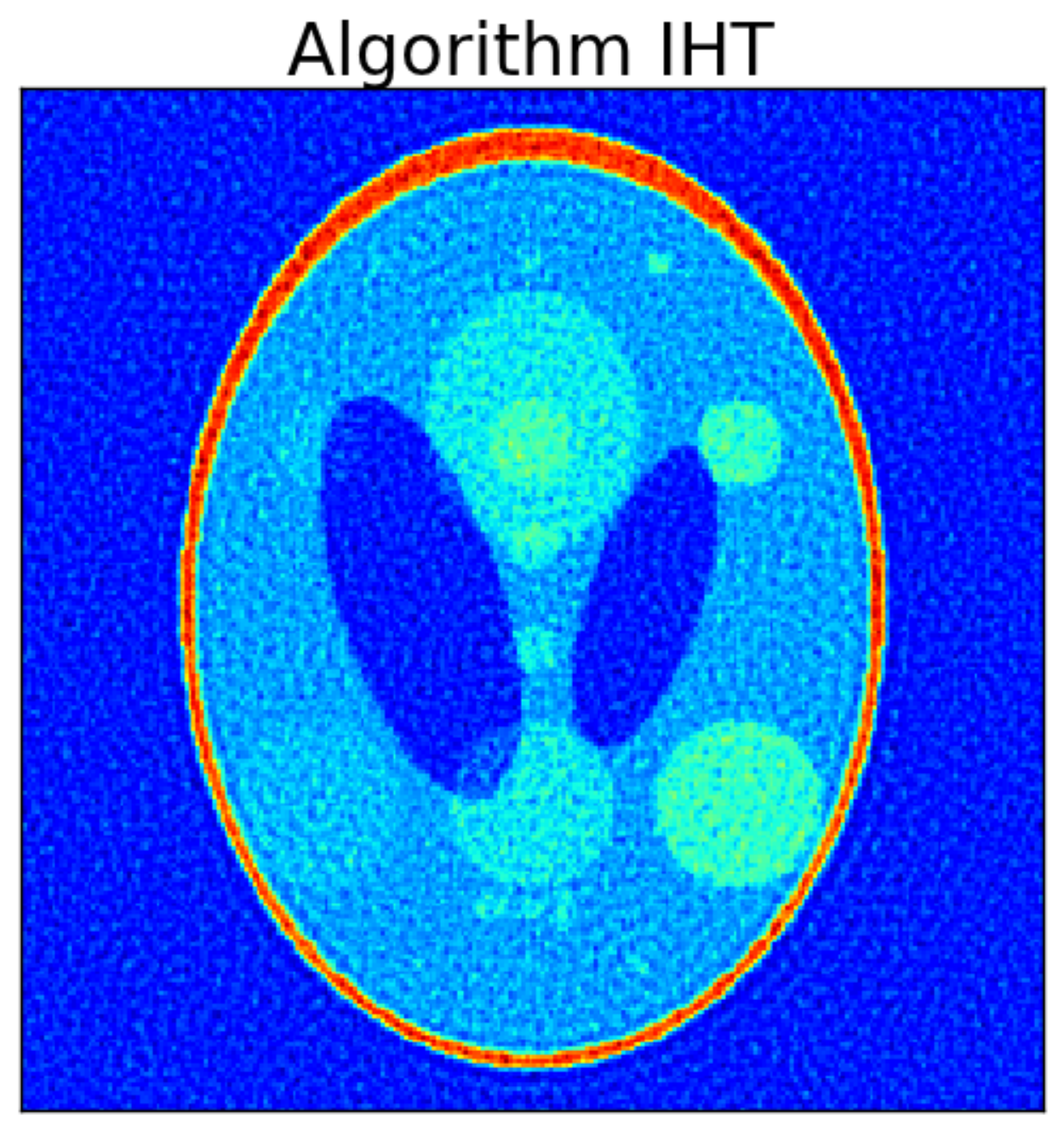}}
  {\includegraphics[scale=.27,clip]{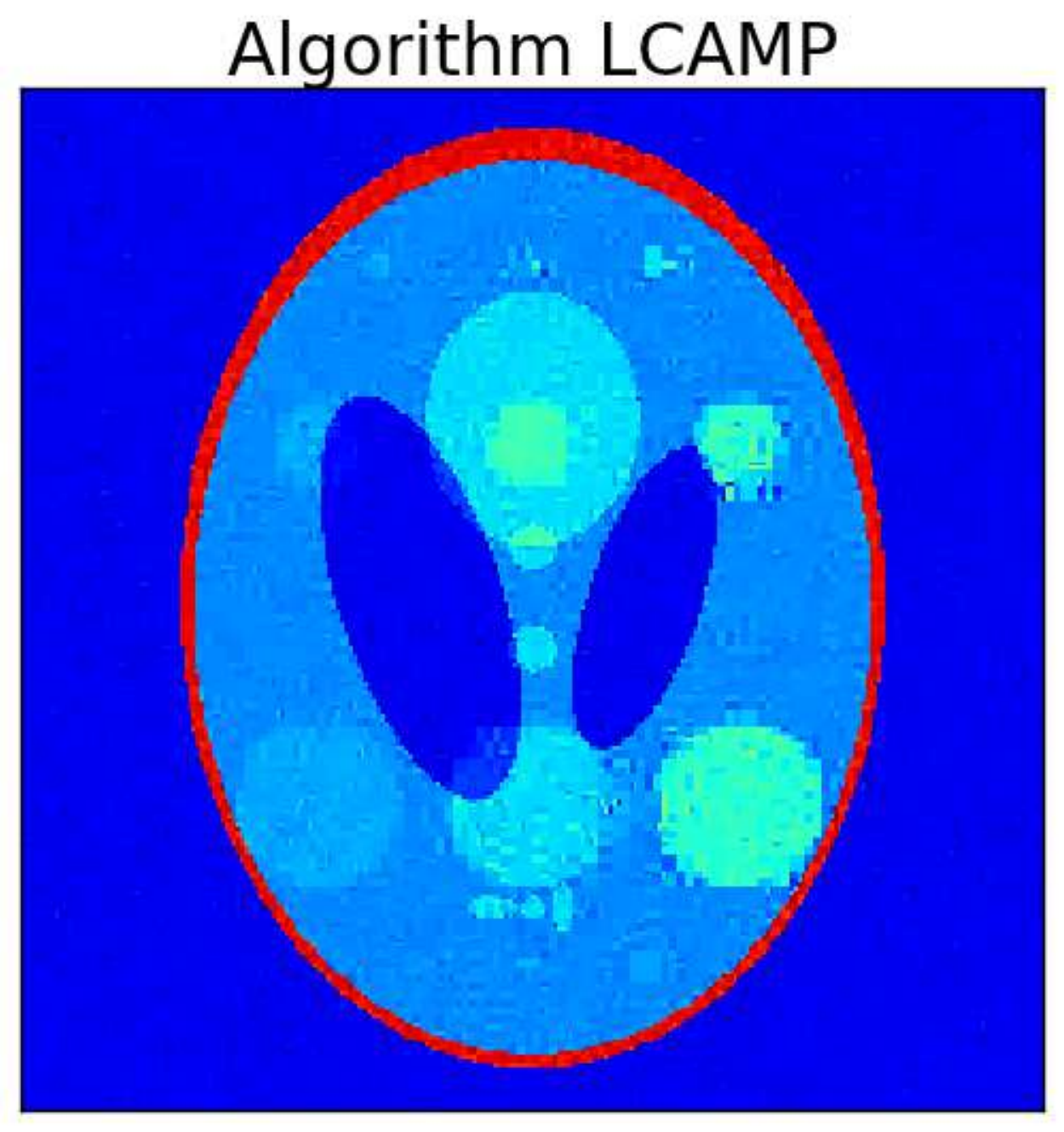}}
  \caption{Recovery of a particular frame of a simulated dataset with noise (20\% of the total number of measures).}
  \label{fig:phantomnoise_rec}
\end{figure}

The recovered images in the figure \ref{fig:phantomnoise_rec} are very similar to the previous ones, the added noise is not sufficiently intense to be visible.

\begin{table}[h]
\begin{center}
\begin{tabular}{|ccccc|}
\hline
Number of measures (on the total):   & 10\%   & 20\%   & 33\%  & 50\%  \\
\hline
1         & 1.80  & 1.56  & 1.32   & 1.01\\
2         & 1.79  & 1.45  & 1.30   & 1.15\\
3         & 1.80  & 1.56  & 1.32   & 1.01\\
4         & 1.74  & 1.48  & 1.31   & 1.14\\
IHT     & 12.67 & 5.82  & 3.51   & 3.34\\
LCAMP     & 3.59  & 2.87  & 3.58   & - \\
\hline
\end{tabular}
\end{center}
\caption{Relative percent errors of the simulated dataset (with noise) reconstructed by the algorithms with different number of measures.}
\label{tab:phantomnoise}
\end{table}

\subsection{Adaptive mask}

As we mentioned in section \ref{algorithms}, we tried to identify better measure sets $J$ by adapting the image that the algorithms receive as input with the last reconstructions.

We modified the algorithm 1 in a way that it updates the input mask $\bar{x}$ for every instant $t>\tau$ according to the formula (\ref{aggiornamento_mask}).

The table \ref{tab:adaptive_mask} shows the relative percent errors, in the $2$-norm, between reconstructed and original images, averaged over the whole sequence, in the case of real images with 10\% of measures.

\begin{figure}[!htbp]
 \centering
  {\includegraphics[scale=.4,clip]{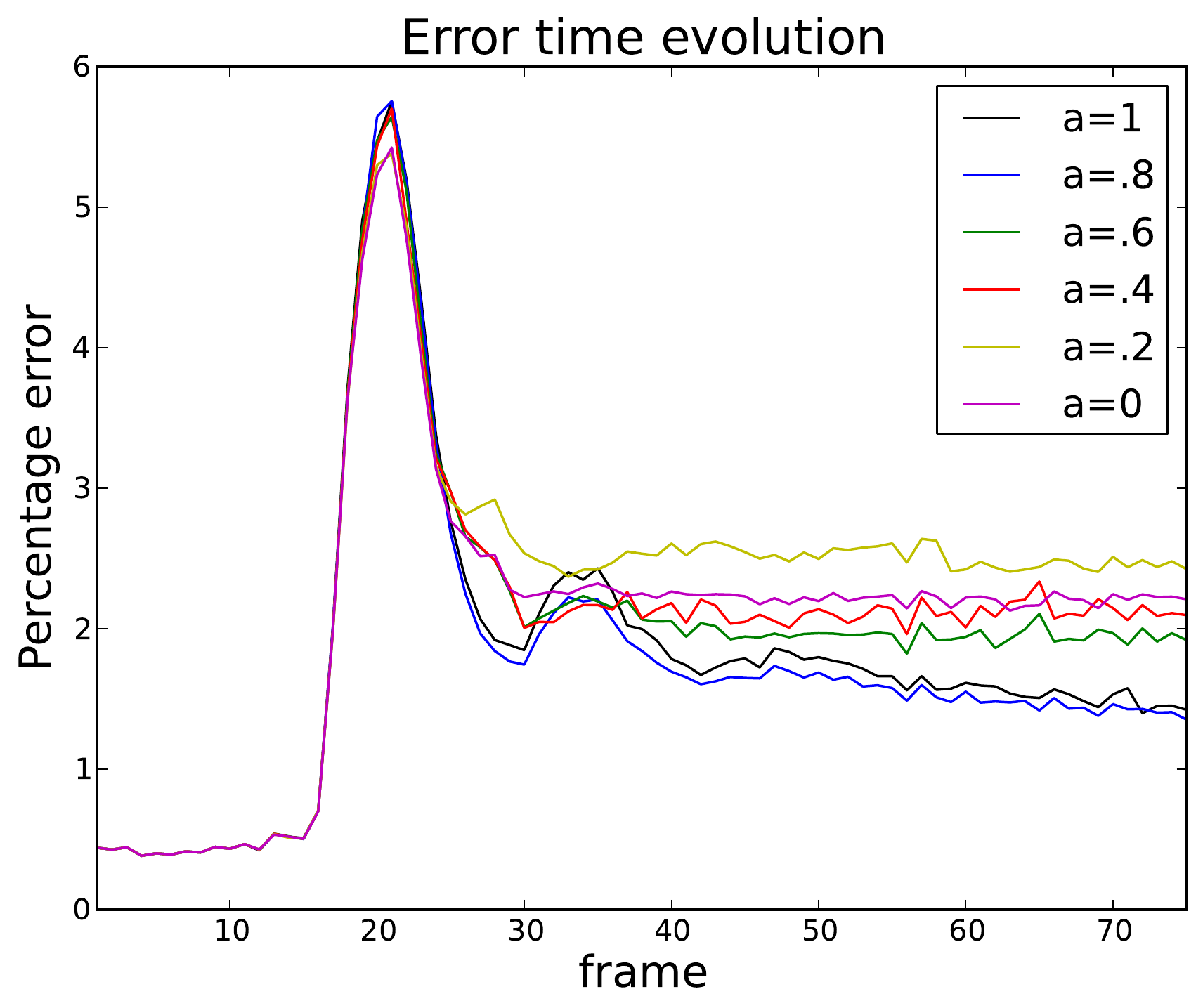}}
  \caption{Comparison of the error time evolution varying the parameter $a$.}
  \label{fig:adaptivemask}
\end{figure}

As $a$ decreases, the error lightly decreases too, then it increases again. The graph in figure \ref{fig:adaptivemask} shows the error time evolution varying the parameter $a$. Note that the most ``prompt'' algorithm ($a=0.0-0.2$) presents a greater global error, but it is sensibly more accurate in the peak corresponding to the moment at which the contrast fluid flows in the region, with an error reduction of about the 8\%. Then one can improve the tracking of the dynamic phase of the test by updating the mask $\bar{x}$. It is worth noting that from a diagnostic point of view, the late portion of the examination (approx. from frame 50 onward), is usually discarded from the analysis. Algorithms that are able to provide better performance in the first part of dynamic phase (rising edge, peak and washout) are preferable.

\begin{table}[h]
\begin{center}
\begin{tabular}{|cccccccccccc|}
\hline
a & 1.0 & 0.9 & 0.8 & 0.7 & 0.6 & 0.5 & 0.4 & 0.3 & 0.2 & 0.1 & 0.0 \\
\hline
err & 1.78 & 1.74 & 1.71 & 1.82 & 1.95 & 2.01 & 2.02 & 2.29 & 2.26 & 2.16 & 2.07 \\
\hline
\end{tabular}\\
\end{center}
\caption{Relative percent errors of the real sequence reconstructed by the modified algorithm 1 with different values of the parameter $a$.}
\label{tab:adaptive_mask}
\end{table}

\section{Conclusions}
\label{conclusions}

\begin{itemize}
\item The adoption of techniques deriving from the mathematical analysis of the DSC-MRI problem enabled us to develop algorithms for the reconstruction of MRI images that perform sensibly better than the random based approaches cited in the literature.
\item The results are sensitive to the error function adopted, e.g. the best images are not necessarily the most accurate in the $2$-norm. Improvements in this direction must be obtained, in our opinion, with a joint work with the physicians.
\item We have also noticed that these results can still be improved by introducing a dynamic model of the problem, that could be improved with an estimator of the direction and speed of the contrast propagation in the time-spatial domain rather than only in the time domain, see e.g. \cite{marcuzzi2} for an example. With this aim, it should be relevant to exploit the fluid dynamics of the contrast liquid to sharpen the prediction.
\end{itemize}

\section*{Acknowledgments}

The first author is supported by the project ex-60\% of the University of Padua, named ``Metodi numerici per problemi inversi in meccanica computazionale''.

\end{document}